\pdfoutput=1
\documentclass[11pt]{article}
\usepackage[preprint]{acl}

\usepackage{array}
\usepackage{xcolor}
\usepackage{booktabs}
\usepackage{colortbl} 
\usepackage{hyperref}
\usepackage{times}
\usepackage{latexsym}
\usepackage[T1]{fontenc}
\usepackage[utf8]{inputenc}
\usepackage{microtype}
\usepackage{inconsolata}
\usepackage{graphicx} 
\usepackage{lscape}
\usepackage{multirow}
\usepackage{listings}
\usepackage{pdfpages}
\usepackage{tikz}
\usepackage{booktabs}
\usepackage{amsmath}
\usepackage{amssymb}
\usepackage{float}
\usepackage{CJKutf8}
\usepackage{booktabs}
\usepackage{array}
\usepackage{geometry}
\usepackage{hyperref}
\usepackage{kotex}
\usepackage{tcolorbox}
\usepackage{cleveref}
\Crefname{section}{\S}{\S\S}

\tcbuselibrary{skins, breakable, theorems}
\usepackage{subcaption}

\tcbset{
    myparagraph/.style={
    colback=blue!5, 
    colframe=blue!75,  
    width=\textwidth,  
    boxrule=0.5mm,  
    arc=4mm,  
    left=4mm, 
    right=4mm,  
    top=2mm, 
    bottom=2mm, 
    fonttitle=\bfseries,
    title={Prompt},
    breakable
    }
}

\tcbset{
    incontextexample/.style={
    colback=blue!5, 
    colframe=blue!75,  
    width=\textwidth,  
    boxrule=0.5mm,  
    arc=4mm,  
    left=4mm, 
    right=4mm,  
    top=2mm, 
    bottom=2mm, 
    fonttitle=\bfseries,
    title={L1 Knowledge Injection Prompt},
    breakable
    }
}

\tcbset{
    traitanalysis/.style={
    colback=blue!5, 
    colframe=blue!75,  
    width=\textwidth,  
    boxrule=0.5mm,  
    arc=4mm,  
    left=4mm, 
    right=4mm,  
    top=2mm, 
    bottom=2mm, 
    fonttitle=\bfseries,
    title={Trait Analysis Prompt},
    breakable
    }
}

\tcbset{
    annotations/.style={
    colback=blue!5, 
    colframe=blue!75,  
    width=\textwidth,  
    boxrule=0.5mm,  
    arc=4mm,  
    left=4mm, 
    right=4mm,  
    top=2mm, 
    bottom=2mm, 
    fonttitle=\bfseries,
    title={Annotation Prompt},
    breakable
    }
}

\tcbset{
    mydialogue/.style={
    colback=blue!5, 
    colframe=blue!75,  
    width=\textwidth,  
    boxrule=0.5mm,  
    arc=4mm,  
    left=4mm, 
    right=4mm,  
    top=2mm, 
    bottom=2mm, 
    fonttitle=\bfseries,
    title={Dialogue},
    breakable
    }
}

\newcommand{\mychar}[1]{
  \begingroup\normalfont
  \includegraphics[height=0.5cm]{#1}%
  \endgroup
}

\definecolor{deepgreen}{rgb}{0.1, 0.5, 0.1}
\providecommand{\dec}[1]{{\color{red} #1}}
\providecommand{\inc}[1]{{\color{deepgreen} #1}}

\title{Can LLMs Simulate L2-English Dialogue? \\ An Information-Theoretic Analysis of L1-Dependent Biases}

\author{Rena Gao$^\heartsuit$\thanks{Equal contribution.}, Xuetong Wu$^\heartsuit$$^{*}$, Tatsuki Kuribayashi$^\diamondsuit$, Mingrui Ye$^\clubsuit$, Siya Qi$^\clubsuit$ \\
\textbf{Carsten Roever$^\heartsuit$, Yuanxing Liu$^\spadesuit$, Zheng Yuan$^\clubsuit$, Jey Han Lau$^\heartsuit$} \\ 
\begin{tabular}{l l}  
  $^\heartsuit$The University of Melbourne & $^\clubsuit$King's College London \\
   $^\diamondsuit$MBZUAI     & $^\spadesuit$Harbin Institute of Technology \\
\end{tabular}
\\
\texttt{\{rena.gao,carsten\}@unimelb.edu.au, \{wfyitf,jeyhan.lau\}@gmail.com} \\ 
\texttt{\{tatsuki.kuribayashi\}@mbzuai.ac.ae}, \texttt{\{yxliu\}@ir.hit.edu.cn} \\
\texttt{\{mingrui.ye, siya.qi, zheng.yuan\}@kcl.ac.uk}
}

\begin{document}
\maketitle

\begin{abstract} 
This study evaluates Large Language Models' (LLMs) ability to simulate non-native-like English use observed in human second language (L2) learners interfered with by their native first language (L1). In dialogue-based interviews, we prompt LLMs to mimic L2 English learners with specific L1s (e.g., Japanese, Thai, Urdu) across seven languages, comparing their outputs to real L2 learner data. Our analysis examines L1-driven linguistic biases, such as reference word usage and avoidance behaviors, using information-theoretic and distributional density measures. Results show that modern LLMs (e.g., Qwen2.5, LLAMA3, DeepseekV3, GPT 4o) replicate L1-dependent patterns observed in human L2 data, with distinct influences from various languages (e.g., Japanese, Korean, and Mandarin significantly affect tense agreement, and Urdu influences noun-verb collocations). Our results reveal LLMs’ potential for L2 dialogue generation and evaluation for future educational applications.  
\end{abstract}

\section{Introduction}\label{sec:intro}  
\begin{figure}[t]
    \centering  \includegraphics[width=0.9\linewidth]{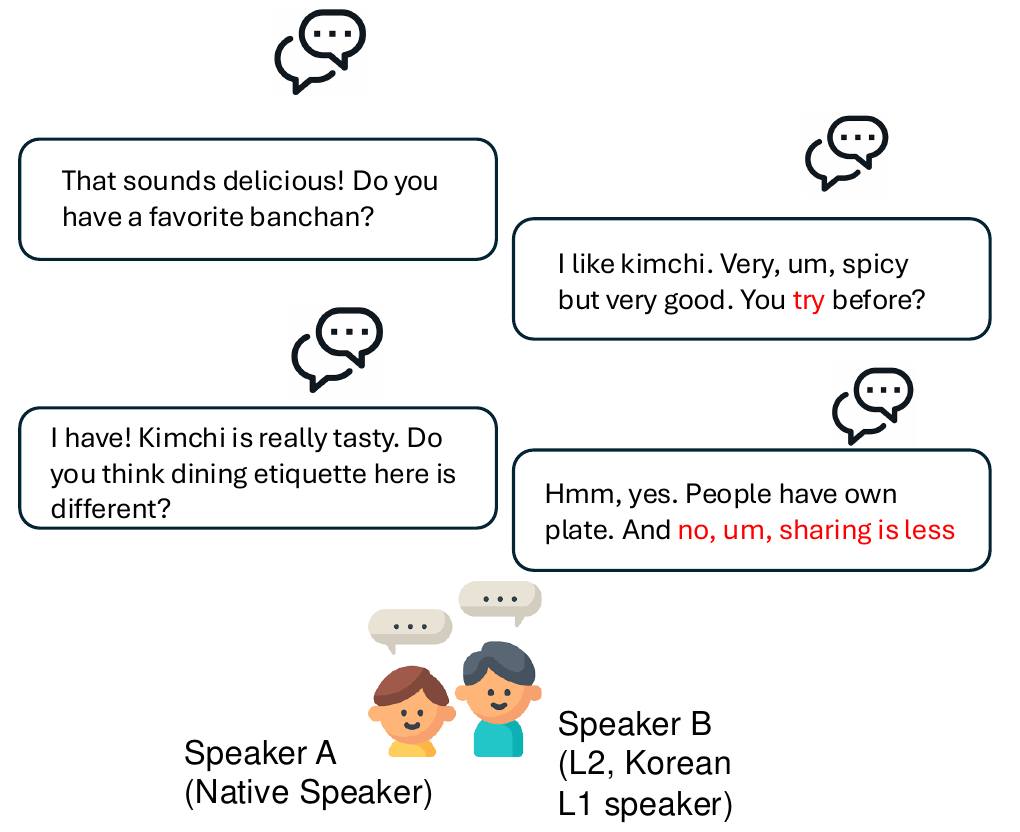}
    \caption{Examples of L2 English dialogue from human speakers, which can generally be biased by their native L1 knowledge, e.g., with particular \textcolor{red}{errors}.}
    \label{fig:L2example}
\end{figure}

The widespread use of Large Language Models (LLMs) in language communication and education has opened opportunities to study their ability to simulate human-like language, particularly in second language (L2) communication~\cite{liang2024controllable,cherednichenko2024large}, as illustrated by Figure~\ref{fig:L2example}. 
Such an L2-speaker simulation will be helpful for, e.g., predicting L2 speakers' biases in a pedagogical situation~\cite{settles-etal-2018-second}, developing an L2-speaking agent~\cite{timpe2022using}, virtual language-learning applications~\cite{bibauw2022dialogue}, emulating diverse agents to simulate the diversity of L2 speakers in real world~\cite{ge2024scaling}, and potentially assessing LLMs' cognitive plausibility from a cross-lingual perspective~\cite{aoyama-schneider-2024-modeling}. 
However, the ability of LLMs to accurately replicate linguistic patterns of non-native speakers and the systematic influence of L1 knowledge on L2 generation remain underexplored~\cite{chen2024roleinteract}, especially in the dialogue domain~\cite{veivo2025dialogue} and in non-native contexts~\cite{fincham2024using}. This leads us to ask: \textbf{Can LLMs effectively mimic human-like dialogue performance in L2 contexts?}

To address this question, it is crucial to understand the role of native linguistic knowledge in areas such as language education and cross-lingual communication~\cite{levenston1971over,Schachter1974-dq,Kleinmann1977-vt,brooke-hirst-2012-measuring}. L2 speakers' use of English is often influenced by their L1 traits~\cite{takahashi2024l1}, especially for Asian native speakers whose first language (L1) is diverse from English~\cite{pan2024impoliteness}, resulting in distinct L1-L2 linguistic patterns~\cite{bailey2021digital}, including grammatical constructions and lexical choices in spoken dialogues~\cite{aoyama2024modeling,downey2023embedding}. To investigate whether LLMs simulate similar patterns, we propose an information-theoretic evaluation framework grounded in multiple linguistic perspectives:  key features from grammatical/semantic accuracy, fluency, discourse-level cohesion, and pragmatics that shape the communicative outcome~\cite{schwandt2001understanding,sun2021conversations,gao2024listenership,santiago2023inclusion}. By analyzing these aspects, we explore how accurately various LLMs simulate L2-like dialogues that align with human linguistic behaviors across different L1 backgrounds.\footnote{All codes and dataset can be found \url{https://github.com/RenaGao/LLMPirorknowledge}} For benchmark data, we utilize the ICNALE dataset~\cite{ishikawa2023icnale}, which includes recordings from 435 human L2 speakers with 18 L1s and manual transcripts comprising approximately 1.6M tokens. 
An information-theoretic analysis is applied to evaluate LLMs' L1-dependent biases by comparing LLM-generated dialogues (with a prompt to simulate L2 English with a specific L1) with human counterparts. 
To address challenges in reflecting L1 background in LLMs' L2 generation, as an initial foray, we employ native knowledge injection prompting (e.g., \texttt{Simulate L2 English dialogue spoken by Japanese based on provided Japanese native linguistic knowledge})~\cite{dong2022survey,santiago2023inclusion,bibauw2022dialogue}.

Through our exploration, we demonstrate that simple L1 prompting has a significant impact on LLM-generated L2 dialogues. For example, Japanese, Korean, and Mandarin L1 influence tense agreement, Thai and Malay L1 affect speech acts, while Urdu L1 impacts noun-verb collocations.
Our information-theoretic evaluation quantifies their human-like output, which is further supported by qualitative analysis. 
Ultimately, our study paves the way for using LLMs to simulate human L2 dialogues. Summarizing our contributions: 
\begin{itemize}
    \item We propose a new evaluation framework with eight linguistic features, covering grammatical/semantic accuracy, fluency, cohesion, and pragmatics perspectives, designed to evaluate the impact of L1 information on LLM-generated dialogues. This framework enables systematic analysis of how native language traits (in humans/LLMs) shape linguistic features in cross-lingual dialogue generation.
    \item We further propose an information-theoretic metric to quantify L1 influence on LLM dialogue generation, revealing L1-dependent differences such as \emph{reference word}, \emph{modifiers} and \emph{numerals} usages.
    \item We show that, through prompting, LLMs can generate dialogues with varying degrees of non-native-like linguistic features influenced by different L1s, paving a new way for LLMs to simulate L2 communications.
\end{itemize}

\section{Related Work} \label{sec:related-work}
\subsection{Bilingual Knowledge for LLMs}
\paragraph{L1 interference in humans and LMs} 
Native language profoundly influences L2 language use in humans~\cite{levenston1971over,Schachter1974-dq,Kleinmann1977-vt,brooke2013native}. This \textit{language interference} effect biases, for example, the syntactic constructions~\cite{felker2021role} and discourse flows ~\cite{bailey2021digital} in L2, and the dialogue patterns are not an exception ~\cite{veivo2025dialogue}.
When it comes to neural LMs, the cross-lingual transferability of LMs and their human-likeness has also gained attention, but prior studies have exclusively focused on sentence-level evaluations~\cite{oba-etal-2023-second,yadavalli-etal-2023-slabert,elshin2024general}. 
Such perspectives can easily be extended to the dialogue level, involving discourse-level cohesion/coherence and L1-dependent, nuanced differences in dialogue strategy~\cite{abe2019interactional,gao2024interaction}.
Moreover, LLMs are now deployed to generate dialogue (e.g., chat interactions); evaluating their ability in a dialogue scenario generally aligns with their practical usage~\cite{jin2024better,veivo2025dialogue}. That said, our scope is limited to just simulating L2-like language use in a behavioral sense; LM's cognitive plausibility as an L2 learner, while interesting and related, is beyond of the scope of this paper.

\paragraph{Bilingual Knowledge in LLMs}
Bilingual knowledge typically impacts LLM in cross-lingual and multilingual tasks ~\cite{miah2024multimodal}. 
For example, leveraging shared grammatical features, bilingual LLM excels with typologically similar language pairs like English-Spanish, improving coherence and fluency through transfer learning ~\cite{jeon2022information}. On the other hand, handling distant cross-lingual pairs, such as English-Chinese, poses challenges (i.e., negative language transfer) due to differences in their grammatical features such as word order~\cite{ranaldi2023does}, requiring targeted training and alignment of grammatical constructs~\cite{pvribavn2024comparative}. 
In the context of dialogue tasks, limited L2 dialogue data and linguistic inconsistencies sometimes hinder LLM performance for non-native English speakers to interact~\cite{gan2024clarq}. There are case studies that optimize bilingual knowledge integration and enhance cross-lingual grammatical understanding~\cite{huzaifah2024evaluating}, as well as improve LLMs' ability to generate accurate and coherent dialogue, benefiting non-native English users~\cite{han2024llm}.

\subsection{Evaluation of L2 Capabilities of LLMs} 
Evaluating human-like LLMs is a key focus in educational NLP. Studies explore their use in online platforms~\cite{manoharanmaximizing}, personalized language tutoring~\cite{mejeh2024taking}, and L2 chatbots~\cite{yigci2024large}, but often rely on human judgments due to the complexity of L2 dialogues. Some worked propose automated evaluation tools for L2 interactions~\cite{gao-etal-2025-interaction} and language practice~\cite{huzaifah2024evaluating}, yet LLM performance of generation of non-native language in these settings remains under explored.


\paragraph{Mimicking Human-like L2 Dialogues} 
Developing effective L2 dialogue generation systems requires a robust evaluation framework that facilitates linguistic knowledge transfer from L1 to L2, particularly for Asian L1 speakers with distinct syntactic structures from English~\cite{li-qiu-2023-finding}. Such a framework is crucial for integrating prior linguistic competence, enabling models to generate more context-aware utterances~\cite{sung-etal-2024-context,gao2024cnima}. To address this, evaluation protocols should incorporate cross-linguistic benchmarking and error analysis to identify language-specific grammatical challenges~\cite{kobayashi-etal-2024-large}. Systematic analysis of these errors provides insights into LLMs’ bilingual grammatical understanding and representation, ensuring they not only grasp cross-lingual constructs but also generate language-specific nuances, enhancing real-world multilingual applications~\cite{cong2025demystifying,gao2024interaction,singh2024three,poole2024llm}.

\section{Evaluation Metrics} \label{sec:design}  
\subsection{Evaluation Framework}
\label{sec:evaluation-framework}

To assess whether LLMs can accurately simulate L2 dialogues,
we target eight linguistic constructs to evaluate their L2 English generation ability, motivated by L1--L2 interference research~\cite{jackson2018production,taguchi2020second,milliere2024language,gao-etal-2025-interaction}.
The constructs cover both structural and functional aspects of languages, including \emph{reference word} usage to assess their cohesion, \emph{noun and verb collocations} to capture native-like lexical patterns, and various forms of \emph{agreement} such as \emph{number}, \emph{tense}, and \emph{subject-verb} consistency, which are critical for grammatical accuracy. Additionally, pragmatic constructs like \emph{speech acts} and \emph{modal verbs and expressions} evaluate contextually appropriate language use in dialogues, reflecting cultural and linguistic nuances often influenced by L1 conversations. Together, these metrics provide a comprehensive framework to measure the effectiveness of LLM-generated L2 dialogues, identifying both strengths and areas for improvement in cross-lingual dialogue generation. We summarize these  constructs in Table~\ref{tab:linguistic_constructs}.

\begin{table*}[t]
    \centering
    \small
    \begin{tabular}{p{1.5cm}p{2.3cm}p{4.3cm}p{2.2cm}p{2.6cm}}
        \toprule
        Categories & Features & Definition & Example & Examples in Prompt \\
        \cmidrule(lr){1-3} \cmidrule(lr){4-4} \cmidrule(lr){5-5}
        \multirow{3}{*}{\shortstack[l]{Grammatical \\ Accuracy}} 
        & Number Agreement (in noun phrase)
        & Adjectives/determiners and nouns must agree in number (sometimes involves grammatical gender, e.g., ``la/las'' and ``el/los'' in Spanish). 
        & \textit{[100] cars} 
        & \textit{``The big cars are red.''...} \\
         \cmidrule(lr){2-3} \cmidrule(lr){4-4} \cmidrule(lr){5-5}
        & Tense Agreement 
        & The verb tense (i.e., past, present, future) must align with temporal expressions. 
        & \textit{I [did] a task [yesterday].} 
        & \textit{``He has finished his homework.''...} \\
         \cmidrule(lr){2-3} \cmidrule(lr){4-4} \cmidrule(lr){5-5}
        & Subject-Verb Agreement 
        & The verb form must agree with the subject's person and number. 
        & \textit{[She] [is] amazing.} 
        & \textit{``They are playing football.''...}  \\
                \cmidrule(lr){1-3} \cmidrule(lr){4-4} \cmidrule(lr){5-5}
        \multirow{2}{*}{\shortstack[l]{Semantic \\ Accuracy}} 
        & Modal Verbs and Expressions
        & Modal verbs that indicate likelihood, ability, permission, or obligation. 
        & \textit{She [might] come to the meeting.} 
        & \textit{``You should complete the project soon.''...} \\ 
         \cmidrule(lr){2-3} \cmidrule(lr){4-4} \cmidrule(lr){5-5}
        & Quantifiers and Numerals
        & Numerical expressions or those related to the amounts, such as quantifiers. 
        & \textit{Some, many, a few} 
        & \textit{``There are ten apples on the table.''...} \\
        \cmidrule(lr){1-3} \cmidrule(lr){4-4} \cmidrule(lr){5-5}
        Fluency & Noun-Verb Collocations 
        & Common collocations that enhance sentence fluency. 
        & \textit{[Drive] a [car]}, \textit{[Do] a [test]} 
        & \textit{``He drives a car every day.''...} \\
        \cmidrule(lr){1-3} \cmidrule(lr){4-4} \cmidrule(lr){5-5}
        Cohesion 
        & Reference Word 
        & Linguistic devices referring to entities mentioned earlier (anaphora) or later (cataphora). 
        & \textit{She}, \textit{her}, \textit{him}, \textit{he} 
        & \textit{``She went home early.''...} \\
        \cmidrule(lr){1-3} \cmidrule(lr){4-4} \cmidrule(lr){5-5}
        Pragmatics 
        & Speech Acts
        & Utterances that serve special functions, such as assertions, questions, requests, or commands. 
        & \textit{``Could you open the window?''} (Indirect request) 
        & \textit{``Can you help me with this task?''...}  \\
        \bottomrule
    \end{tabular}
    \caption{Linguistic features targeted in our L2-like dialogue generation capability tests for LLMs} \label{tab:linguistic_constructs}
\end{table*}

\subsection{Information-Theoretic Metrics}
\label{ssec:information_theory}
\paragraph{Overview}
We quantify how similar the specific L2-English usages simulated by LLMs are to those exhibited by human L2-English speakers.
This is quantified by a particular information-theoretic distance between the dialogues produced by those two groups (LLMs vs. humans); that is, the smaller the score is, the better the LLMs could simulate the real L2-English speakers' patterns. 


\paragraph{Theoretical Introduction}
We propose an information-theoretic framework to explore how a person's first language influences their use of a second language. We use a random variable $Y$ to represent a specific linguistic phenomenon, as shown in Table~\ref{tab:linguistic_constructs}, and the distribution $p(Y)$ describes how frequently this phenomenon occurs.\footnote{Thus, our focus is on the avoidance behavior of L2 speakers~\cite{levenston1971over,Schachter1974-dq,Kleinmann1977-vt}, and analyzing the correctness of the phenomenon is left to be our future work.} 
In the case of L1 English acquisition, the English language exposures $D$ are generated by $Y$, following the likelihood $p(D|Y)$. By combining $p(Y)$ and $p(D|Y)$, we model the learning of English dialogue structure through the posterior $p(Y|D)$, which quantifies how well an English L1 learner can infer $Y$ from $D$. 
Now, extending this to L2 English acquisition/learning, we define another random variable $X$ to represent linguistic properties for L1 \textbf{human} native speakers of that L1 language (non-English). 
Here, $X$ acts as a \textbf{priori knowledge}. 
According to L2 development theory (\citealt{roever2024relationship}; \textit{inter alia.}), L2 learners usually acquire a new language by interacting with native speakers and learning from the linguistic patterns present in the spoken input. 
With the English (L2) language exposures $D$ and the effect of L1 properties $X$, the updated posterior for learning $Y$ for L2 becomes $p(Y|D, X) \propto p(D|Y, X)p(Y|X)$ with the assumption that $p(D|Y,X) = p(D|Y)$ as the context $D$ hinges solely on the English linguistic properties $Y$, which also incorporates the \textbf{prior distribution} $p(Y|X)$. The human-like $p(Y|D,X)$ is estimated with the dialogues produced by the real human L2-English speakers of L1 natives (\cref{sec:dataset}). Then, when it comes to LLMs with the respective L1 and L2, their L1 prior knowledge (and their general learning bias) is noted as $X'$. Our focus is on whether they can have a human-like $X$ (that is, similar to $X'$) when prompted to behave like respective human L2 speakers, which is analyzed through the lens of the L2 behavior similarity between LLMs' $p(Y|D,X')$ and humans' $p(Y|D,X)$. These differences are quantified as a density between these distributions in our experiments. Mathematically, we can characterize this difference with the logarithmic loss function $\ell(Q) = -\log Q$, leading to the following evaluation\footnote{The dependence of $D$ on all $X$, $X'$, and $Y$ ensures that the posterior $p(Y|D, X)$ differs from $p(Y|D, X')$ due to the different priors $p(Y|X)$ and $p(Y|X')$. }:

\begin{small}
\begin{align*}
  &d = \mathbb{E}_{XX'YD}\left[\ell(p(Y|D, X')) - \ell(p(Y|D, X))\right] \\
   & = \mathbb{E}_{XYD}\left[\log\frac{p(Y|D, X)}{p(Y|D)}\right] - \mathbb{E}_{X'YD}\left[\log\frac{p(Y|D, X')}{p(Y|D)}\right]  \\
   &= I(X;Y|D) - I(X';Y|D)
\end{align*}
\end{small}%
where $I(X;Y|D)$ quantifies the mutual information between $X$ and $Y$ given $D$, and it represents the shared information between English ($Y$) and the native language ($X$) conditioned on the context $D$. Similarly, $I(X';Y|D)$ measures the effectiveness of LLMs in generating L2 English by quantifying the mutual information between the LLM's native language ($X'$) and English ($Y$) given the context $D$.\footnote{We leave it as future work to align $D$ between humans' and LLMs' L1/L2 learning --- an important topic that investigates the cognitive/developmental plausibility of LM's language learning ability.}
We report $d_\mathrm{bi}$ in the case that LLMs are instructed to mimic an L2-English speaker with the respective L1. As a baseline, we also compute $d_\mathrm{mono}$ when no valid L1 information is provided to LLMs as $X'$.
We will consistently use $\ell(Q) = -\log Q$ in our experiments.

\section{Evaluation Framework Annotation Design}\label{sec:dataset}

We now describe how we annotate the linguistics constructs for dialogues based on the evaluation framework in
\cref{sec:evaluation-framework}.
We used a hybrid approach combining automated methods with manual review. This annotation process targeted eight key linguistic constructs that influence dialogue construction from grammatical accuracy to pragmatics, as outlined in Table~\ref{tab:linguistic_constructs}. To this end, we utilize the International Corpus Network of Asian Learners of English (ICNALE) dataset~\cite{ishikawa2018icnale}, which includes dialogue response utterances from speakers of 18 diverse native Asian languages: Bahasa Indonesia, Cantonese, English, Mandarin, Japanese, Korean, Filipino, Javanese, Malay, Pakistani, Pashto, Pashtoo, Punjabi, Urdu, Pushto, Tagalog, Thai, and Uyghur as statistical data in Table~\ref{tab:stats_dia}.\footnote{For more details, see ~\url{https://language.sakura.ne.jp/icnale/}} This dataset offers comprehensive information about L2 English speakers with varied L1 backgrounds. Each file in ICNALE contains transcripts of a single L2 speaker's recorded responses on different discussion topics. Examples of these dialogue transcripts can be found in Appendix~\ref{sec:icanleexample}.
For this study, we selected seven linguistically divergent native languages from the dataset~\cite{philippy2023identifying}: Korean (kor), Mandarin (cmn), Japanese (jpn), Cantonese (yue), Thai (tha), Malay (msa), and Urdu (urd). 

\begin{table}[ht]
    \centering
    \tabcolsep=0.1cm
    \resizebox{0.48\textwidth}{!}{
    \begin{tabular}{llcc} 
    \toprule 
     Stats & Dialogues & Tokens &  Participants  \\
     \cmidrule(lr){1-1} \cmidrule(lr){2-2} \cmidrule(lr){3-3} \cmidrule(lr){4-4}
    \# ICANLE (Human)   & 4,250 & 1,600K & 425  \\
    \cmidrule(lr){1-1} \cmidrule(lr){2-2} \cmidrule(lr){3-3} \cmidrule(lr){4-4}
    \# LLM Generated &  2,600 &  1,344K &  NA \\
    \cmidrule(lr){1-1} \cmidrule(lr){2-2} \cmidrule(lr){3-3} \cmidrule(lr){4-4}
    \# Example Dialogue & 7 sets (one per each L1) & 10K &  NA  \\
    \bottomrule 
    \end{tabular}
    }
    \caption{Statistics of L2 Dialogue dataset, including human benchmarks, generated L2 dialogue datasets, and those used in prompting}\label{tab:stats_dia}
\end{table}
\subsection{Automated Annotation with GPT 4o}
The initial annotation by GPT 4o \cite{achiam2023gpt} with few-shot prompting, used four examples per linguistics feature. For \textit{Reference Word}, we selected four sentences from a dialogue, highlighting reference words (e.g., he, she, her) and presenting them in a few-shot format (detailed prompts in Appendix~\ref{sec:LLMannoations}).
Each dialogue in the dataset was analyzed using GPT 4o to identify and annotate the specified linguistic entities using a \textit{span-annotation} approach. The resulting annotations were stored in a structured format (JSON).

\subsection{Human Validation of LLMs Annotations}
To assess the quality of the automated annotations, three volunteer annotators who are proficient bi-lingual speakers and are all PhD students in NLP,
manually reviewed 15\% (randomly sampled) of the annotated dialogues. The annotators are required to make a binary judgement whether the span-annotation output is correct. 
This manual assessment found that the GPT 4o annotations had an accuracy of 84.1\%, suggesting that it is a viable approach for automatic annotation.

\subsection{Prompt Refinement}
Our manual validation revealed consistent errors in constructs like \emph{Noun-Verb Collocations}, where non-collocating tokens (e.g., a little bit \textit{trouble}) were incorrectly annotated by GPT-4o with unnecessary token \textit{trouble}. To address this, we refined the few-shot examples and improved the instructions and conducted a second human validation. As we see improved accuracy for these constructs, we adopted these updated prompts for all experiments.\footnote{Full prompts are published in ~\url{https://github.com/RenaGao/LLMPirorknowledge}} under the instructions folder.

\section{L2 Dialogue Generation}\label{sec:three-steppipeline}

\begin{figure}[t]
    \centering
    \includegraphics[width=0.80\linewidth]{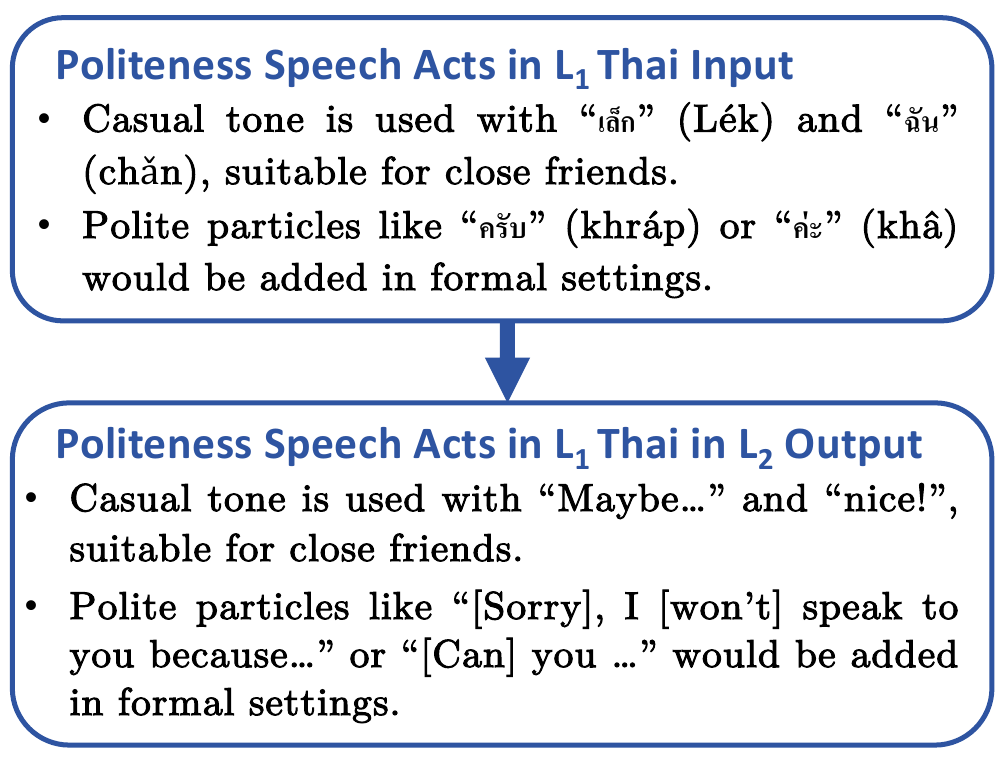}
    \caption{An example for Thai L1 knowledge injection prompting of Speech Acts, we provided full sentences in a complete dialogue context, the  utterances were omitted as ``...'' in this figure}
    \label{fig:thai-polite}
\end{figure}

To generate L2 dialogues using LLMs, we experiment with L1 knowledge injection through prompting: we design an instruction that contains high-level meta-linguistic information of the L1 language and examples of carefully crafted dialogue pairs that capture key dialogue grammatical traits~\cite{chen-2023-large,hu2022context}. 
Detailed instructions and sample L1 knowledge injection dialogue pairs are provided in Appendix~\ref{sec:in-context learning}. 
Each pair consists of at least 20 turns of conversation in L1, with corresponding English (L2) translation. These examples emphasize specific linguistic features, such as speech act politeness with Thai~\cite{srisuruk2011politeness} as shown in Figure~\ref{fig:thai-polite}. In addition to the L1 injection prompt, we also provide another set of instructions to generate the L2 dialogues.\footnote{Generation temperature is set to 0 in our experiments.} The LLM is instructed to ``role-play'' as an L2 English speaker, emulating realistic behaviors such as tense agreement and politeness strategies. For example, the model is prompted to act as an L2 speaker in an interview scenario, where the interviewer (an English native speaker) follows predefined templates based on ICNALE benchmark datasets. All prompts can be found in Appendix~\ref{sec:LLMGenprompts}. \footnote{These example conversations are derived from human L1 dialogues from xDial-Eval~\cite{zhang-etal-2023-xdial}, a multilingual open-domain dialogue dataset.}

\section{Results and Analysis}\label{sec:experiment} 
We conduct experiments using five large language models (LLMs) for L2 dialogue generation: \textbf{LLAMA3-8B, LLAMA3-70B} \textit{(April 2024)}, \textbf{Qwen2.5-72B} \textit{(September 2024)}, \textbf{DeepSeekV3-685B} \textit{(December 2024)}, and \textbf{GPT-4o} \textit{(December 2024)}.


\subsection{L1-Specification Impact across LLMs}\label{sec:staticalanalysis}

\begin{table*}[!t]
\centering
\resizebox{0.9\textwidth}{!}{%
\begin{tabular}{llllllllll}
\toprule
 & & \multicolumn{8}{c}{Distribution distance between humans' and LLMs' generated dialogues ($\downarrow$)} \\
 \cmidrule(lr){3-10}
\textbf{Lang.} & \textbf{Condition} & \textbf{\begin{tabular}[c]{@{}c@{}}Number \\ Agreement\end{tabular}} &
  \textbf{\begin{tabular}[c]{@{}c@{}}Tense \\ Agreement\end{tabular}} &
  \textbf{\begin{tabular}[c]{@{}c@{}}Subject-Verb \\ Agreement\end{tabular}} &
  \textbf{\begin{tabular}[c]{@{}c@{}}Modal Verbs \\ Expressions\end{tabular}} &
  \textbf{\begin{tabular}[c]{@{}c@{}}Quantifiers \\ Numerals\end{tabular}} &
  \textbf{\begin{tabular}[c]{@{}c@{}}Noun-Verb \\ Collocation\end{tabular}} &
  \textbf{\begin{tabular}[c]{@{}c@{}}Reference \\ Word\end{tabular}} &
  \textbf{\begin{tabular}[c]{@{}c@{}}Speech \\ Acts\end{tabular}} \\
  \cmidrule(lr){1-1} \cmidrule(lr){2-2} \cmidrule(lr){3-3} \cmidrule(lr){4-4} \cmidrule(lr){5-5} \cmidrule(lr){6-6} \cmidrule(lr){7-7} \cmidrule(lr){8-8} \cmidrule(lr){9-9} \cmidrule(lr){10-10}
\multirow{2}{*}{Cantonese} 
& $d_{\mathrm{bi}}$   
  & \inc{0.086} & \inc{0.021} & \inc{0.045} & \dec{0.313} & \dec{0.367} & \dec{0.073} & \inc{0.138} & \inc{0.373} \\
& $d_{\mathrm{mono}}$
  & 0.231       & 0.034       & 0.326       & 0.158       & 0.136       & 0.001       & 0.449       & 0.817       \\
  \cmidrule(lr){1-1} \cmidrule(lr){2-2} \cmidrule(lr){3-3} \cmidrule(lr){4-4} \cmidrule(lr){5-5} \cmidrule(lr){6-6} \cmidrule(lr){7-7} \cmidrule(lr){8-8} \cmidrule(lr){9-9} \cmidrule(lr){10-10}
\multirow{2}{*}{Thai}
& $d_{\mathrm{bi}}$
  & \dec{0.096} & \inc{0.072} & \inc{0.311} & \inc{0.064} & \dec{0.182} & \inc{0.050} & \inc{0.504} & \inc{0.561} \\
& $d_{\mathrm{mono}}$
  & 0.038       & 0.257       & 0.589       & 0.127       & 0.024       & 0.142       & 0.625       & 1.099       \\
  \cmidrule(lr){1-1} \cmidrule(lr){2-2} \cmidrule(lr){3-3} \cmidrule(lr){4-4} \cmidrule(lr){5-5} \cmidrule(lr){6-6} \cmidrule(lr){7-7} \cmidrule(lr){8-8} \cmidrule(lr){9-9} \cmidrule(lr){10-10}
\multirow{2}{*}{Japanese}
& $d_{\mathrm{bi}}$
  & \inc{0.010} & \inc{0.031} & \inc{0.160} & \inc{0.073} & \dec{0.104} & \inc{0.014} & \inc{0.183} & \inc{1.154} \\
& $d_{\mathrm{mono}}$
  & 0.037       & 0.305       & 0.685       & 0.321       & 0.090       & 0.185       & 0.695       & 1.894       \\
  \cmidrule(lr){1-1} \cmidrule(lr){2-2} \cmidrule(lr){3-3} \cmidrule(lr){4-4} \cmidrule(lr){5-5} \cmidrule(lr){6-6} \cmidrule(lr){7-7} \cmidrule(lr){8-8} \cmidrule(lr){9-9} \cmidrule(lr){10-10}
\multirow{2}{*}{Korean}
& $d_{\mathrm{bi}}$
  & \inc{0.043} & \inc{0.019} & \inc{0.136} & \inc{0.049} & \dec{0.094} & \inc{0.026} & \inc{0.321} & \inc{1.241} \\
& $d_{\mathrm{mono}}$
  & 0.103       & 0.035       & 0.394       & 0.173       & 0.017       & 0.144       & 0.542       & 2.268       \\
  \cmidrule(lr){1-1} \cmidrule(lr){2-2} \cmidrule(lr){3-3} \cmidrule(lr){4-4} \cmidrule(lr){5-5} \cmidrule(lr){6-6} \cmidrule(lr){7-7} \cmidrule(lr){8-8} \cmidrule(lr){9-9} \cmidrule(lr){10-10}
\multirow{2}{*}{Malay}
& $d_{\mathrm{bi}}$
  & \inc{0.069} & \inc{0.156} & \inc{0.062} & \inc{0.042} & \dec{0.113} & \inc{0.016} & \inc{0.164} & \inc{0.771} \\
& $d_{\mathrm{mono}}$
  & 0.109       & 0.167       & 0.369       & 0.082       & 0.022       & 0.031       & 0.438       & 1.184       \\
  \cmidrule(lr){1-1} \cmidrule(lr){2-2} \cmidrule(lr){3-3} \cmidrule(lr){4-4} \cmidrule(lr){5-5} \cmidrule(lr){6-6} \cmidrule(lr){7-7} \cmidrule(lr){8-8} \cmidrule(lr){9-9} \cmidrule(lr){10-10}
\multirow{2}{*}{Mandarin}
& $d_{\mathrm{bi}}$
  & \inc{0.030} & \inc{0.028} & \inc{0.133} & \inc{0.023} & \dec{0.070} & \inc{0.037} & \inc{0.261} & \inc{0.618} \\
& $d_{\mathrm{mono}}$
  & 0.099       & 0.208       & 0.455       & 0.109       & 0.028       & 0.091       & 0.530       & 1.175       \\
  \cmidrule(lr){1-1} \cmidrule(lr){2-2} \cmidrule(lr){3-3} \cmidrule(lr){4-4} \cmidrule(lr){5-5} \cmidrule(lr){6-6} \cmidrule(lr){7-7} \cmidrule(lr){8-8} \cmidrule(lr){9-9} \cmidrule(lr){10-10}
\multirow{2}{*}{Urdu}
& $d_{\mathrm{bi}}$
  & \inc{0.041} & \dec{0.133} & \inc{0.057} & \inc{0.091} & \dec{0.251} & \inc{0.010} & \inc{0.205} & \inc{0.311} \\
& $d_{\mathrm{mono}}$
  & 0.102       & 0.078       & 0.291       & 0.117       & 0.052       & 0.035       & 0.529       & 0.822       \\
\bottomrule
\end{tabular}
}
\caption{The distribution divergences $d_\mathrm{bi}$ and $d_\mathrm{mono}$ of GPT 4o generated L2 dialogues for different native languages: Korean (kor), Mandarin (cmn), Japanese (jpn), Cantonese (yue), Thai (tha), Malay (msa), and Urdu (urd) where \inc{green} indicates $d_{\mathrm{bi}}$ is less than $d_{\mathrm{mono}}$, while \dec{red} indicates the opposite.}\label{tab:gpt4o-distributions}
\end{table*}

\begin{figure*}[!t]
    \centering
    \begin{subfigure}[b]{0.275\textwidth}
        \includegraphics[width=\textwidth]{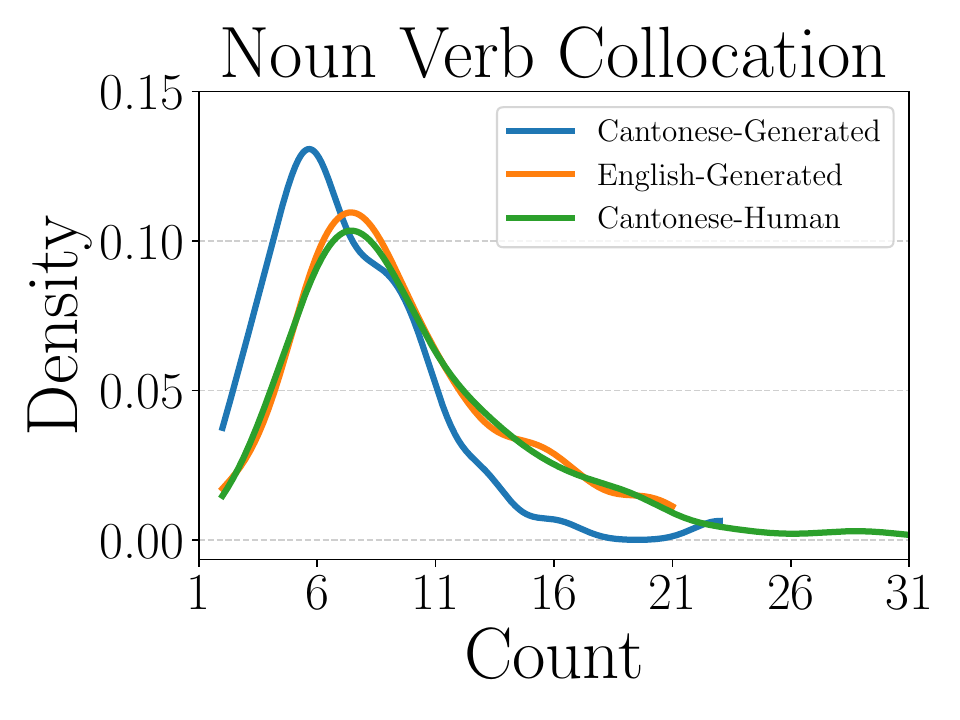}
        \caption{Cantonese-L1 with NVC}
        \label{fig:can_nvc}
    \end{subfigure}
    \hfill
    \begin{subfigure}[b]{0.275\textwidth}
        \includegraphics[width=\textwidth]{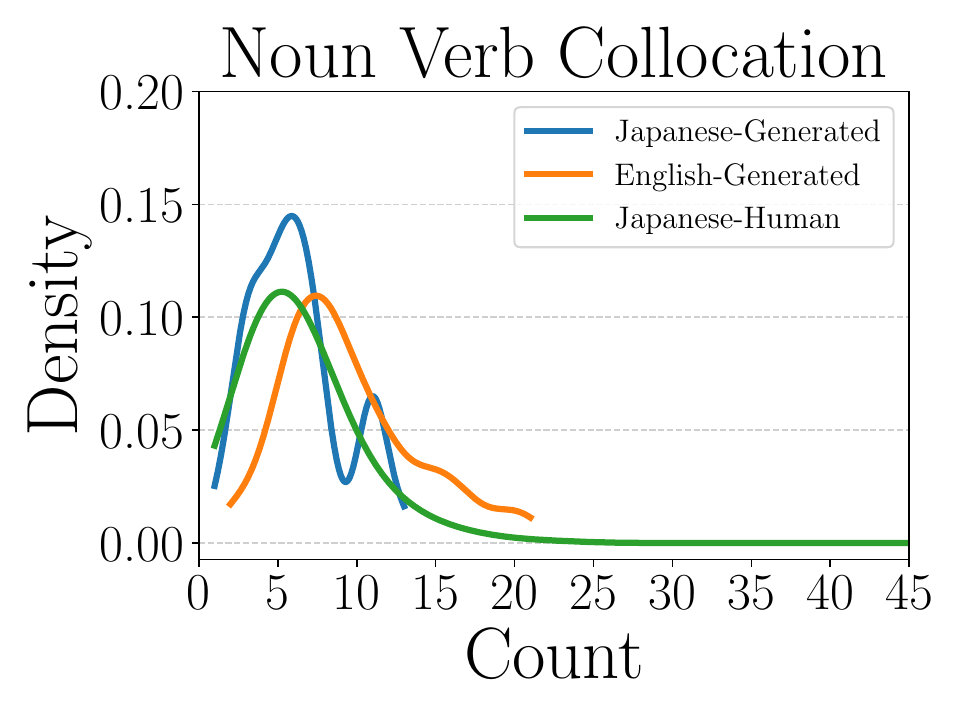}
        \caption{Japanese-L1 with NVC}
        \label{fig:jpn_nvc} 
    \end{subfigure}
    \hfill
    \begin{subfigure}[b]{0.275\textwidth}
        \includegraphics[width=\textwidth]{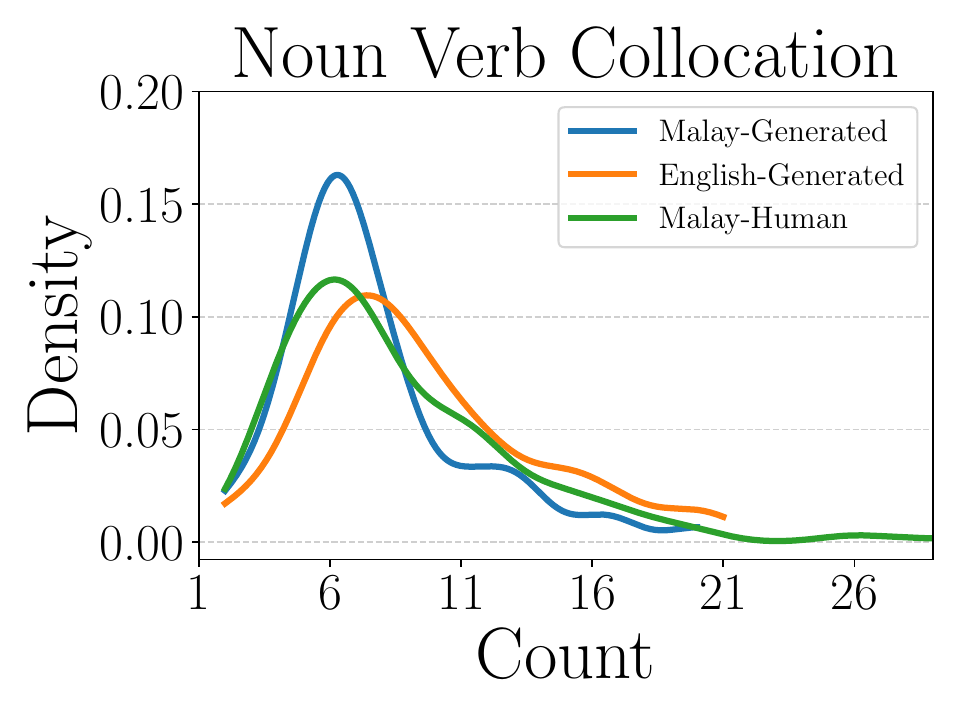}
        \caption{Malay-L1 with NVC}
        \label{fig:msa_nvc}
    \end{subfigure}
    \hfill \\
    \begin{subfigure}[b]{0.275\textwidth}
        \includegraphics[width=\textwidth]{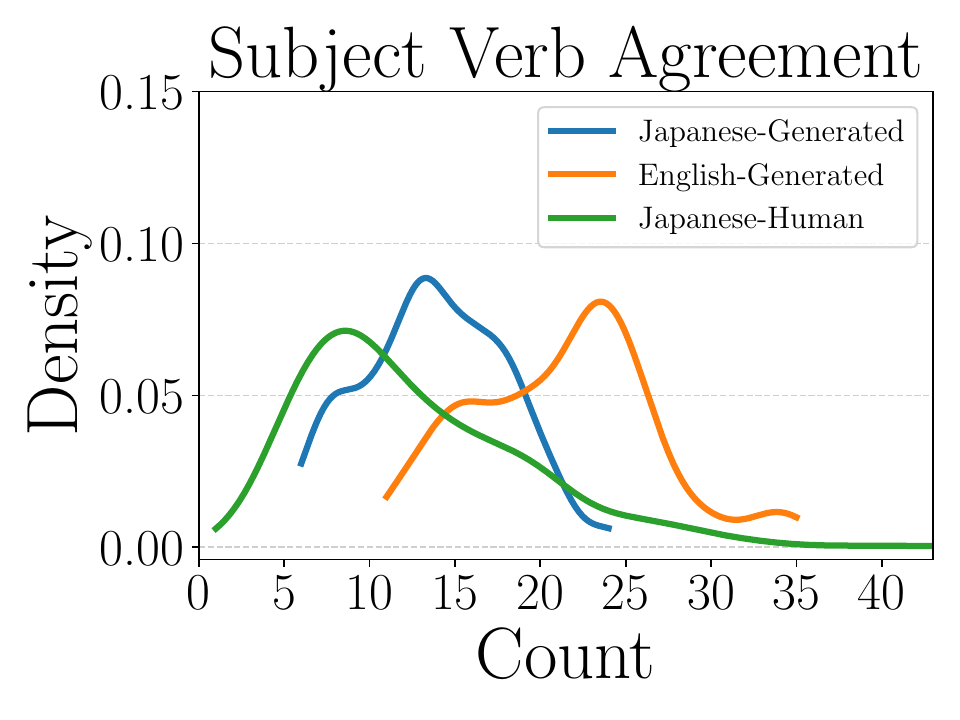}
        \caption{Japanese-L1 with SVA}
        \label{fig:jpn_sva}
    \end{subfigure}
    \hfill
    \begin{subfigure}[b]{0.275\textwidth}
        \includegraphics[width=\textwidth]{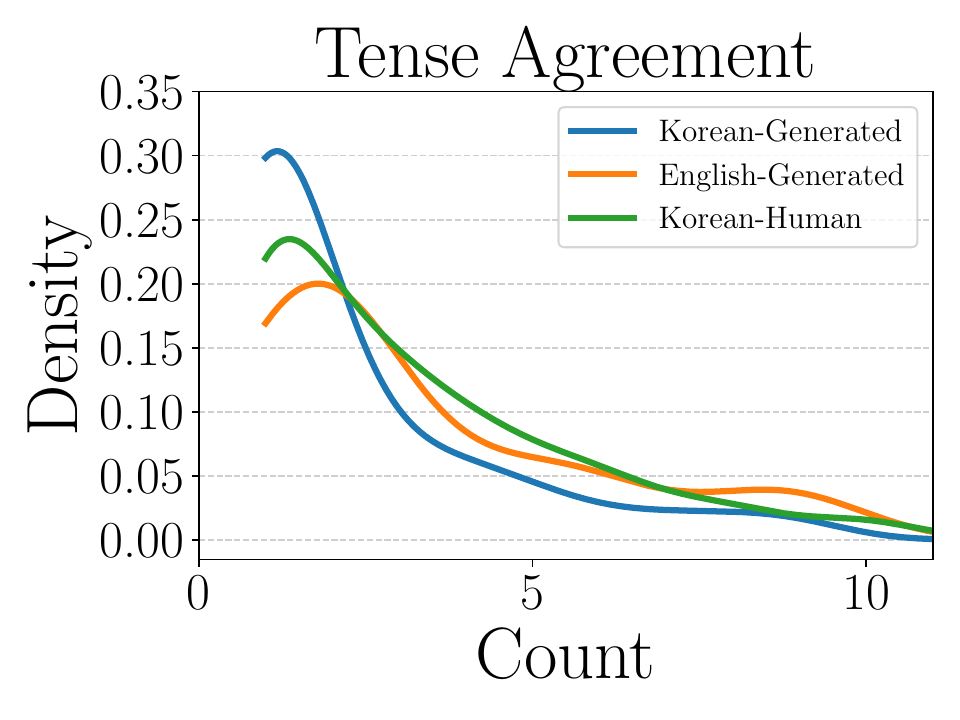}
        \caption{Korean-L1 with TA}
        \label{fig:jpn_na}
    \end{subfigure}
    \hfill
    \begin{subfigure}[b]{0.275\textwidth}
        \includegraphics[width=\textwidth]{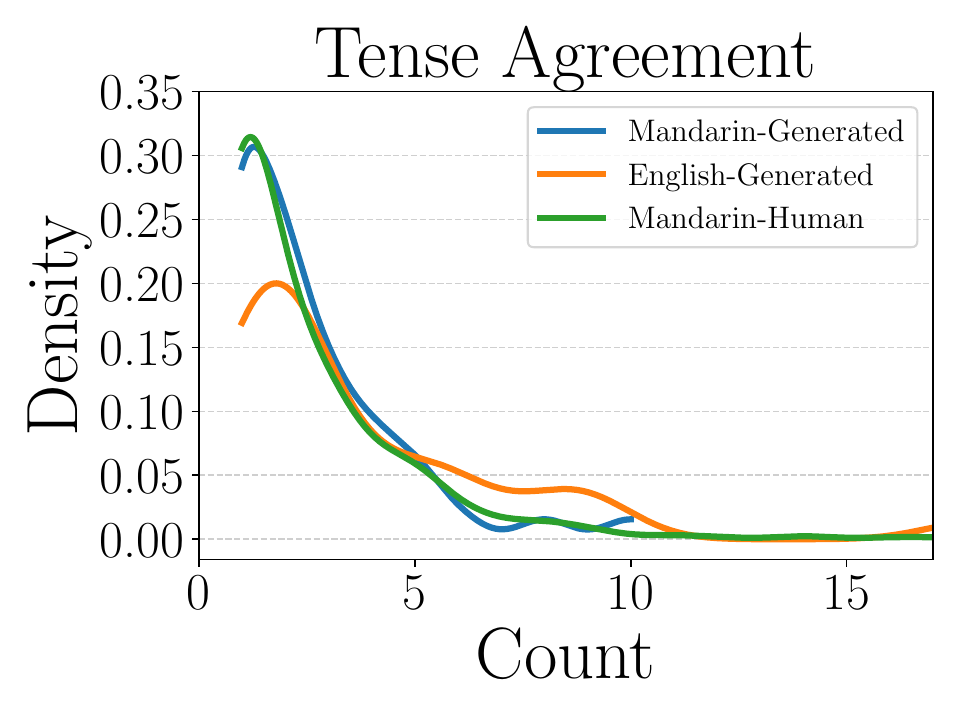}
        \caption{Mandarin-L1 with TA}
        \label{fig:jpn_ta}
    \end{subfigure}
    \caption{Density results for L2 GPT-4o generation dialogue via different L1s where NVC represents noun and verb collocations, TA for tense agreement and NA for number agreement. The blue lines (L2-Generated), orange lines (English-Generated), and green lines (L2-Humans) correspond to LLM-generated dialogue with L1 prompting, that without L1 knowledge injection prompting, and respective human dialogue.}
    \label{fig:L2density}
\end{figure*}
\begin{figure*}[!t]
    \centering
    \begin{subfigure}[b]{0.275\textwidth}
    \includegraphics[width=\textwidth]{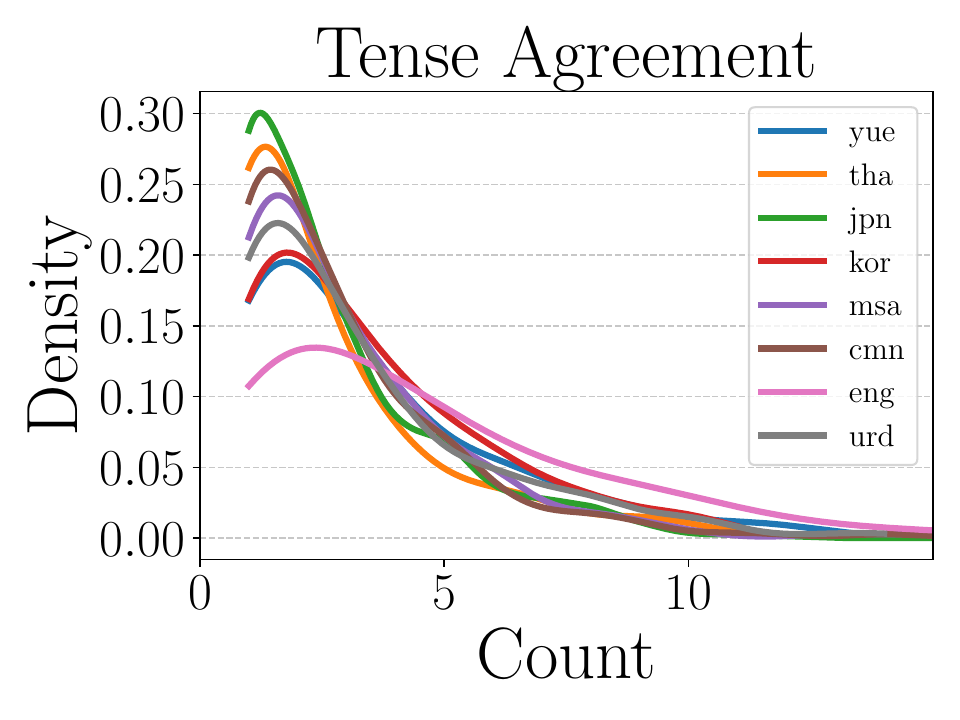}
        \caption{TA of Real L1}
    \label{fig:real_ta} 
    \end{subfigure}
    \hfill
    \begin{subfigure}[b]{0.275\textwidth}
    \includegraphics[width=\textwidth]{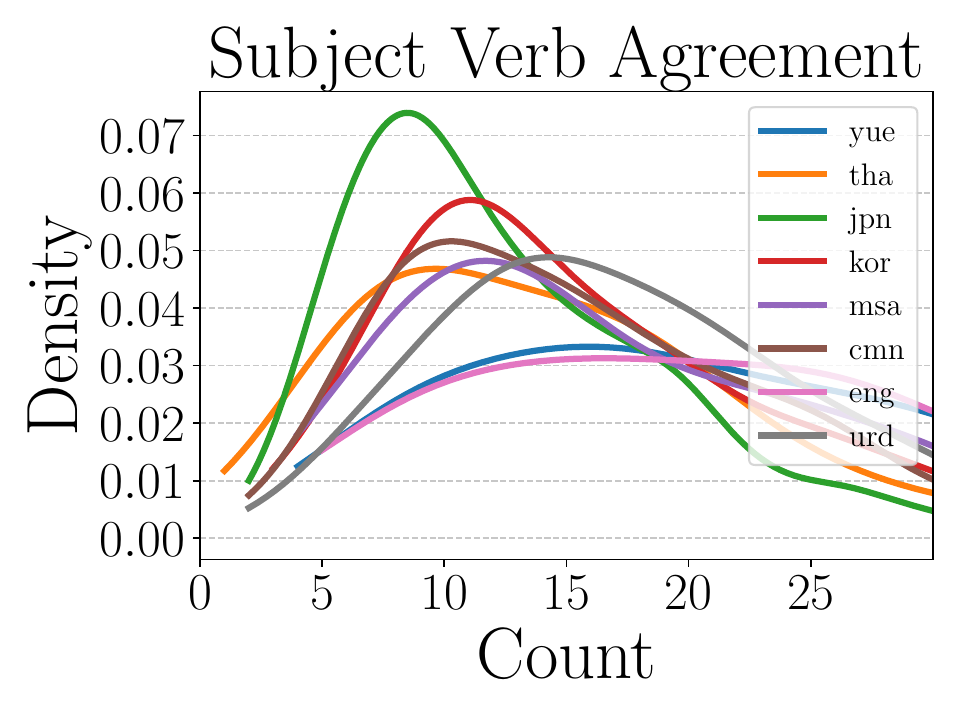}
        \caption{SVA of Real L1}
        \label{fig:real_sva}
    \end{subfigure}
    \hfill
    \begin{subfigure}[b]{0.275\textwidth}
    \includegraphics[width=\textwidth]{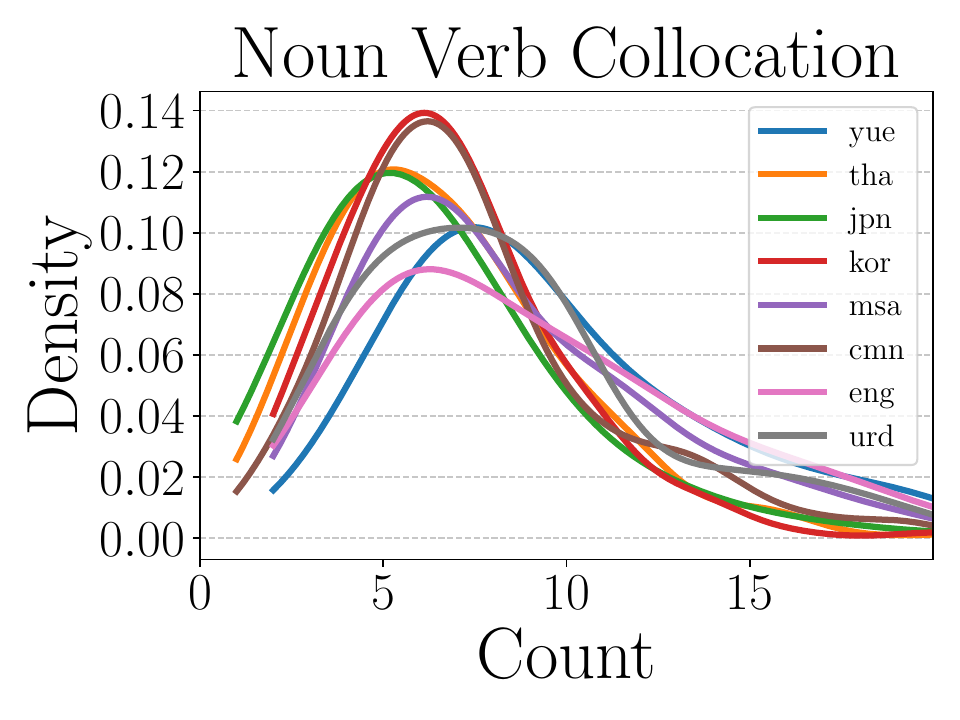}
        \caption{NVC of Real L1}
        \label{fig:real_nvc}
    \end{subfigure}    
    \caption{Density results of human-baseline dialogues of different L1s, where NVC represents \emph{Noun and Verb Collocations}, TA for \emph{Tense Agreement} and NA for \emph{Number Agreement}.} 
    \label{fig:real_density}
\end{figure*}

We present GPT-4o results in 
in Table~\ref{tab:gpt4o-distributions} (full results for all models are in Appendix~\ref{sec:allresults}). 
GPT-4o generated L2 dialogues exhibit generally consistent and significant improvements across all seven languages after prompting with the L1 information injection prompting, given the decrease (in distance) of $d_\mathrm{bi}$ from $d_\mathrm{mono}$, expect for \emph{Quantifiers Numerals}. This shows the effectiveness of promoting native linguistic information in L2-like dialogue generation. The eight grammatical constructs listed in Table~\ref{tab:linguistic_constructs} demonstrate human-like distribution patterns when leveraging native knowledge through L1 knowledge injection learning. This is particularly evident in the categories of Agreement--\emph{Tense Agreement, Number Agreement,} and \emph{Subject-Verb Agreement}, Pragmatics--\emph{Speech Acts}, and \emph{Reference Words}, which play important roles in oral communications~\cite{gao2024cnima}. Interestingly, GPT-4o struggles with \emph{Noun-Verb Collocation} for Cantonese speakers, likely due to the differences in word order between Cantonese and English~\cite{chan2010toward}.

Looking at the results across different LLMs (Appendix~\ref{sec:allresults}), the summary is that performance varies depending on the exact model, but broadly speaking L1 knowledge injection appears to help most models to mimic the L2 dialogue patterns, suggesting our approach is not LLM-dependent. Diving a bit deeper, DeepSeekV3 shows very strong performance similar to GPT-4o's, and
LLAMA3-8B performs the worst, although this is perhaps unsurprising since it's the smallest model. This does suggest, however, that model size is an important factor when it comes to the selection of LLM.
\begin{table}[!h]
\centering
\resizebox{0.48\textwidth}{!}{%
\begin{tabular}{llp{1.5cm}p{1.5cm}p{1.5cm}p{1.5cm}p{1.5cm}}
\toprule
 & & \multicolumn{5}{c}{Human--LLM dialogue distribution distance ($\downarrow$)} \\
\cmidrule(lr){3-7}
\textbf{Lang.} & \textbf{Cond.} & \textbf{DeepSeek V3} & \textbf{QWEN 72B} & \textbf{LLaMA 70B} & \textbf{LLaMA 8B} & \textbf{GPT 4o} \\
  \cmidrule(lr){1-1} \cmidrule(lr){2-2} \cmidrule(lr){3-3} \cmidrule(lr){4-4} \cmidrule(lr){5-5} \cmidrule(lr){6-6} \cmidrule(lr){7-7}
\multirow{2}{*}{Cantonese}
& $d_{\mathrm{bi}}$   & \dec{0.102} & \dec{0.137} & \dec{0.376} & \dec{0.294} & \dec{0.367} \\
& $d_{\mathrm{mono}}$ & 0.051       & 0.026       & 0.046       & 0.086       & 0.136       \\
  \cmidrule(lr){1-1} \cmidrule(lr){2-2} \cmidrule(lr){3-3} \cmidrule(lr){4-4} \cmidrule(lr){5-5} \cmidrule(lr){6-6} \cmidrule(lr){7-7}
\multirow{2}{*}{Thai}
& $d_{\mathrm{bi}}$   & \dec{0.057} & \inc{0.061} & \dec{0.093} & \dec{0.064} & \dec{0.182} \\
& $d_{\mathrm{mono}}$ & 0.037       & 0.098       & 0.025       & 0.019       & 0.024       \\
  \cmidrule(lr){1-1} \cmidrule(lr){2-2} \cmidrule(lr){3-3} \cmidrule(lr){4-4} \cmidrule(lr){5-5} \cmidrule(lr){6-6} \cmidrule(lr){7-7}
\multirow{2}{*}{Japanese}
& $d_{\mathrm{bi}}$   & \inc{0.060} & \inc{0.094} & \inc{0.060} & \inc{0.025} & \dec{0.104} \\
& $d_{\mathrm{mono}}$ & 0.181       & 0.286       & 0.156       & 0.113       & 0.090       \\
  \cmidrule(lr){1-1} \cmidrule(lr){2-2} \cmidrule(lr){3-3} \cmidrule(lr){4-4} \cmidrule(lr){5-5} \cmidrule(lr){6-6} \cmidrule(lr){7-7}
\multirow{2}{*}{Korean}
& $d_{\mathrm{bi}}$   & \inc{0.021} & \inc{0.017} & \dec{0.262} & \dec{0.085} & \dec{0.094} \\
& $d_{\mathrm{mono}}$ & 0.070       & 0.117       & 0.031       & 0.004       & 0.017       \\
  \cmidrule(lr){1-1} \cmidrule(lr){2-2} \cmidrule(lr){3-3} \cmidrule(lr){4-4} \cmidrule(lr){5-5} \cmidrule(lr){6-6} \cmidrule(lr){7-7}
\multirow{2}{*}{Malay}
& $d_{\mathrm{bi}}$   & \dec{0.078} & \inc{0.060} & \dec{0.174} & \dec{0.149} & \dec{0.113} \\
& $d_{\mathrm{mono}}$ & 0.034       & 0.103       & 0.017       & 0.015       & 0.022       \\
  \cmidrule(lr){1-1} \cmidrule(lr){2-2} \cmidrule(lr){3-3} \cmidrule(lr){4-4} \cmidrule(lr){5-5} \cmidrule(lr){6-6} \cmidrule(lr){7-7}
\multirow{2}{*}{Mandarin}
& $d_{\mathrm{bi}}$   & \dec{0.108} & \inc{0.022} & \dec{0.103} & \dec{0.085} & \dec{0.070} \\
& $d_{\mathrm{mono}}$ & 0.037       & 0.064       & 0.008       & 0.010       & 0.028       \\
  \cmidrule(lr){1-1} \cmidrule(lr){2-2} \cmidrule(lr){3-3} \cmidrule(lr){4-4} \cmidrule(lr){5-5} \cmidrule(lr){6-6} \cmidrule(lr){7-7}
\multirow{2}{*}{Urdu}
& $d_{\mathrm{bi}}$   & \dec{0.072} & \dec{0.140} & \dec{0.313} & \dec{0.054} & \dec{0.251} \\
& $d_{\mathrm{mono}}$ & 0.071       & 0.075       & 0.025       & 0.004       & 0.052       \\
\bottomrule
\end{tabular}
}
\caption{The distribution divergences $d_\mathrm{bi}$ and $d_\mathrm{mono}$ with different models for \emph{Quantifier Numerals}.} \label{tab:quantifier_res}
\end{table}

Another consistent result across all LLMs is the poor performance for \emph{Quantifiers and Numerals}; see Table~\ref{tab:quantifier_res}.
A likely explanation is that many LLMs prioritize natural, concise, conversational English, where quantifiers and numerals are often omitted when the meaning is clear (e.g., ``We bought apples'' instead of ``We bought \textit{some} apples''). Qwen2.5-72B appears to do better than other LLMs, and it is perhaps due to its stronger exposure to Mandarin~\cite{yang2024qwen2} where quantifiers and numerals are more structurally important. This linguistic influence likely helps the model retain explicit quantifiers and numerals, even in contexts where native English speakers might naturally omit them during conversation.

\subsection{L2 Generation Power via L1 Distance} \label{sec:LLML2}

The density comparison results in Figure~\ref{fig:L2density} show a consistent yet subtle influence of a speaker's L1 on GPT-4o's performance in generating L2 English dialogues. We focus on GPT-4o here due to its strong performance in mimicking L1 language patterns (Appendix~\ref{fig:L2density} presents density results for all LLMs).

For speakers of Cantonese, Japanese, and Malay (Figure~\ref{fig:can_nvc}, \ref{fig:jpn_nvc}, \ref{fig:msa_nvc}) --- languages that share certain structural similarities with English such as \emph{Noun and Verb Collocations} --- the generated dialogues generally resemble human-like patterns (except for the Cantonese with NVC). This alignment is supported by the L1 distance density results, suggesting that in most cases, LLMs successfully transfer these L1 linguistic features into the L2 English dialogues, particularly when grammatical agreement plays a key role in conveying meaning and maintaining semantic clarity.

To further investigate the impact of L1, we put together density results for different L1s given a construct in Figure~\ref{fig:real_density}. The \emph{eng} line (pink) serves as the baseline, representing native English speakers. We find that L1s with more distant grammatical structures from English --- such as those with an `SOV' (Subject-Object-Verb) word order --- tend to induce greater deviations in the generated dialogues. For instance, Japanese and Korean, which follow an SOV structure, exhibit a greater divergence from the English baseline in \emph{Subject-Verb Agreement} (Figure~\ref{fig:real_sva}). 
This is also reflected by the decrease from $d_\mathrm{mono}$ to $d_\mathrm{bi}$ in Table~\ref{tab:gpt4o-distributions}. In the case of \emph{Tense Agreement} (Figure~\ref{fig:real_ta}), all L1s show relatively similar distributions, as observed in the density patterns for Korean and Mandarin LLM-generated dialogues, despite their typological differences. 



\subsection{Qualitative Analysis: LLM L2 Human-like Dialogue Generation} 
Beyond the statistical analysis, we qualitatively examined 30 LLM-generated L2 dialogues per language across five models, providing an in-depth analysis of L1-specific patterns and LLM-L2 generated traits where the models' generation diverged from target-language norms in ways that go beyond typical L2 transfer patterns. Below, we summarize key features observed: DeepseekV3 reproduces the characteristic omission errors of numerals and quantifiers found among some Asian native L2 learners. An example for Mandarin L1 is: ``We make some food. Like... sandwich, fruit... how to say... drink?" This kind of omission often arises from L1 transfer, where speakers rely on L1 structures or habitual omission of certain function words~\cite{macuch2024strategic}. GPT 4o exhibits word order, agreement, and collocation traits often observed in Urdu-speaking learners following the Subject–object–verb (SOV) structure~\cite{saleem2021social}. For instance, ``Lahore University I study'' and ``It has been about three year now.'' Additionally, when generating outputs for Thai speakers, GPT 4o captures instances of politeness mismatches by occasionally omitting formal markers, as shown in ``Yes, a lot of plant. Many flower, very beautiful.'' These intriguing patterns may reflect how learners from such L1 backgrounds may transfer default word ordering or honorific usage from their native language to L2 contexts~\cite{chansamrong2014effectiveness}. LLAMA3-70B replicates the common challenge with speech acts and modal verbs seen among Korean L1 learners, producing examples like ``What time you think is good to go?'' This often occurs because of essential differences in how Korean grammar encodes modalities compared to target languages~\cite{mott2024thing}. Qwen2.5-72B demonstrates pragmatic choices that closely resemble human dialogue, though it occasionally displays unusual modal expressions across languages, such as ``Yes, I try. But my picture not very good.'' for Mandarin L1. These reflections of L1-driven structures and lexical choices show how underlying knowledge of a first language can shape second language productions. These interesting influences reflect the extent to \textit{which LLMs} internalize L1-specific structures and how to apply them to L2 production, even when the target language's norms differ.

\section{Conclusions} 
This study introduces an automated dialogue annotation framework and an information-theoretic method to evaluate LLMs' performance in simulating L2 English dialogues with L1 influence. Using the ICNALE dataset, we compared LLM outputs with human data, showing that LLMs can capture L1-specific patterns through L1 knowledge injection. The results show strong alignment with human speakers in dialogue cohesion, grammar, and pragmatic use, offering insights to improve multilingual dialogue systems for educational applications.

\section*{Limitations}\label{sec:limitations} 
This study has several limitations. First, it relies on the ICNALE dataset as only benchmark, which may limit the generalizability of the results to languages beyond Asian languages. Second, the use of predefined templates for few-shot prompting ensures consistency but may constrain the analysis of \textit{spontaneous L2 language behaviors}, such as chit-chat. Furthermore, the study focuses on linguistics features, overlooking the potential impact of socio-cultural bias on each native language use. Future work should address these limitations by incorporating more diverse datasets and examining unscripted interactions to enhance the validity and applicability of the results.

\section*{Ethics Statement}
This study is conducted under the guidance of the ACL Code of Ethics. The volunteer annotators were all NLP PhD students who are willing to participate in manual checking for this study. We removed all information related to the identification of human volunteer annotators. This study was approved by the The University of Melbourne ethics board (Human Ethics Committee LNR 1D), Reference Number 2022-24988-32929-3, and data acquisition and analysis has been taken out to the according ethical standards., and data acquisition and analysis has been taken out to the according ethical standards.

\bibliography{custom}
\newpage
\onecolumn
\appendix

\section{Appendix} \label{sec:appendix}
\subsection{High-Level Instructions and L1 Injection Prompts} \label{sec:in-context learning}

\subsubsection{General Prompts}

Depending on the language, we design explicit L1 knowledge injection learning examples adopted from L2 human data and based on the grammatical traits in expression from each native language 
\begin{tcolorbox}[myparagraph]
Your goal is to generate a realistic conversation in English between one \{target language\} native speaker and a native English speaker.

Read and learn the provided \{target language\} dialogue and the analysis of grammatical traits. 

Scene [Optional]: Two friends, \{speaker 1\} and \{speaker 2\}, are planning to visit the mall over the weekend and discuss what to do there.
\end{tcolorbox}

\begin{tcolorbox}[incontextexample]
\textbf{In this section, we only show a piece of L1 knowledge injection example prompts  for different L1s. For more examples from full dialogues, please refer to the context instructions folder in:~\url{https://github.com/RenaGao/LLMPirorknowledge}}\\

\textbf{Scene:} Two friends, \{speaker A\} and \{speaker B\}, are meeting at a \{certain place\} for \{some discussions\}. \textbf{Note that this is a template for different example prompt depending on the scene and the contents \{...\}} are put here as placeholders.\\
------------ \\
\textbf{Malay Example}\\
\textbf{Aiman:} Farah, awak ada rancangan hujung minggu ni?\\
\textit{(Farah, awak ada rancangan hujung minggu ni?)}\\
``Farah, do you have any plans this weekend?''\\
\textbf{Farah:} Tak ada apa-apa pun. Kenapa? \\
\textit{(Tak ada apa-apa pun. Kenapa?)} \\
“No, nothing at all. Why?”\\
------------ \\
\textbf{Urdu Example}\\
\textbf{Ayesha:} \mychar{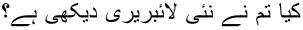} \\ 
\textit{(Kya tum ne nayi library dekhi hai?)} \\
“Have you seen the new library?” \\
\textbf{Bilal:} \mychar{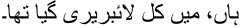} \\
\textit{(Haan, main kal library gaya tha.)} \\
“Yes, I went to the library yesterday.” \\
------------ \\
\textbf{Japanese Example}\\
\textbf{Sora:} \begin{CJK*}{UTF8}{min}こんにちは、明日何をする予定ですか？\end{CJK*} \\
\textit{(Konnichiwa, ashita nani o suru yotei desu ka?)} \\
“Hello, what are your plans for tomorrow?” \\
\textbf{Aki:} \begin{CJK*}{UTF8}{min} 明日は特に予定がありませんが、どうしてですか\end{CJK*} \\
\textit{(Ashita wa toku ni yotei ga arimasen ga, doushite desu ka?)} \\ 
``I don't have any particular plans for tomorrow. Why do you ask?'' \\
------------ \\
\textbf{Korean Example}\\
\textbf{Minji:}  지수야, 이번 주말에 시간 있어? \\
\textit{(Jisoo-ya, ibeon jumal-e sigan isseo?)} \\
``Jisoo, do you have time this weekend?” \\
\textbf{Jisoo}: 응, 있어. 왜? \\
\textit{(Eung, isseo. Wae?)} \\ 
“Yes, I do. Why?” \\
------------ \\
\textbf{Thai Example}\\
\textbf{Nuch}: \mychar{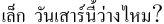}  \\ 
\textit{(Lék wan sǎo níi wâang mái?)} \\ 
“Lek, are you free this Saturday?” \\
\textbf{Lek}: \mychar{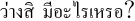}\\
\textit{(Wâang sì. Mii à-rai rǒe?)} \\
“I free. What’s up?” \\
------------ \\
\textbf{Mandarin Example}\\
\textbf{Xiao Ming:} \begin{CJK*}{UTF8}{bsmi}我想去公圜玩儿，最近天气很好。\end{CJK*} \\
\textit{(Wǒ xiǎng qù gōng yuán wánr, zuì jìn tiān qì hěn hǎo.)} \\
“I want to go to the park; the weather has been great recently.” \\
\textbf{Xiao Li:} \begin{CJK*}{UTF8}{bsmi}好主意！你想做什么？\end{CJK*} \\
\textit{(Hǎo zhǔ yì! Nǐ xiǎng zuò shén me?)} \\ 
“Good idea! What do you want to do?” \\
------------ \\
\textbf{Cantonese Example}\\
\textbf{Mei:} \begin{CJK*}{UTF8}{bsmi} 喂，阿Wing，星期六有冇時間呀？\end{CJK*}\\
\textit{(Wai, a Wing, sing1 kei4 luk6 jau5 mou5 si4 gaan3 aa3?)}\\
“Hey, Wing, do you have time on Saturday?”\\
\textbf{Wing:} \begin{CJK*}{UTF8}{bsmi}有呀，你想做啲咩呀？\end{CJK*}\\
\textit{(Jau5 aa3, nei5 soeng2 zou6 di1 me1 aa3?)}\\
“Yes, what do you want to do?” 
\end{tcolorbox}

\begin{tcolorbox}[traitanalysis]
Make sure to follow the following idiomatic expressions and cultural nuances commonly used by \{target language\} speakers. Keep the tone respectful and in line with traditional \{target language\} communication styles. \textbf{Here we give Malay as an example while we do have specific trait analysis prompts for other languages.}
\begin{enumerate}
    \item \textbf{Particles}
    \begin{itemize}
        \item ``pun'': Used for emphasis, e.g., ``Tak ada apa-apa pun.'' (Nothing at all).
        \item ``ke'': Indicates direction, e.g., ``pergi ke pusat membeli-belah'' (go to the mall).
    \end{itemize}

    \item \textbf{Aspect Markers}
    \begin{itemize}
        \item ``nak'': Informal future marker, e.g., ``Saya nak pergi'' (I want to go).
        \item ``dengar'': Implied past aspect in ``saya dengar food court dia besar'' (I heard their food court is big).
    \end{itemize}

    \item \textbf{Topic-Comment Structure}
    \begin{itemize}
        \item ``Wayang apa yang awak nak tengok?'' (What movie do you want to watch?): Topic ``Wayang apa'' introduces the subject, and ``awak nak tengok'' comments on it.
    \end{itemize}
    \item \textbf{Politeness Levels}
    \begin{itemize}
        \item Formal tone with ``saya'' (I) and ``awak'' (you) is polite but casual, suitable for friendly conversations.
        \item Politeness can be enhanced with ``Encik'' or ``Cik'' for formal contexts.
    \end{itemize}

    \item \textbf{Verb Serialization}
    \begin{itemize}
        \item ``Makan tengah hari di sana. Lepas tu, nak tengok wayang?'' (Have lunch there. After that, shall we watch a movie?): Actions are listed sequentially.
    \end{itemize}

    \item \textbf{Conjunctions}
    \begin{itemize}
        \item ``dan'': Connects clauses, e.g., ``banyak kedai baru, dan saya dengar'' (many new shops, and I heard).
        \item ``Lepas tu'': Informal for ``after that.''
    \end{itemize}

    \item \textbf{Time Expressions}
    \begin{itemize}
        \item ``hujung minggu ni'' (this weekend).
        \item ``pukul 10 pagi'' (10 a.m.).
    \end{itemize}

    \item \textbf{Expressions of Agreement}
    \begin{itemize}
        \item ``Setuju!'' (Agreed!).
        \item ``Boleh!'' (Sure!).
    \end{itemize}

    \item \textbf{Conditional Suggestions}
    \begin{itemize}
        \item ``Kita tengok jadual wayang nanti.'' (Let's check the movie schedule later): Indicates a planned action.
    \end{itemize}

    \item \textbf{Adjectives for Excitement}
    \begin{itemize}
        \item ``Bagus tu!'' (That's great!) expresses enthusiasm.
    \end{itemize}
\end{enumerate}
\end{tcolorbox}

\subsection{L2 Dialogue Generation Prompts} \label{sec:LLMGenprompts}
\begin{tcolorbox}[myparagraph]
Given the topic: {text}. Generate a realistic conversation IN ENGLISH with 20 turns between two native Cantonese speakers. Make sure the output is not cut off. Provide the complete English conversation below. 
\begin{enumerate}
    \item \textbf{Speaker A (Native Speaker, NS)}
    \begin{itemize}
        \item Fluent and natural English speaker with clear, concise, and polite phrasing.
        \item Provides guidance, asks questions, and may clarify misunderstandings when necessary.
        \item Avoids overly complex words or idioms to make the conversation accessible for L2 learners.
    \end{itemize}
    \item \textbf{Speaker B (Second-Language Speaker)}
    \begin{itemize}
        \item A non-native English speaker whose proficiency reflects an intermediate-to-upper-intermediate level.
        \item Their native language is \{language\}, please follow the idiomatic expressions and cultural nuances commonly used by \{language\} speakers.
        \item Exhibits typical linguistic influences from their native language, such as:
        \begin{itemize}
            \item Grammatical mistakes (e.g., ``He have'' instead of ``He has'').
            \item Limited vocabulary leading to overuse of simple words or circumlocution (e.g., ``thing for fixing paper'' instead of ``stapler'').
            \item Pronunciation hints if relevant.
            \item Uses filler phrases or pauses to reflect real-time language processing (e.g., ``Um'', ``How to say...'').
        \end{itemize}
    \end{itemize}
    \item \textbf{Context}: The conversation is around for some topics or scenes. The L2 speaker is trying to express their thoughts, answer questions, or solve a problem, while the native speaker responds supportively to maintain the flow of the conversation.
    \item \textbf{Requirements}
    \begin{itemize}
        \item \textbf{Cultural Nuances}: Reflect the L2 speaker's cultural communication style.
        \item \textbf{Balanced Exchange}: Ensure the dialogue alternates between the two speakers.
        \item \textbf{Error Patterns}: Highlight realistic mistakes in the L2 speaker's grammar, vocabulary, or syntax. Include occasional self-corrections or clarifications prompted by the native speaker.
        \item \textbf{Clarity and Empathy}: The native speaker provides clear, friendly responses, avoiding judgment of language mistakes.
        \item \textbf{Length and Focus}: The conversation should be concise, focusing on the L2 speaker's ability to express their ideas despite language barriers.
    \end{itemize}
\end{enumerate}
\end{tcolorbox}

\begin{tcolorbox}[incontextexample]
\textbf{Speaker A (NS):} Hi! Thanks for meeting with me today. Can you tell me a little about yourself?

\textbf{Speaker B (L2):} Um, yes. My name is Mei. I am from Hong Kong. I, uh... work in marketing for... four years.

\textbf{Speaker A (NS):} That’s great! What kind of marketing work do you do?

\textbf{Speaker B (L2):} I do, um, online... how to say... advertisement? On social media, and also write article.

\textbf{Speaker A (NS):} Oh, social media advertising and content writing?

\textbf{Speaker B (L2):} Yes, yes! Content writing. Sometimes for product launch, or... uh, promotion.

\textbf{Speaker A (NS):} I see. Do you enjoy writing for different audiences?

\textbf{Speaker B (L2):} Yes, very much. But, um... sometime hard because need many idea. Creative, you know?

\textbf{Speaker A (NS):} Absolutely, coming up with fresh ideas can be challenging. How do you find inspiration?

\textbf{Speaker B (L2):} I... ah, read other, um, campaign? And look what people like. Sometimes ask my teammate.

\textbf{Speaker A (NS):} That’s a smart approach! Collaboration always helps. What’s a campaign you’re particularly proud of?

\textbf{Speaker B (L2):} Oh, um, last year I make one for new phone. We use... uh, storytelling to show family connect. Many people like.

\textbf{Speaker A (NS):} Storytelling is very effective. How did you measure its success?

\textbf{Speaker B (L2):} We see, uh, number of share on Facebook and, um... how to say... comment? And we also check sale data.
\end{tcolorbox}

\subsection{L2 Annotation Prompts}  \label{sec:LLMannoations}
\begin{tcolorbox}[annotations]
\begin{itemize}
    \item \textit{You are a linguist expert specializing in doing text annotation in the English second language. You will be tasked with making annotations to a given dialogue texts based on some linguistics aspects to compare grammatical features in machine learning models for cross-lingual tasks.}
    \item The given text are samples in the dialogue passage from second language speakers of English.
    \item Make sure to keep the annotation format without any change in passage when giving the annotation output.
    \item A task may ask for one or multiple annotations. Each annotation should be an object with 5 fields:\begin{itemize}
        \item type: the type of annotation
        \item annotation sentence: the annotated sentence
        \item annotation token: the annotated tokens
        \item rationale: the reason why you give the annotation
        \item grammar correctness: the annotated grammar feature is aligned with the native English speaker's grammar usage
    \end{itemize}
\item Please return a json object which consists of one or multiple modifications.
\end{itemize}
\end{tcolorbox}

\subsection{Examples from the ICNALE Datasets} \label{sec:icanleexample}

\begin{tcolorbox}[mydialogue]
\begin{itemize}
    \item Uh, I think a 100 points is a full points maybe.  I think that I have - I maybe have 70 or 75 points.
    \item No, I - no.
    \item Um, I think this, uh, starting a new - a new thing I think, this will take a little time, uh, maybe for a month, 2 months, or maybe half an year, but finally you - you will not feel nervous about this.
    \item Because, uh, when - when you start something, people always - all of the people will feel nervous I think.
    \item So, just develop your English speaking skills and you will feel confident about.
    \item I think part-time jobs because I - I am now a student and I - I have no part-time job experience.
    \item Uh, I think role play.  This - uh, because I can have some communications with the teachers and this - this picture - storytelling, I think this is a bit - a little familiar with the TOEFL test, uh, speaking test.
\end{itemize}
\end{tcolorbox}

\subsection{LLM Generated L2 Dialogue Examples}\label{sec:l26.3examples}

\subsubsection{English Example} \label{tab:example_dialogue_english} 

\begin{tcolorbox}[mydialogue] 
        \textbf{Speaker A (NS):} Hey Mike, got any plans for the weekend? I was thinking we could do something fun. \\
        \textbf{Speaker B (NS):} Not really. I was just gonna stay home and binge some Netflix. Why, what’s up? Do you have something in mind? \\
        \textbf{Speaker A (NS):} Yeah, how about a quick road trip? We haven’t done that in ages! I’ve been itching to get out of the city. \\    
        \textbf{Speaker B (NS):} Hmm, that sounds cool. Where were you thinking? Someplace nearby, I hope. Gas prices are crazy. \\
        \textbf{Speaker A (NS):} True! I was thinking we could hit the beach. Maybe that little spot down by Clearwater? It’s not too far, and I hear the weather’s supposed to be great. \\
        \textbf{Speaker B (NS):} Oh, Clearwater? That could work. What would we do there, though? Just lounge around or do something active? \\
        \textbf{Speaker A (NS):} A bit of both, maybe? We could rent paddleboards or bikes. Then just chill on the sand after. I mean, it’s the beach—pretty hard to get bored. \\
        \textbf{Speaker B (NS):} I like the sound of that. Should we leave Saturday morning? Beats sitting around all weekend. \\
        \textbf{Speaker A (NS):} Yeah, bright and early! Say 8 a.m.? I’ll bring snacks if you handle the playlist. \\
        \textbf{Speaker B (NS):} Deal. But no weird indie stuff this time, okay? I’m still scarred from that experimental jazz phase you went through. \\
        \textbf{Speaker A (NS):} Ha! No promises, but I’ll try to keep it mainstream. By the way, do you have a cooler? We’ll need one for drinks. \\
        \textbf{Speaker B (NS):} Yeah, I’ve got one. What about parking? Isn’t it a nightmare there? \\
        \textbf{Speaker A (NS):} I think there’s a lot near the main boardwalk. We’ll just have to get there early to snag a spot. Another reason to leave on time! \\
        \textbf{Speaker B (NS):} Gotcha. Alright, sounds like a plan. Let’s touch base Friday night to confirm everything. \\
        \textbf{Speaker A (NS):} Sure thing. This is gonna be awesome! See you Friday! \\
        \textbf{Speaker B (NS):} Later, Sarah! 
    \end{tcolorbox}

\subsubsection{Thai Example}
\begin{tcolorbox}[mydialogue] 
\textbf{Speaker A (NS):} Hello! Is this Nuch? \\
\textbf{Speaker B (L2, Thai):} Yes, yes, it is Nuch. Who is speaking, please? \\
\textbf{Speaker A (NS):} It’s Lek, your cousin! How have you been doing? \\
\textbf{Speaker B (L2, Thai):} Oh! Lek! I am good, thank you. How about you? \\
\textbf{Speaker A (NS):} I’m doing well, thanks. It’s been a long time since we talked. How is everything with the family? \\ \textbf{Speaker B (L2, Thai):} Family is very okay. My mom, she has new hobby... um, plant... planting tree. \\
\textbf{Speaker A (NS):} Oh, gardening! That sounds nice. Does she have a big garden now? \\ 
\textbf{Speaker B (L2, Thai):} Yes, a lot of plant. Many flower, very beautiful. And your family? \\
\textbf{Speaker A (NS):} They’re good too, thanks for asking. My dad is enjoying painting these days. \\
\textbf{Speaker B (L2, Thai):} Painting? That is nice. Like... picture, um, with... colors? \\ 
\textbf{Speaker A (NS):} Exactly! He loves painting landscapes and nature scenes. So, are you still working at the same company? \\ 
\textbf{Speaker B (L2, Thai):} Yes, I still work there. It is... um, quite busy. Many project now. 
\end{tcolorbox}

\subsubsection{Mandarin Example}
\begin{tcolorbox}[mydialogue] 
        \textbf{Speaker A (NS):} Hi, Xiao Li! How are your preparations going for the upcoming exam? \\
        \textbf{Speaker B (L2, Mandarin):} Oh, hello! Uh, I am study very hard, but some part still difficult for me. \\
        \textbf{Speaker A (NS):} I understand. Which parts are you finding difficult? \\
        \textbf{Speaker B (L2, Mandarin):} Um, the... how to say... the part about economy numbers? \\
        \textbf{Speaker A (NS):} Ah, you mean the statistics section? \\
        \textbf{Speaker B (L2, Mandarin):} Yes, statistics. It is, uh, hard to remember all the formulas. \\
        \textbf{Speaker A (NS):} I can help you with that! We can go over some key formulas together if you’d like. \\
        \textbf{Speaker B (L2, Mandarin):} Really? That would be very nice! Thank you. \\
        \textbf{Speaker A (NS):} No problem! Have you tried using flashcards to memorize them? \\
        \textbf{Speaker B (L2, Mandarin):} Flashcard? Oh, I think I know. The small card with... word and answer? \\
        \textbf{Speaker A (NS):} Exactly! They can be very useful for memorizing details quickly. \\
        \textbf{Speaker B (L2, Mandarin):} I will try that. Um, you think study together good idea? \\
        \textbf{Speaker A (NS):} Yes, definitely! Studying together can make it easier to understand tough concepts. 
\end{tcolorbox}

\subsubsection{Japanese Example}
\begin{tcolorbox}[mydialogue] 
        \textbf{Speaker A (NS):} Hi, everyone. So, we’re here to plan our group vacation. Any thoughts on where we should go? \\
        \textbf{Speaker B (L2, Japanese):} Ah, yes. I think... maybe Kyoto is nice? It have many... uh, historical... um, places. \\
        \textbf{Speaker A (NS):} That’s a great idea! Kyoto is full of history and culture. What do you think we could do there? \\
        \textbf{Speaker B (L2, Japanese):} We can see... um, temples. And maybe, ah, try... uh, Kyoto food? How you say... kyou-ryouri? \\
        \textbf{Speaker A (NS):} Yes, Kyoto cuisine. That sounds delicious! Are there any other activities you'd like to do? \\
        \textbf{Speaker B (L2, Japanese):} I want to... um, wear kimono. It is very... uh, traditional, right? \\
        \textbf{Speaker A (NS):} Absolutely, wearing a kimono in Kyoto would be quite the experience. How do you feel about transportation? Should we drive or take the train? \\
        \textbf{Speaker B (L2, Japanese):} Um, I think... shinkansen is very fast. But maybe... too expensive? \\
        \textbf{Speaker A (NS):} The bullet train is quick but can be pricey. We could consider an express bus? \\
        \textbf{Speaker B (L2, Japanese):} Oh, yes! Express bus is good idea. Maybe, uh, cheaper? \\
        \textbf{Speaker A (NS):} Definitely could be more budget-friendly. When do you think we should go? \\
        \textbf{Speaker B (L2, Japanese):} Um, maybe next month? I check my... schedule. 
\end{tcolorbox}

\subsubsection{Korean Example}
\begin{tcolorbox}[mydialogue] 
        \textbf{Speaker A (NS):} Hi Minji, are you prepared for the exam next week? \\
        \textbf{Speaker B (L2, Korean):} Oh, hello! Um, yes, I think so... but not very sure. It’s difficult, yes? \\
        \textbf{Speaker A (NS):} It can be challenging. Which part do you find the hardest? \\
        \textbf{Speaker B (L2, Korean):} The, um, history part. Too many dates and name to remember. \\
        \textbf{Speaker A (NS):} I understand. Have you tried making flashcards? They can help with memorization. \\
        \textbf{Speaker B (L2, Korean):} Flashcard? Ah, yes! I make some, but still... um, need more practice. \\
        \textbf{Speaker A (NS):} That sounds like a good start! Maybe we can study together? \\
        \textbf{Speaker B (L2, Korean):} Oh, that would be great! When... um, when can we meet? \\
        \textbf{Speaker A (NS):} How about this weekend? Saturday or Sunday work for you? \\
        \textbf{Speaker B (L2, Korean):} Saturday is good. Um, maybe afternoon? \\
        \textbf{Speaker A (NS):} Perfect! We can meet at the library around 2 p.m.? \\
        \textbf{Speaker B (L2, Korean):} Yes, yes, 2 p.m. good. I will bring flashcards. \\
        \textbf{Speaker A (NS):} Awesome. We can quiz each other and go over the main topics. \\
        \textbf{Speaker B (L2, Korean):} Sounds nice. Thank you for help! 
\end{tcolorbox}

\subsubsection{Urdu Example}
\begin{tcolorbox}[mydialogue]  
\textbf{Speaker A (NS):} Oh, you work with computers? That's interesting! What do you do exactly? \\
\textbf{Speaker B (L2, Urdu):} Yes, um, I am doing software develop... developing. I make, uh, programs and apps. \\ \textbf{Speaker A (NS):} Software development, that sounds exciting! How long have you been doing that? \\ 
\textbf{Speaker B (L2, Urdu):} It has been, um, about three year now. I start after university. \\ 
\textbf{Speaker A (NS):} That's quite some time. Which university did you attend? \\ 
\textbf{Speaker B (L2, Urdu):} I study at Lahore University. It is good for science, technology... um, these things. \\ \textbf{Speaker A (NS):} Lahore University is well-known. What was your major? \\ 
\textbf{Speaker B (L2, Urdu):} My major was computer science. I always like computers. \\ 
\textbf{Speaker A (NS):} That's great! What inspired you to get into computer science? \\ 
\textbf{Speaker B (L2, Urdu):} Um, I like solve problems. And, uh, computers are very... um, how to say... powerful for this? 
\end{tcolorbox}

\subsubsection{Cantonese Example}
\begin{tcolorbox}[mydialogue] 
        \textbf{Speaker A (NS):} Hi there! Can I help you find something today? \\
        \textbf{Speaker B (L2, Cantonese):} Um, yes, please. I looking for... uh, how to say... cleaning thing? \\
        \textbf{Speaker A (NS):} Do you mean cleaning supplies, like a mop or detergent? \\
        \textbf{Speaker B (L2, Cantonese):} Yes, yes! Detergent. I need for washing clothes. \\
        \textbf{Speaker A (NS):} Alright, the laundry detergent is in aisle six. Do you need any help choosing a brand? \\
        \textbf{Speaker B (L2, Cantonese):} Ah, too many brand. Can you recommend? Which is good? \\
        \textbf{Speaker A (NS):} Of course! Tide is quite popular and cleans well. Do you have a preference for liquid or powder? \\
        \textbf{Speaker B (L2, Cantonese):} Uh, I think maybe liquid. Easier to use, I think. \\
        \textbf{Speaker A (NS):} Great choice! Is there anything else you need today? \\
        \textbf{Speaker B (L2, Cantonese):} Um, yes, maybe... how you say... remove spot? On clothes? \\
        \textbf{Speaker A (NS):} Spot remover or stain remover. It's where the laundry detergent is too. \\
        \textbf{Speaker B (L2, Cantonese):} Okay, thank you. I will buy it. Um, question... do you have bags that... um, recycle? \\
        \textbf{Speaker A (NS):} Yes, we have reusable bags at the checkout area. They're a great option for the environment. \\
        \textbf{Speaker B (L2, Cantonese):} Ah, good! I will buy that also. Thank you so much. 
\end{tcolorbox}

\subsubsection{Malay Example}

\begin{tcolorbox}[mydialogue]
\textbf{Speaker A (NS):} Hi there! I heard Malaysia has a lot of interesting festivals. Can you tell me about one of them? \\ 
\textbf{Speaker B (L2, Malay):} Oh, yes! We have many. Um, one famous is Hari Raya Aidilfitri. \\
\textbf{Speaker A (NS):} Sounds interesting! Can you explain what happens during it? \\
\textbf{Speaker B (L2, Malay):} Yes, sure. It is, uh... celebration after fasting month, Ramadan.\\
\textbf{Speaker A (NS):} Oh, right. So, what do people usually do during Hari Raya?\\
\textbf{Speaker B (L2, Malay):} We, uh, visit family. Have... big meals. Um, special food like rendang, ketupat.\\
\textbf{Speaker A (NS):} That sounds delicious! Is there anything else that’s part of the celebration? \\
\textbf{Speaker B (L2, Malay):} Yes, we also... um, give... how to say... small money packets to children.\\
\textbf{Speaker A (NS):} Ah, like gifts?\\
\textbf{Speaker B (L2, Malay):} Yes, but... um, we call it ``duit raya.''\\
\end{tcolorbox}

For Other languages generated data, please refer to ~\url{https://github.com/RenaGao/LLMPirorknowledge} for each dialogues. 

\subsection{L2 Density Results for GPT-4o Generations}\label{sec:DensityResults}
For Other LLMs generated data, please refer to ~\url{https://github.com/RenaGao/LLMPirorknowledge} for each LLM density figure. 

\begin{figure*}[!h]
    \centering
    \begin{subfigure}[b]{0.24\textwidth} 
    \includegraphics[width=\textwidth]{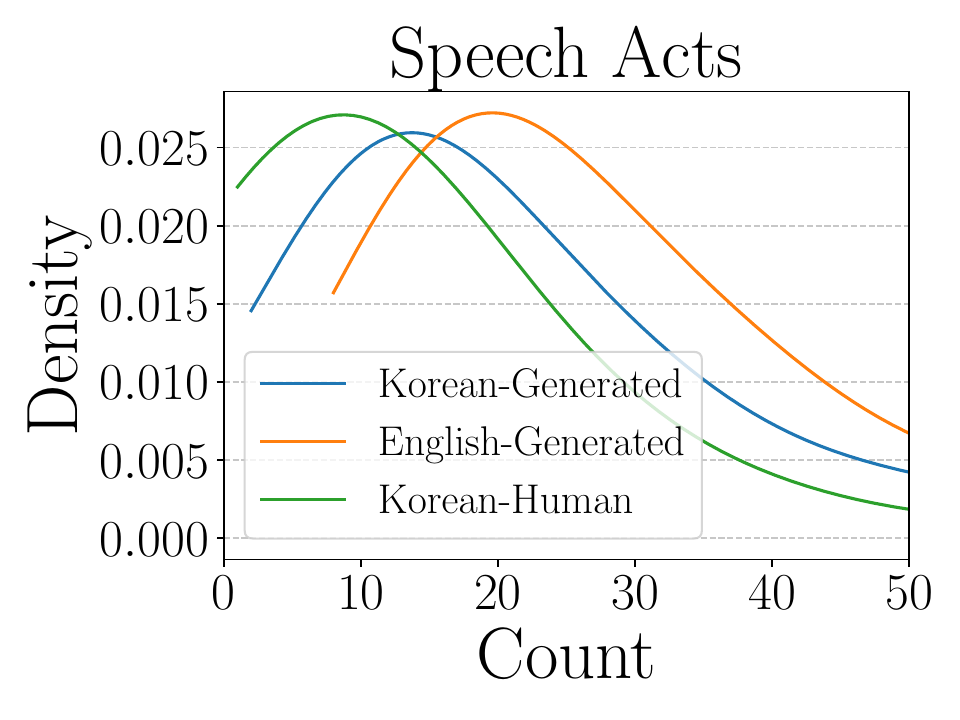}
        \caption{Speech Acts}
    \end{subfigure}
    \hfill 
    \begin{subfigure}[b]{0.24\textwidth} 
    \includegraphics[width=\textwidth]{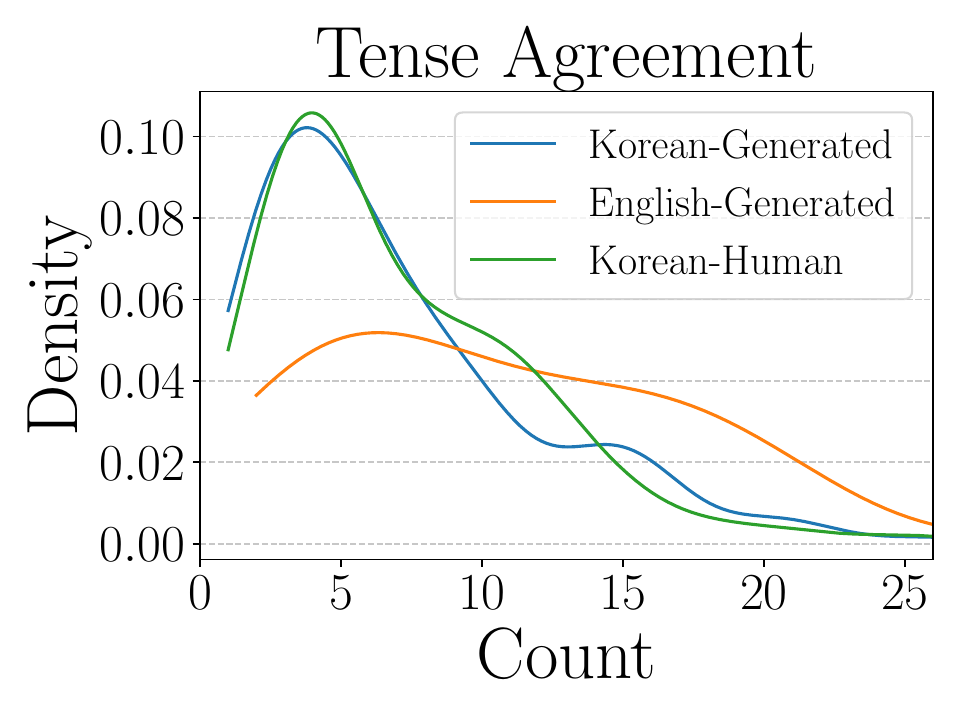}
        \caption{Tense Agreement}
    \end{subfigure}
    \hfill
    \begin{subfigure}[b]{0.24\textwidth}
    \includegraphics[width=\textwidth]{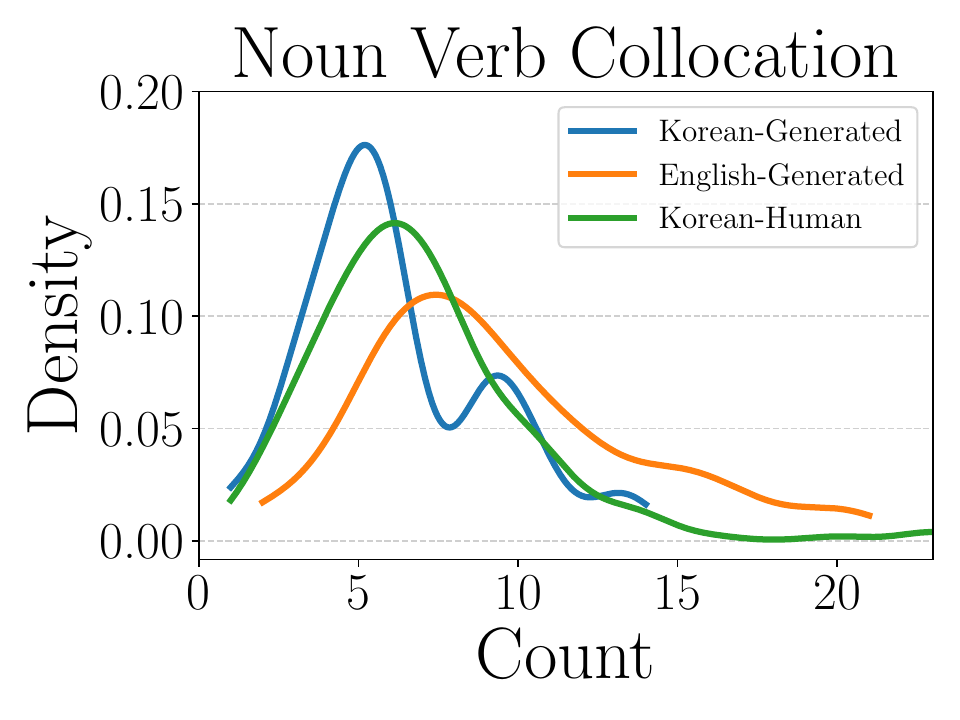}
        \caption{Noun-Verb Collocation}
    \end{subfigure}
    \hfill
    \begin{subfigure}[b]{0.24\textwidth}
    \includegraphics[width=\textwidth]{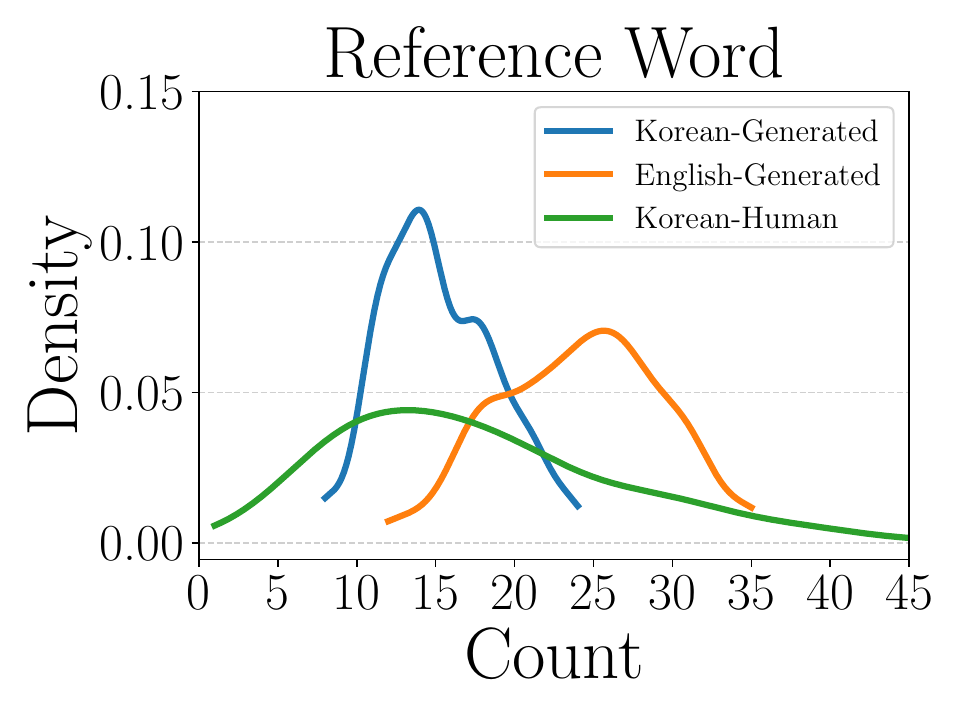}
        \caption{Reference Word}
    \end{subfigure}
    \\
    \begin{subfigure}[b]{0.24\textwidth}
    \includegraphics[width=\textwidth]{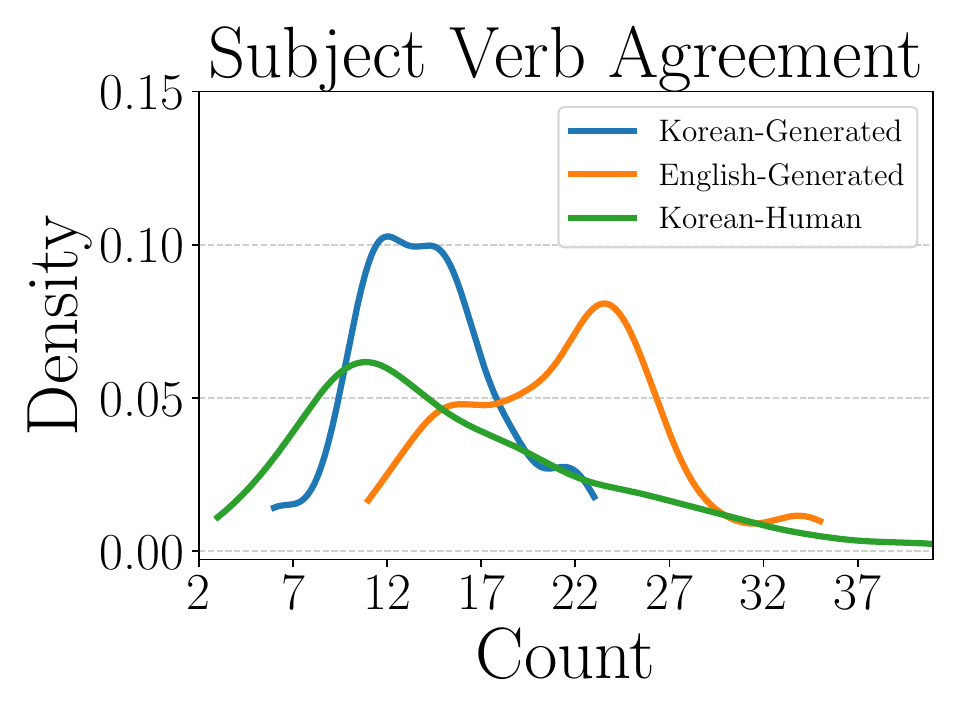}
        \caption{Subject Verb Agreement}
    \end{subfigure}
    \hfill
    \begin{subfigure}[b]{0.24\textwidth}
        \includegraphics[width=\textwidth]{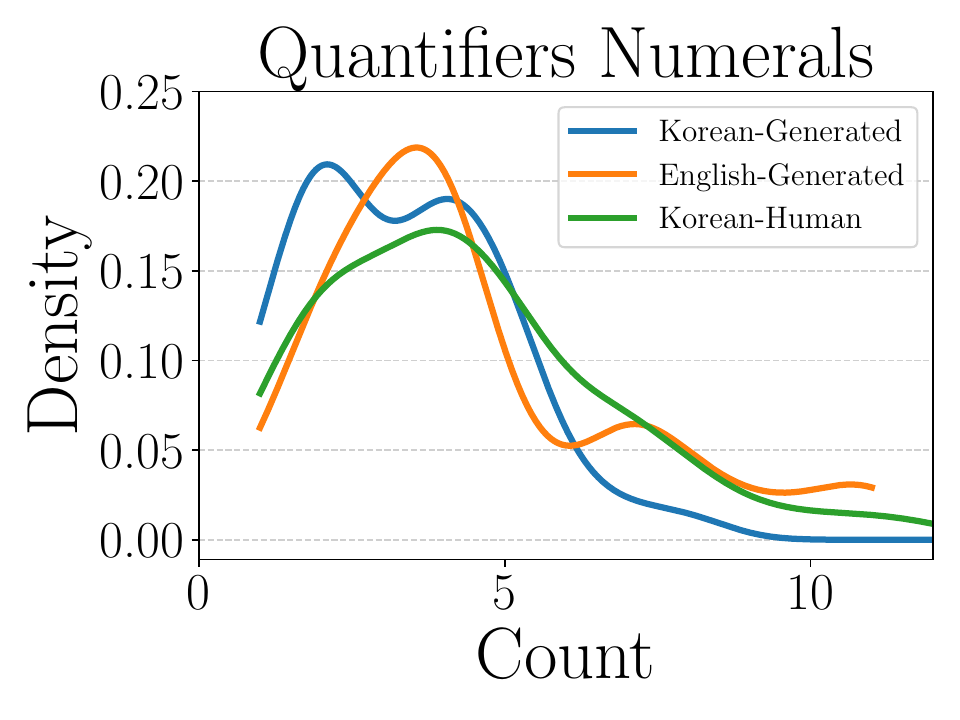}
        \caption{Quantifiers Numerals}
    \end{subfigure}
    \hfill
    \begin{subfigure}[b]{0.24\textwidth}
        \includegraphics[width=\textwidth]{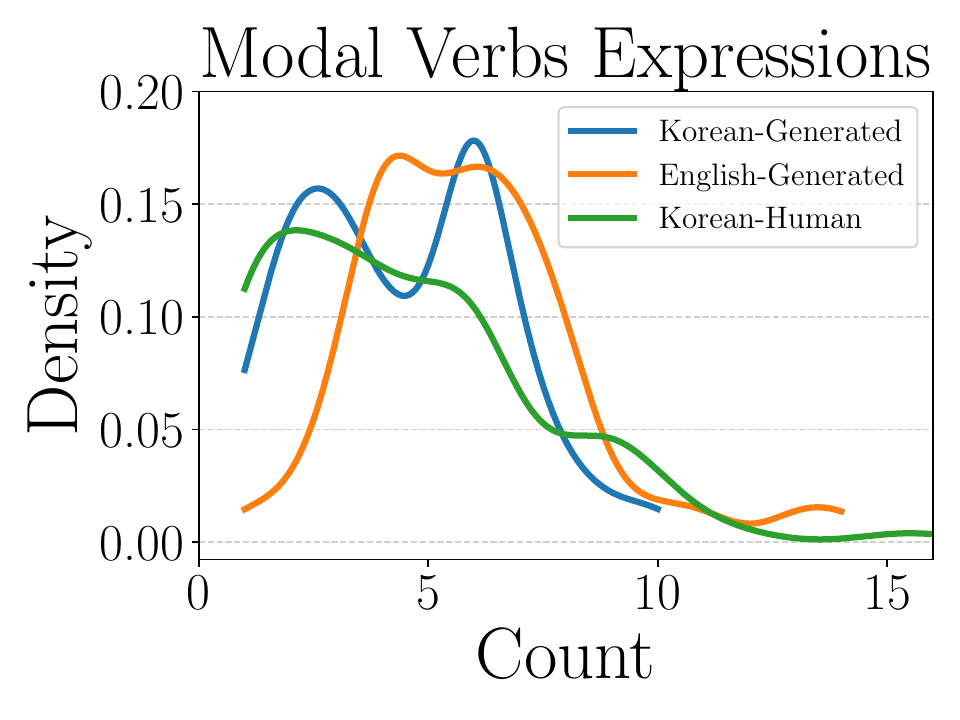}
        \caption{Modal Verbs Expressions}
    \end{subfigure}
    \hfill
    \begin{subfigure}[b]{0.24\textwidth}
    \includegraphics[width=\textwidth]{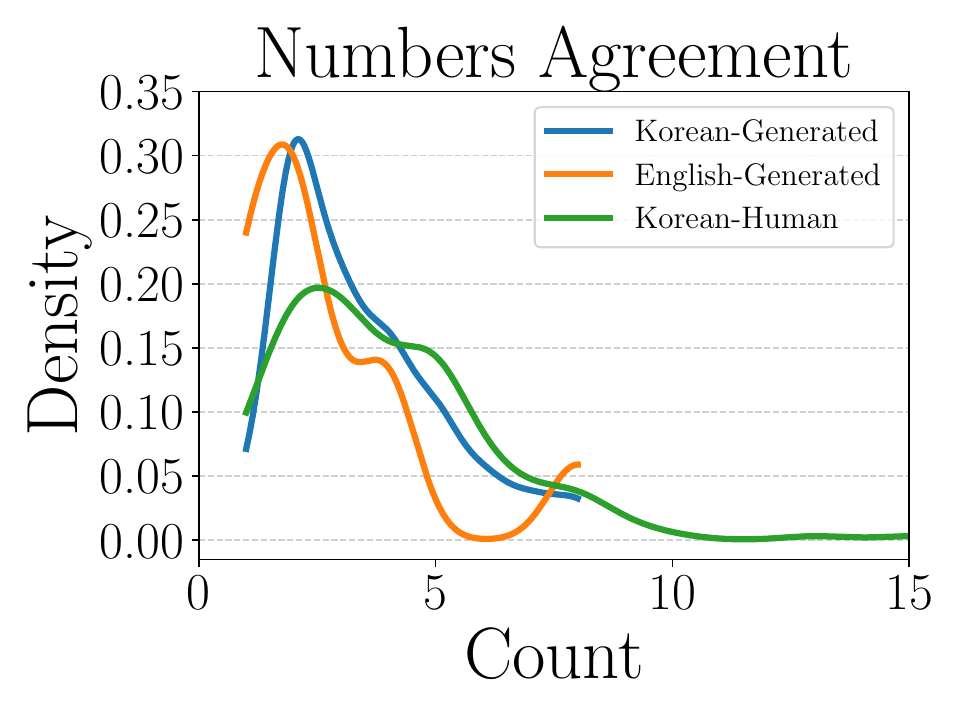}
        \caption{Numbers Agreement}
    \end{subfigure}
    \caption{Full density results for L2 generation dialogue via Korean L1s}
    \label{fig:L2alldens_kor}
\end{figure*}

\begin{figure*}[htbp]
    \centering
    \begin{subfigure}[b]{0.24\textwidth} 
        \includegraphics[width=\textwidth]{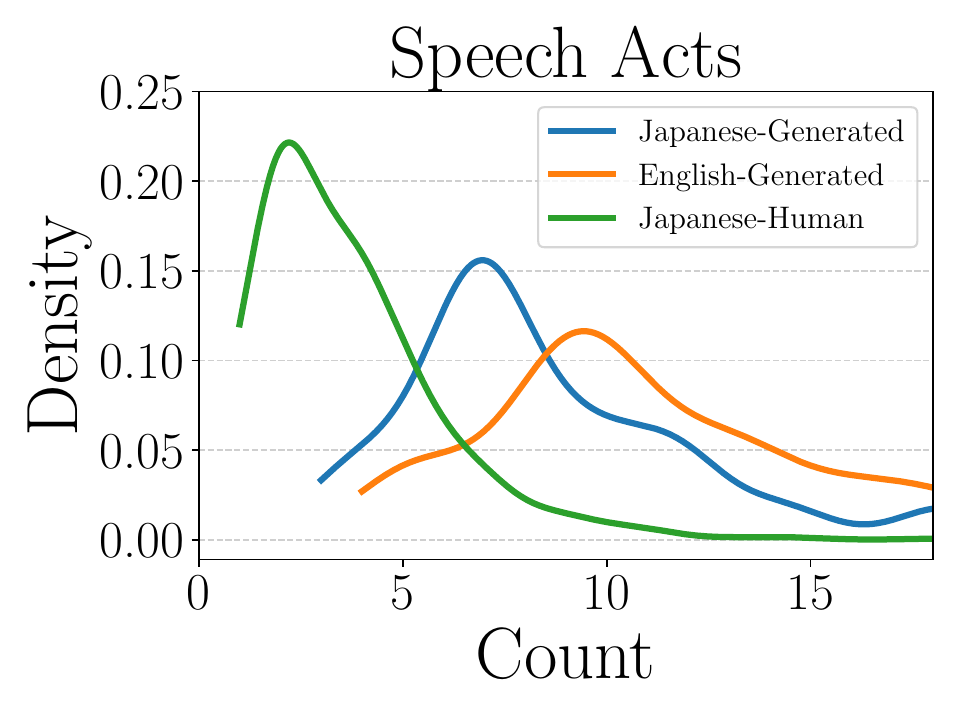}
        \caption{Speech Acts}
    \end{subfigure}
    \hfill 
    \begin{subfigure}[b]{0.24\textwidth} 
        \includegraphics[width=\textwidth]{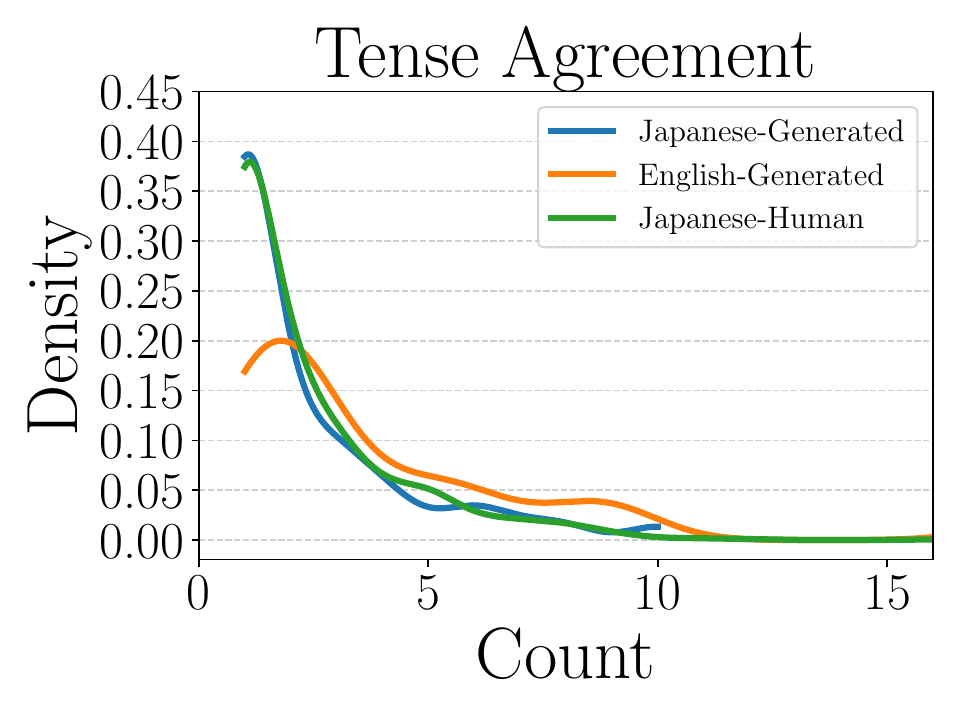}
        \caption{Tense Agreement}
    \end{subfigure}
    \hfill
    \begin{subfigure}[b]{0.24\textwidth}
        \includegraphics[width=\textwidth]{appendix_figs/Noun_Verb_Collocation_Japanese.pdf}
        \caption{Noun-Verb Collocation}
    \end{subfigure}
    \hfill
    \begin{subfigure}[b]{0.24\textwidth}
    \includegraphics[width=\textwidth]{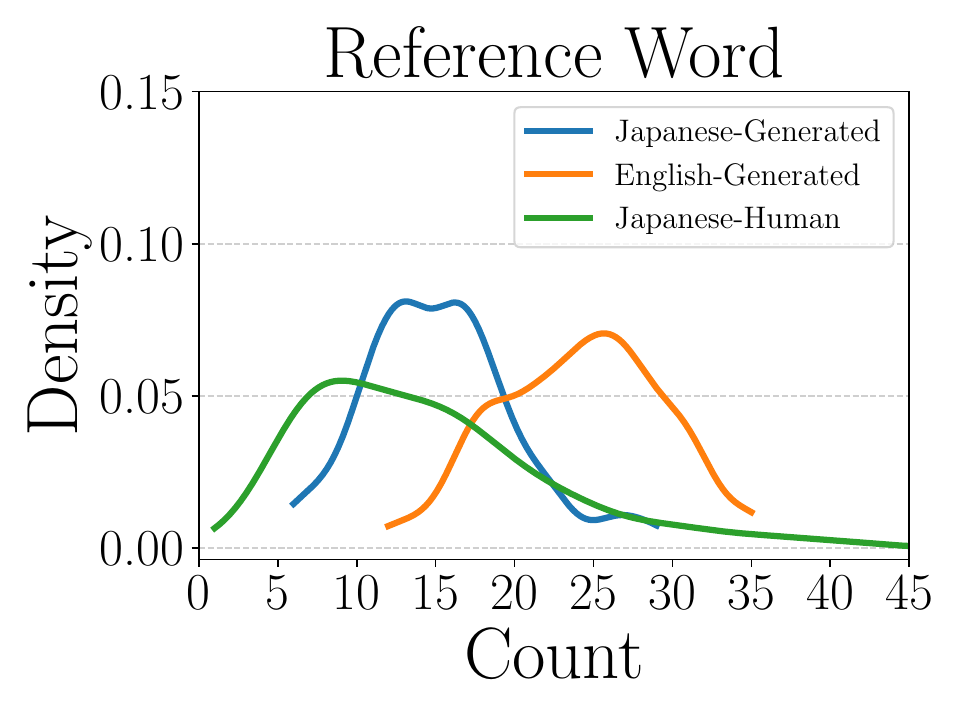}
        \caption{Reference Word}
    \end{subfigure}
    \\
    \begin{subfigure}[b]{0.24\textwidth}
    \includegraphics[width=\textwidth]{appendix_figs/Subject_Verb_Agreement_Japanese.pdf}
        \caption{Subject Verb Agreement}
    \end{subfigure}
    \hfill
    \begin{subfigure}[b]{0.24\textwidth}
        \includegraphics[width=\textwidth]{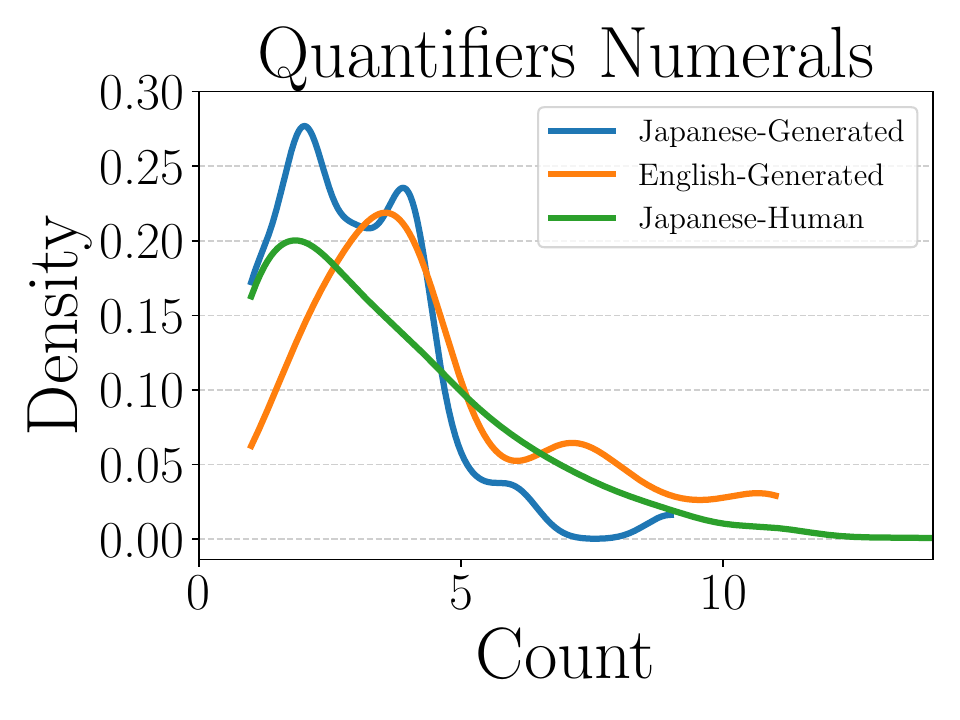}
        \caption{Quantifiers Numerals}
    \end{subfigure}
    \hfill
    \begin{subfigure}[b]{0.24\textwidth}
        \includegraphics[width=\textwidth]{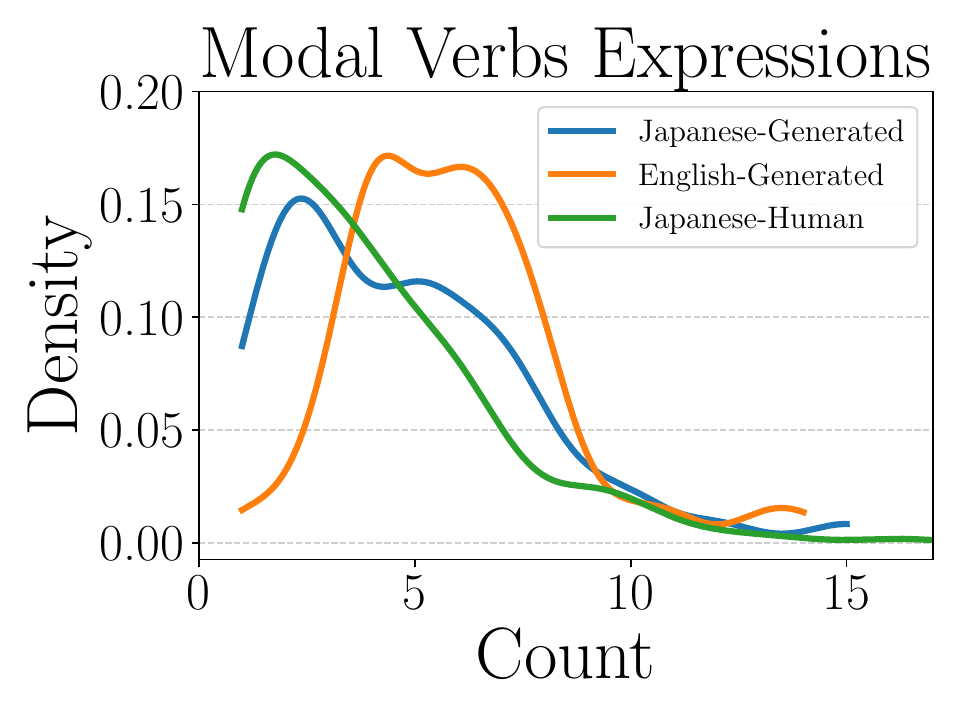}
        \caption{Modal Verbs Expressions}
    \end{subfigure}
    \hfill
    \begin{subfigure}[b]{0.24\textwidth}
    \includegraphics[width=\textwidth]{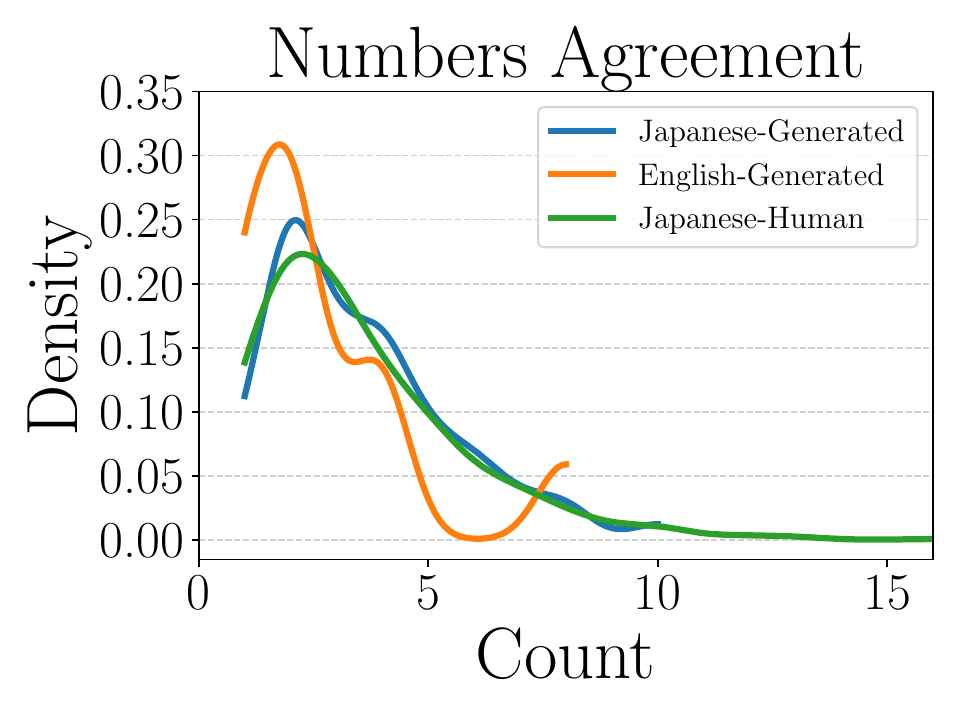}
        \caption{Numbers Agreement}
    \end{subfigure}
    \caption{Full density results for L2 generation dialogue via Japanese L1s} 
    \label{fig:L2alldens_jpn}
\end{figure*}
\begin{figure*}[!h]
    \centering
    \begin{subfigure}[b]{0.24\textwidth} 
        \includegraphics[width=\textwidth]{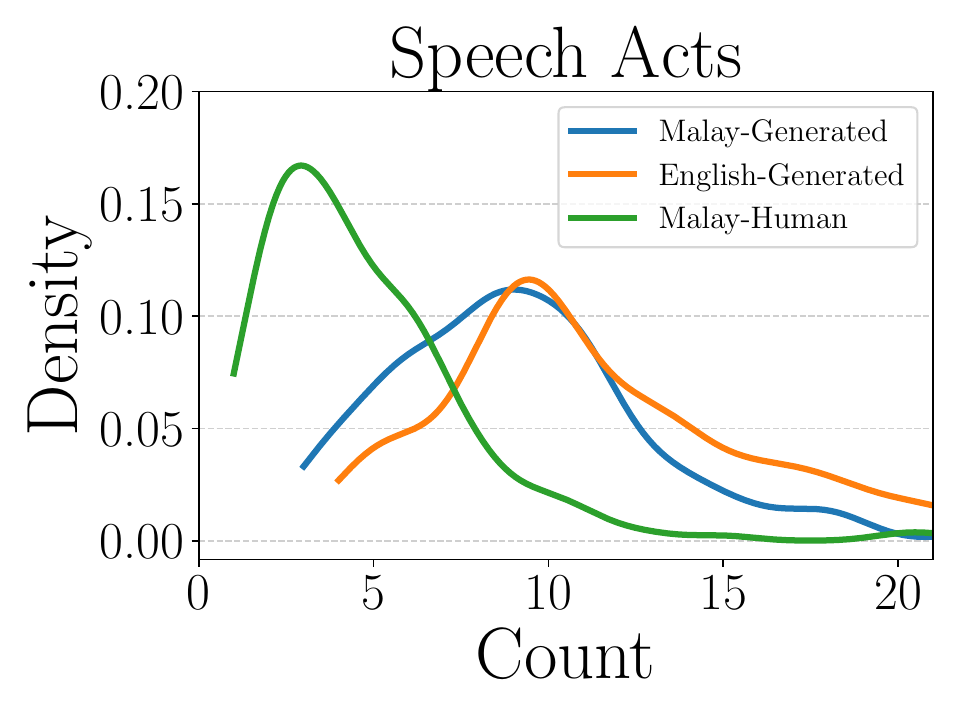}
        \caption{Speech Acts}
    \end{subfigure}
    \hfill 
    \begin{subfigure}[b]{0.24\textwidth} 
        \includegraphics[width=\textwidth]{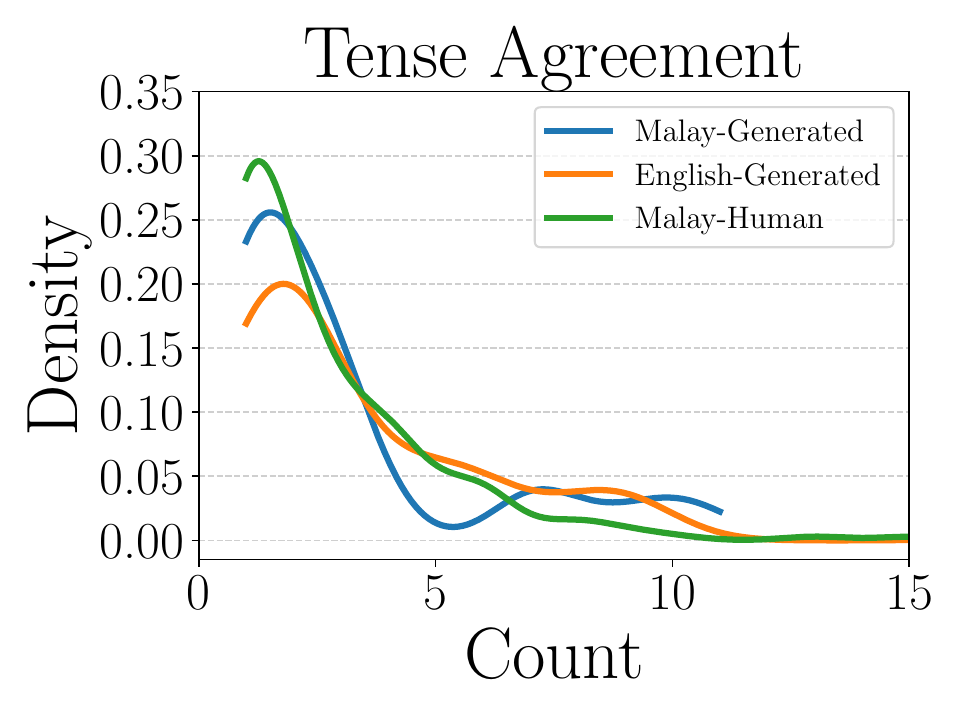}
        \caption{Tense Agreement} 
    \end{subfigure}
    \hfill
    \begin{subfigure}[b]{0.24\textwidth}
        \includegraphics[width=\textwidth]{appendix_figs/Noun_Verb_Collocation_Malay.pdf}
        \caption{Noun-Verb Collocation}
    \end{subfigure}
    \hfill
    \begin{subfigure}[b]{0.24\textwidth}
    \includegraphics[width=\textwidth]{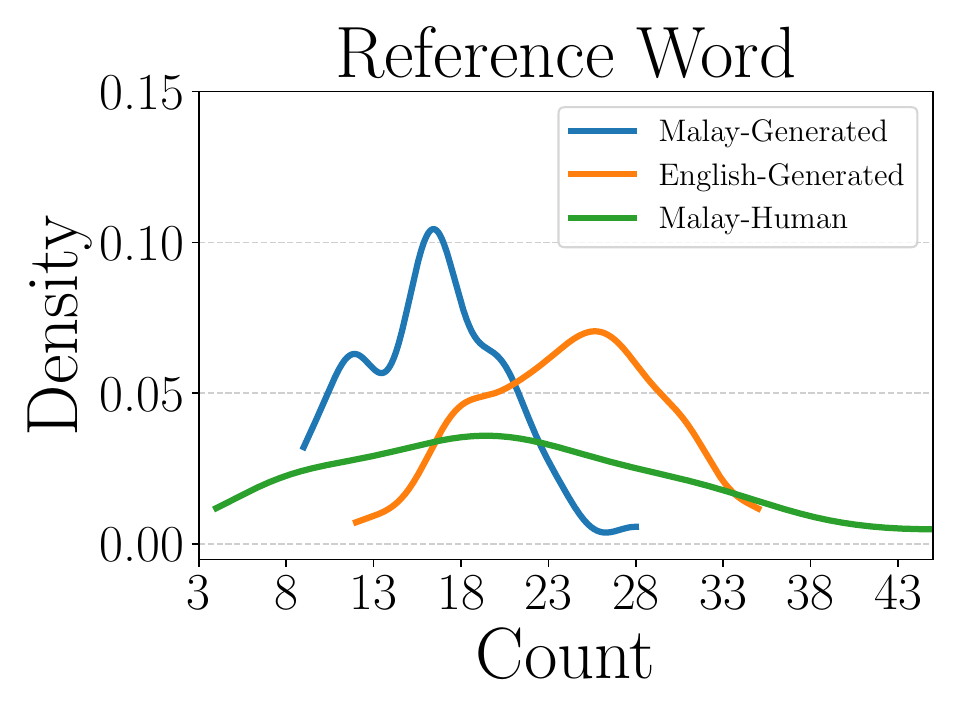}
        \caption{Reference Word}
    \end{subfigure}
    \\
    \begin{subfigure}[b]{0.24\textwidth}
        \includegraphics[width=\textwidth]{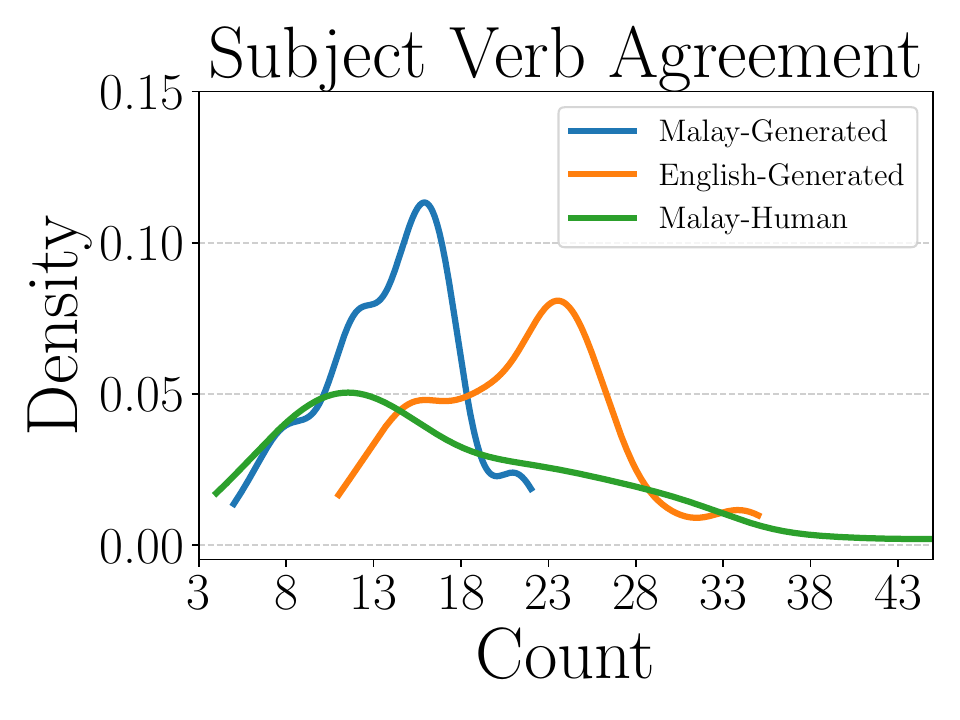}
        \caption{Subject Verb Agreement}
    \end{subfigure}
    \hfill
    \begin{subfigure}[b]{0.24\textwidth}
        \includegraphics[width=\textwidth]{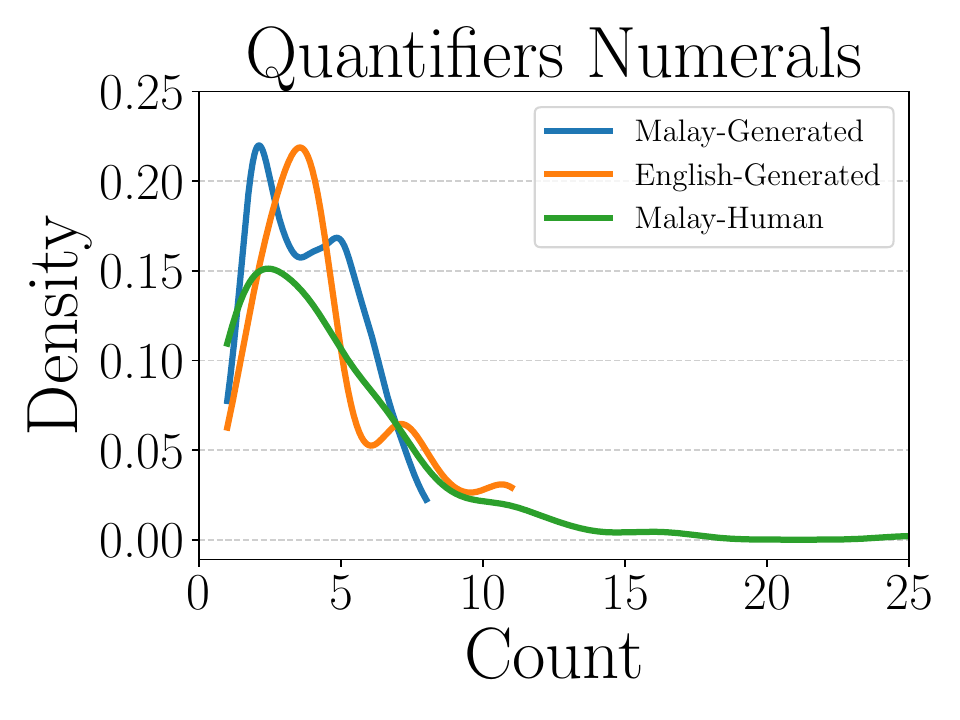}
        \caption{Quantifiers Numerals}
    \end{subfigure}
    \hfill
    \begin{subfigure}[b]{0.24\textwidth}
        \includegraphics[width=\textwidth]{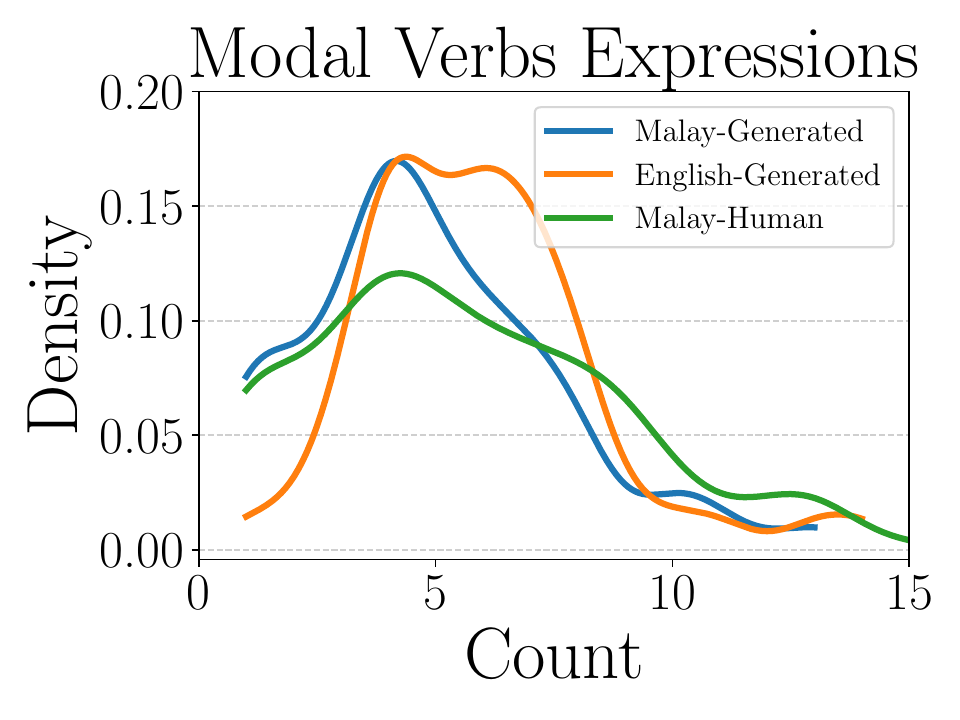}
        \caption{Modal Verbs Expressions}
    \end{subfigure}
    \hfill
    \begin{subfigure}[b]{0.24\textwidth}
    \includegraphics[width=\textwidth]{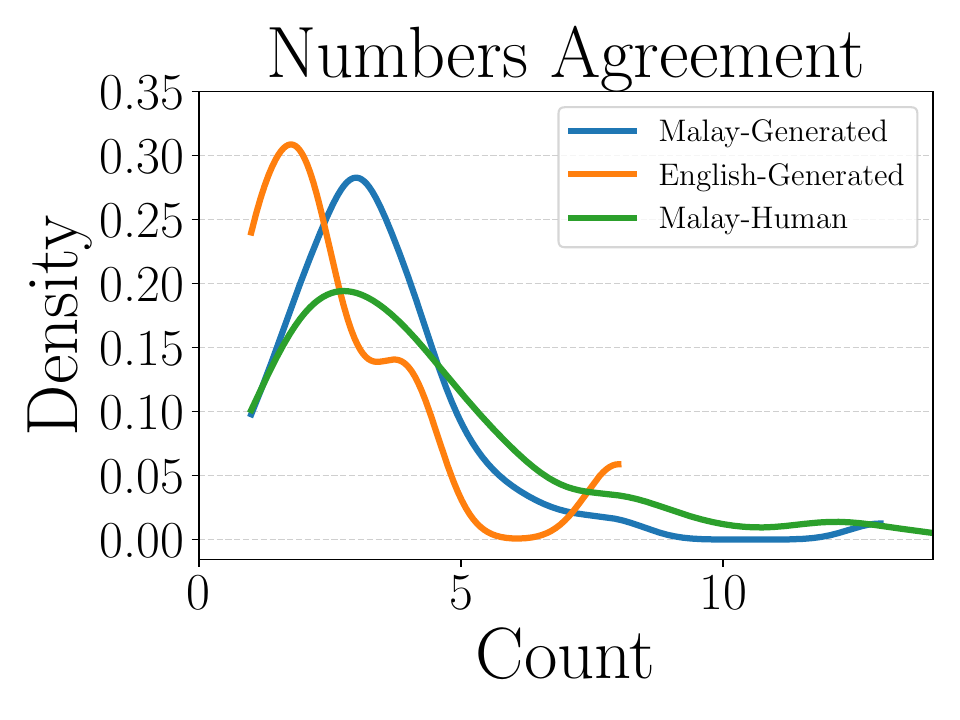}
        \caption{Numbers Agreement}
    \end{subfigure}
    \caption{Full density results for L2 generation dialogue via Malay L1s} 
    \label{fig:L2alldens_malay}
\end{figure*}
\begin{figure*}[!h]
    \centering
    \begin{subfigure}[b]{0.24\textwidth} 
        \includegraphics[width=\textwidth]{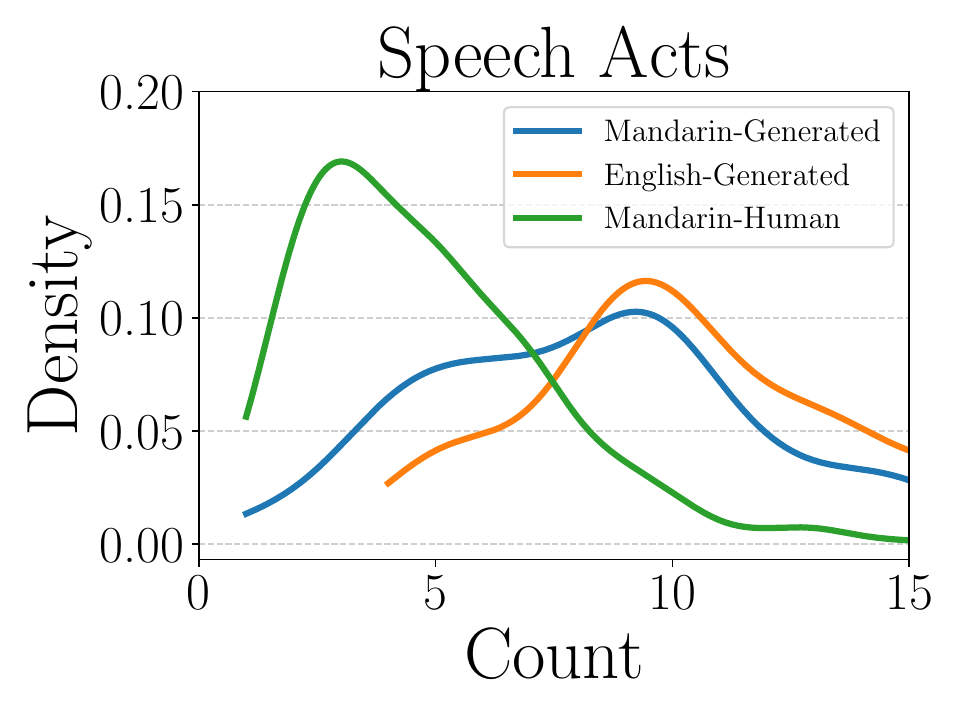}
        \caption{Speech Acts}
    \end{subfigure}
    \hfill 
    \begin{subfigure}[b]{0.24\textwidth} 
        \includegraphics[width=\textwidth]{appendix_figs/Tense_Agreement_Mandarin.pdf}
        \caption{Tense Agreement}
    \end{subfigure}
    \hfill
    \begin{subfigure}[b]{0.24\textwidth}
        \includegraphics[width=\textwidth]{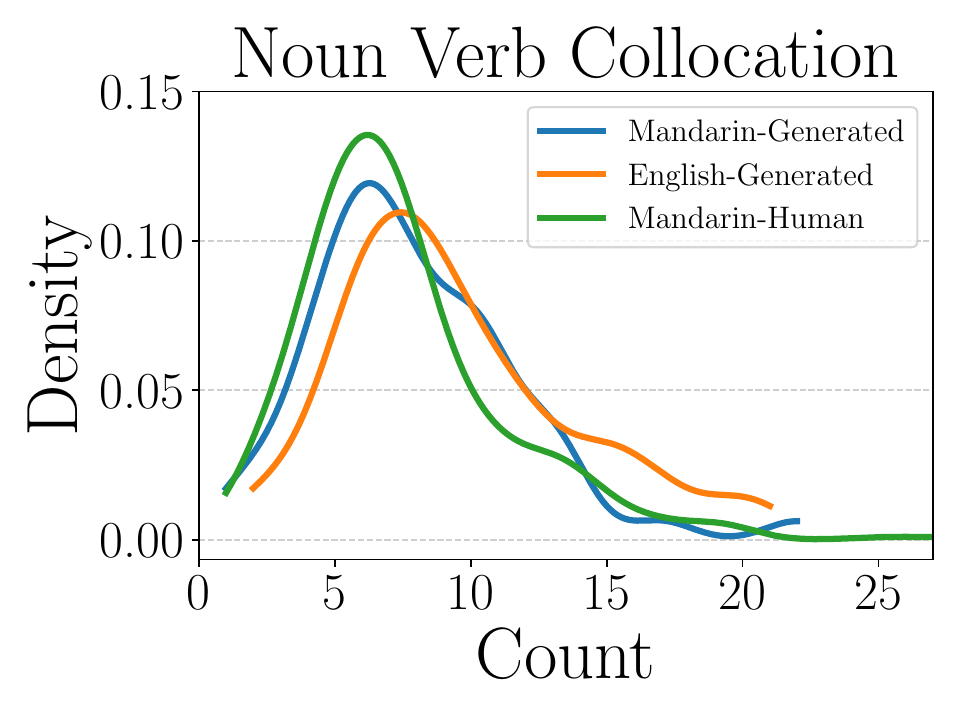}
        \caption{Noun-Verb Collocation}
    \end{subfigure}
    \hfill
    \begin{subfigure}[b]{0.24\textwidth}
    \includegraphics[width=\textwidth]{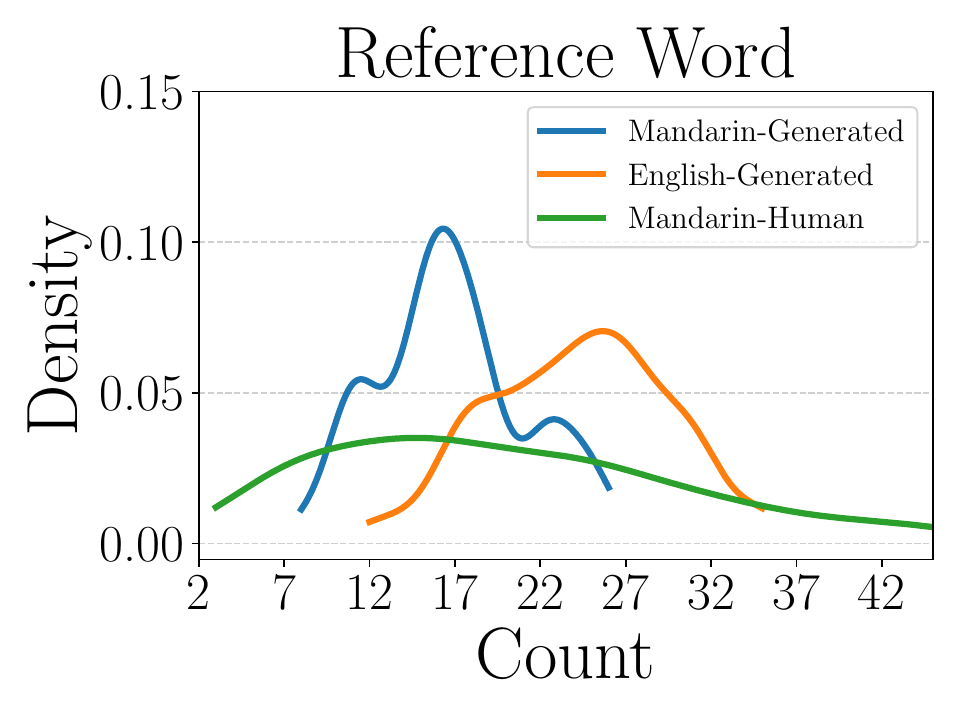}
        \caption{Reference Word}
    \end{subfigure}
    \\
    \begin{subfigure}[b]{0.24\textwidth}
        \includegraphics[width=\textwidth]{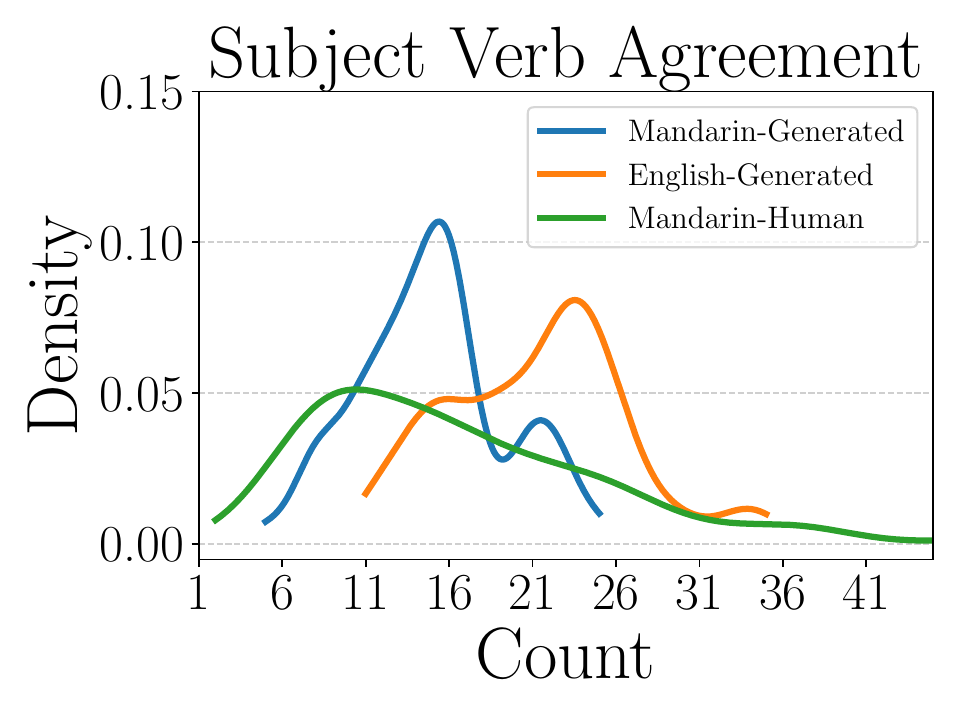}
        \caption{Subject Verb Agreement}
    \end{subfigure}
    \hfill
    \begin{subfigure}[b]{0.24\textwidth}
        \includegraphics[width=\textwidth]{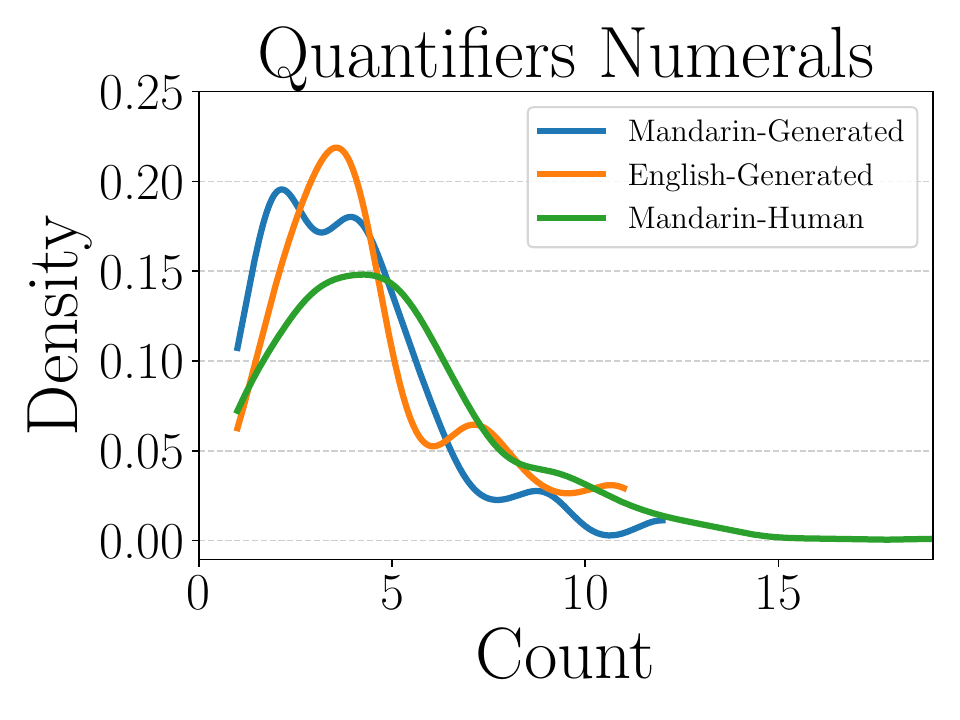}
        \caption{Quantifiers Numerals}
    \end{subfigure}
    \hfill
    \begin{subfigure}[b]{0.24\textwidth}
        \includegraphics[width=\textwidth]{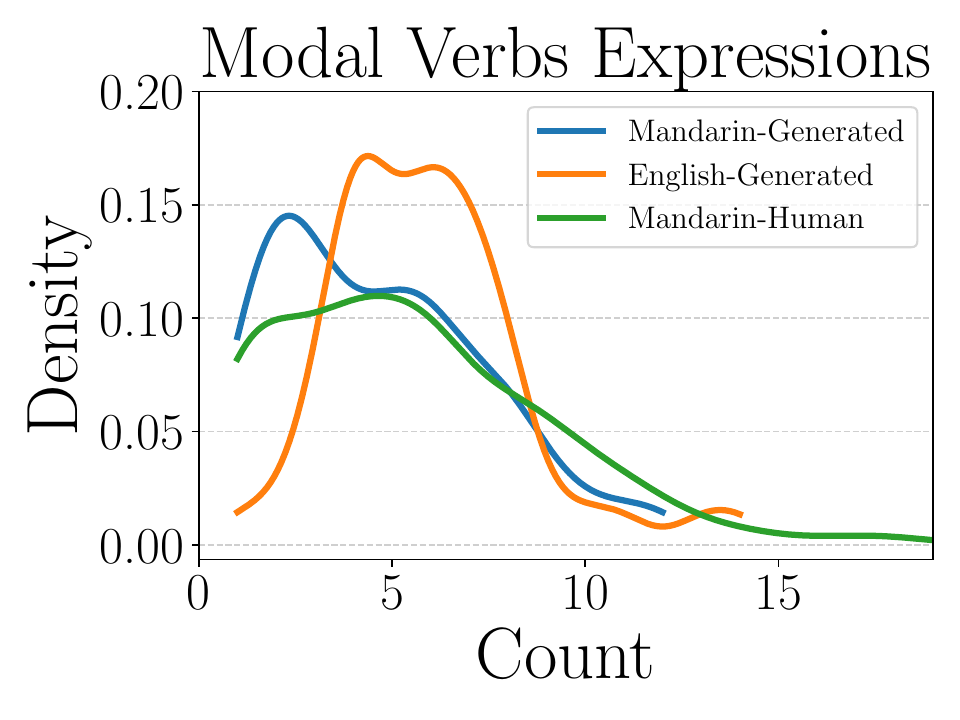}
        \caption{Modal Verbs Expressions}
    \end{subfigure}
    \hfill
    \begin{subfigure}[b]{0.24\textwidth}
    \includegraphics[width=\textwidth]{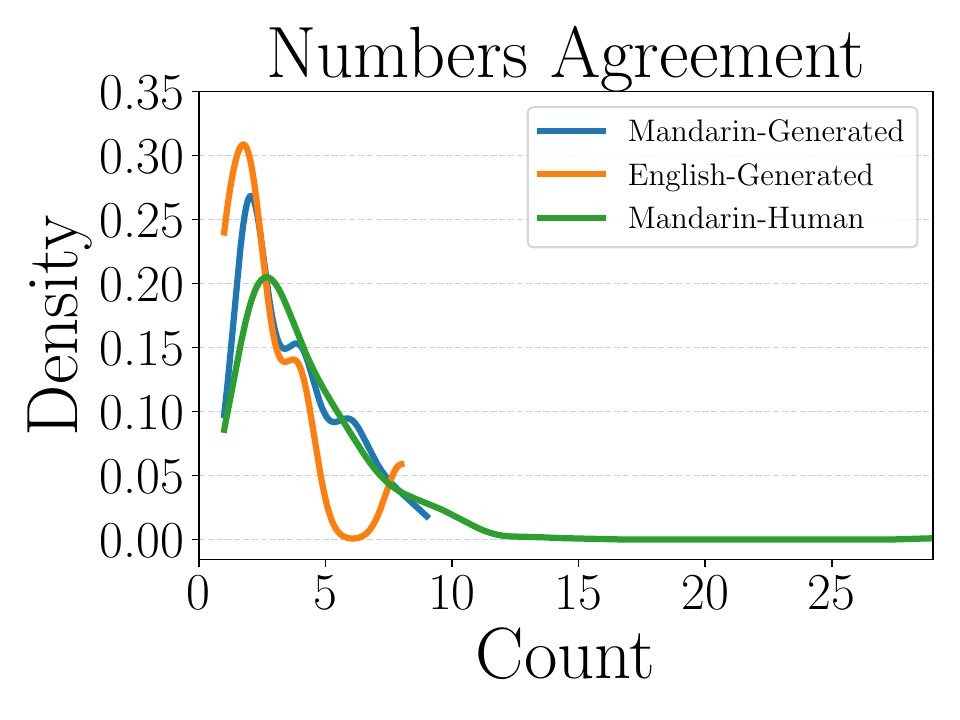}
        \caption{Numbers Agreement}
    \end{subfigure}
    \caption{Full density results for L2 generation dialogue via Mandarin L1s} 
    \label{fig:L2alldens_man}
\end{figure*}

\begin{figure*}[!h]
    \centering
    \begin{subfigure}[b]{0.24\textwidth} 
        \includegraphics[width=\textwidth]{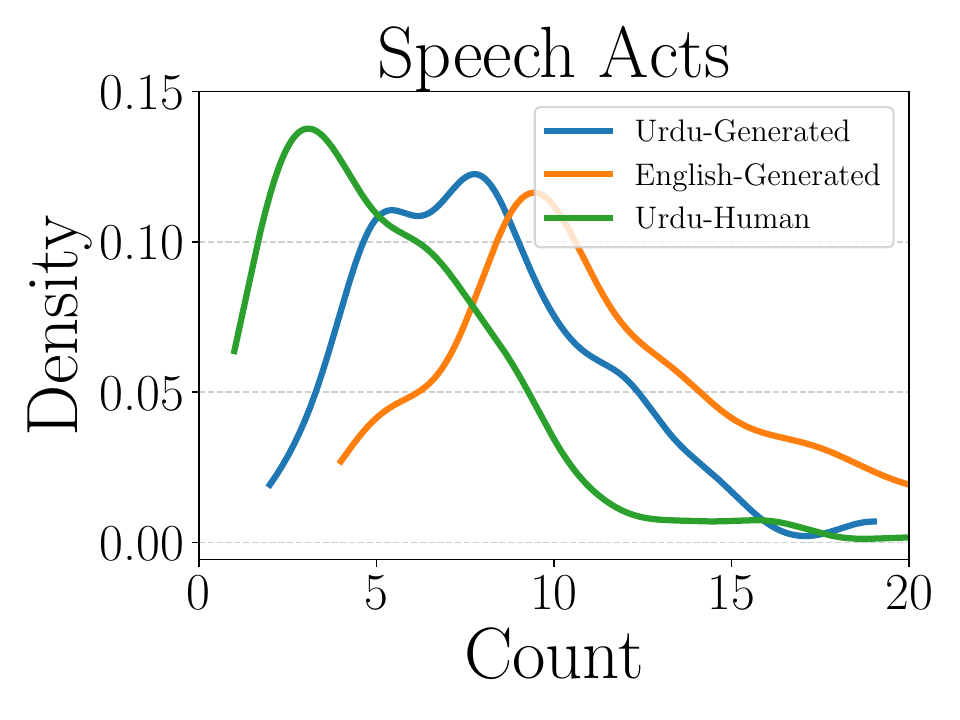}
        \caption{Speech Acts}
    \end{subfigure}
    \hfill 
    \begin{subfigure}[b]{0.24\textwidth} 
        \includegraphics[width=\textwidth]{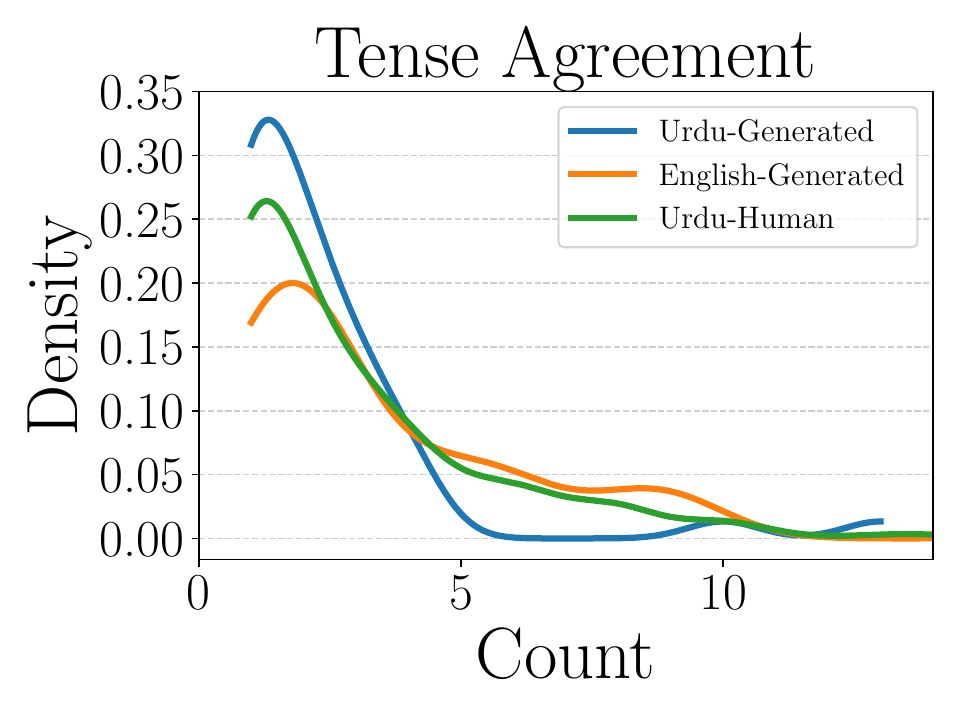}
        \caption{Tense Agreement} 
    \end{subfigure}
    \hfill
    \begin{subfigure}[b]{0.24\textwidth}
        \includegraphics[width=\textwidth]{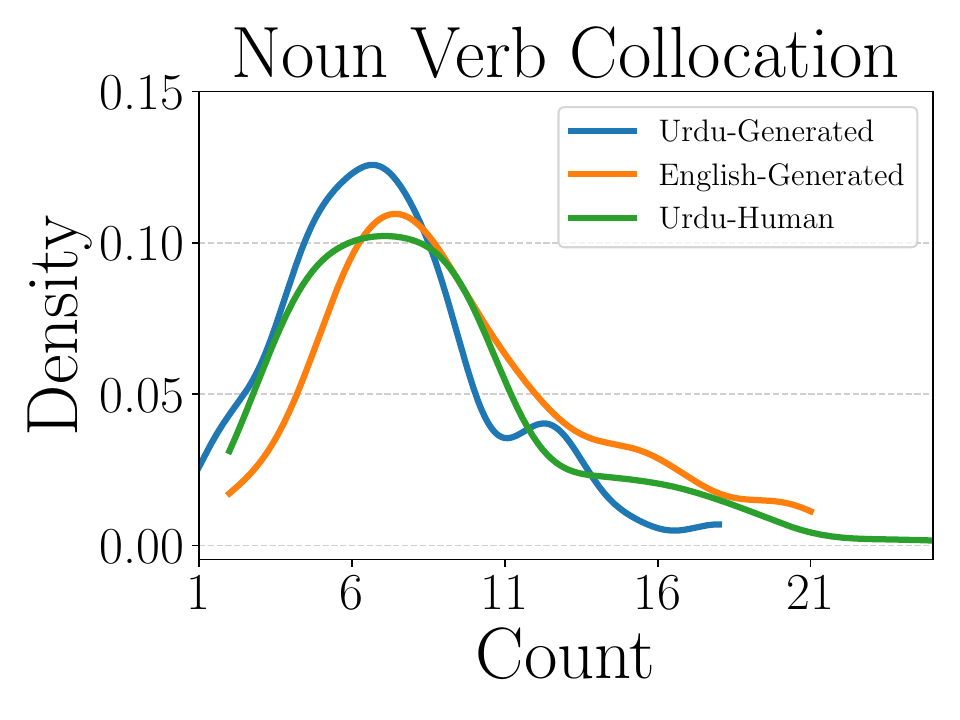}
        \caption{Noun-Verb Collocation}
    \end{subfigure}
    \hfill
    \begin{subfigure}[b]{0.24\textwidth}
    \includegraphics[width=\textwidth]{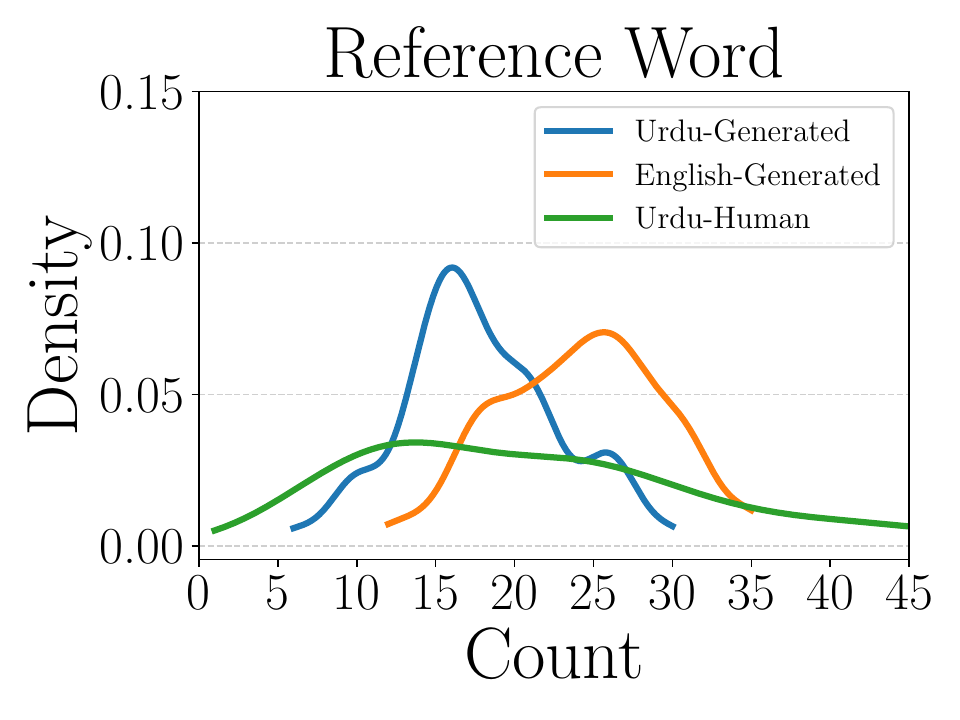}
        \caption{Reference Word}
    \end{subfigure}
    \\
    \begin{subfigure}[b]{0.24\textwidth}
        \includegraphics[width=\textwidth]{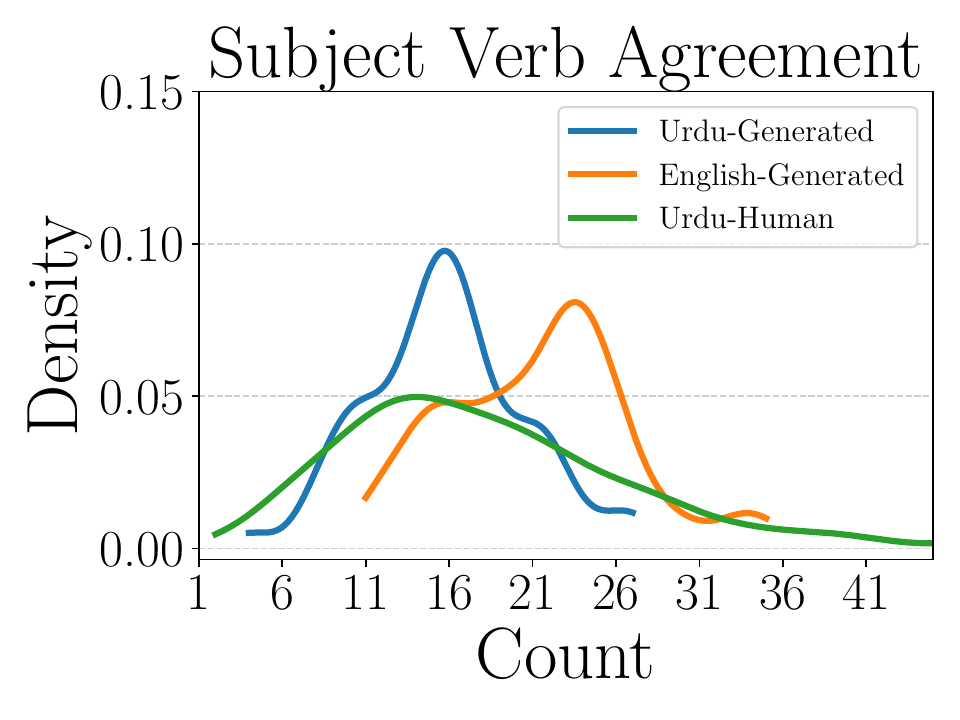}
        \caption{Subject Verb Agreement}
    \end{subfigure}
    \hfill
    \begin{subfigure}[b]{0.24\textwidth}
        \includegraphics[width=\textwidth]{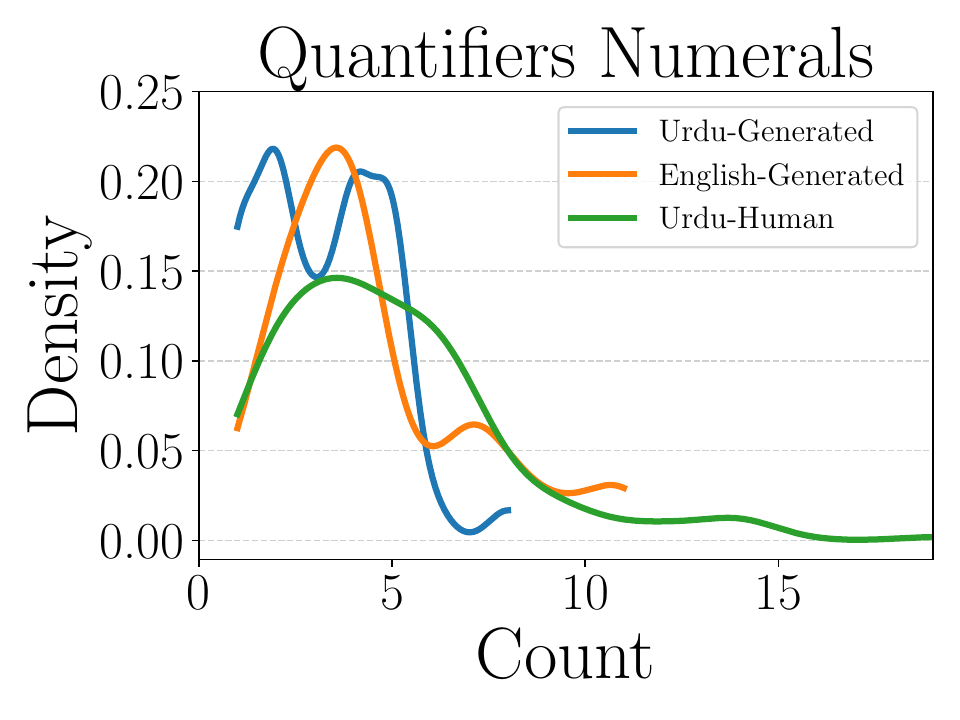}
        \caption{Quantifiers Numerals}
    \end{subfigure}
    \hfill
    \begin{subfigure}[b]{0.24\textwidth}
        \includegraphics[width=\textwidth]{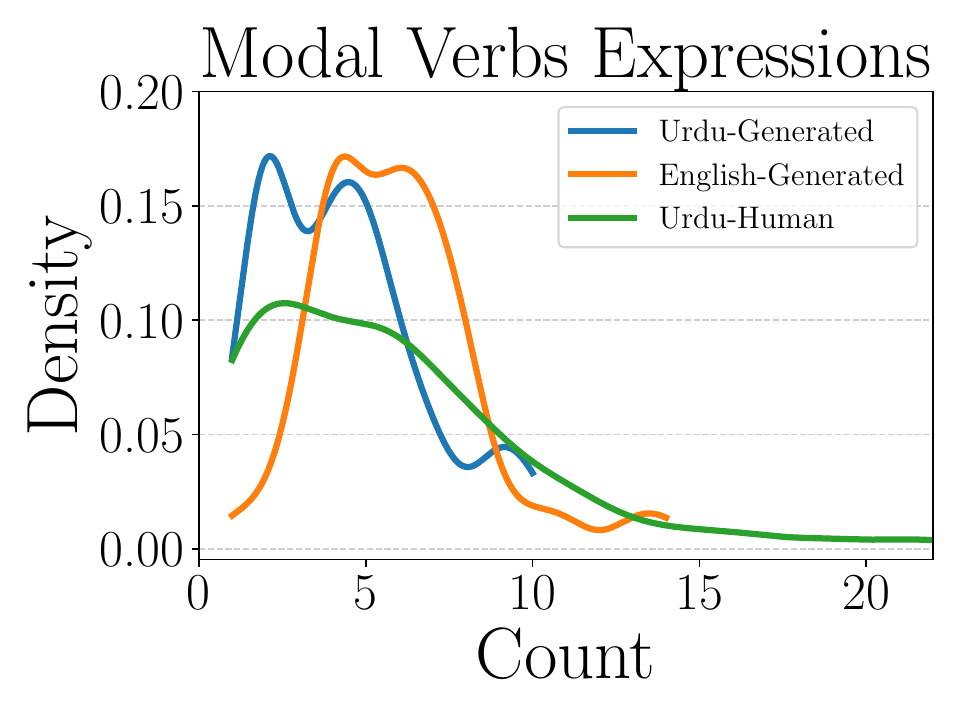}
        \caption{Modal Verbs Expressions}
    \end{subfigure}
    \hfill
    \begin{subfigure}[b]{0.24\textwidth}
    \includegraphics[width=\textwidth]{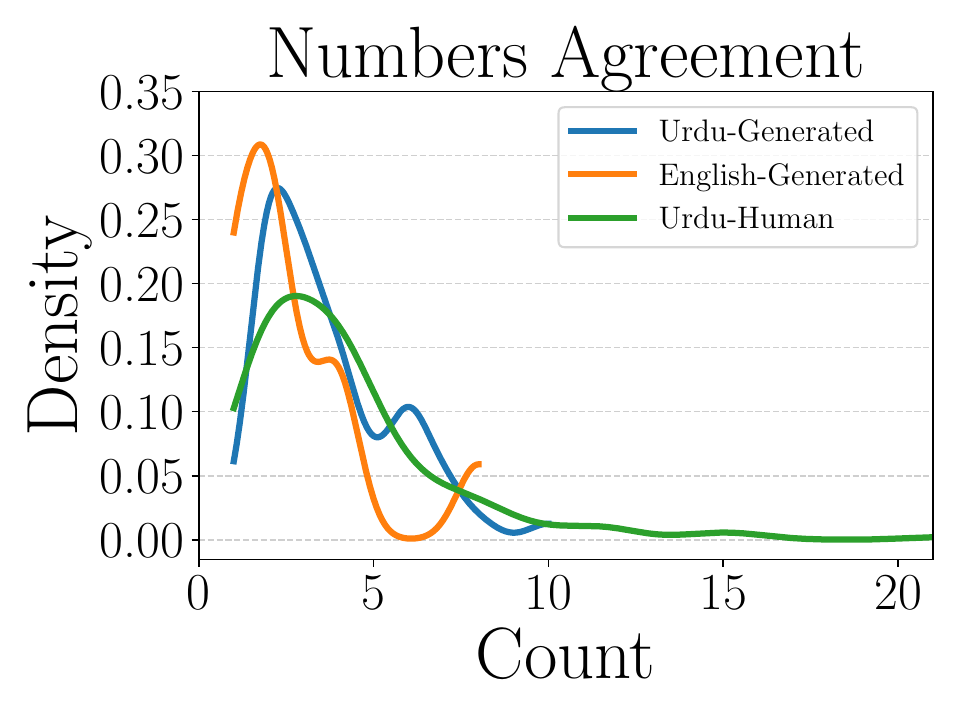}
        \caption{Numbers Agreement}
    \end{subfigure}
    \caption{Full density results for L2 generation dialogue via Urdu L1s} 
    \label{fig:L2alldens_urdu}
\end{figure*}
\begin{figure*}[!h]
    \centering
    \begin{subfigure}[b]{0.24\textwidth} 
        \includegraphics[width=\textwidth]{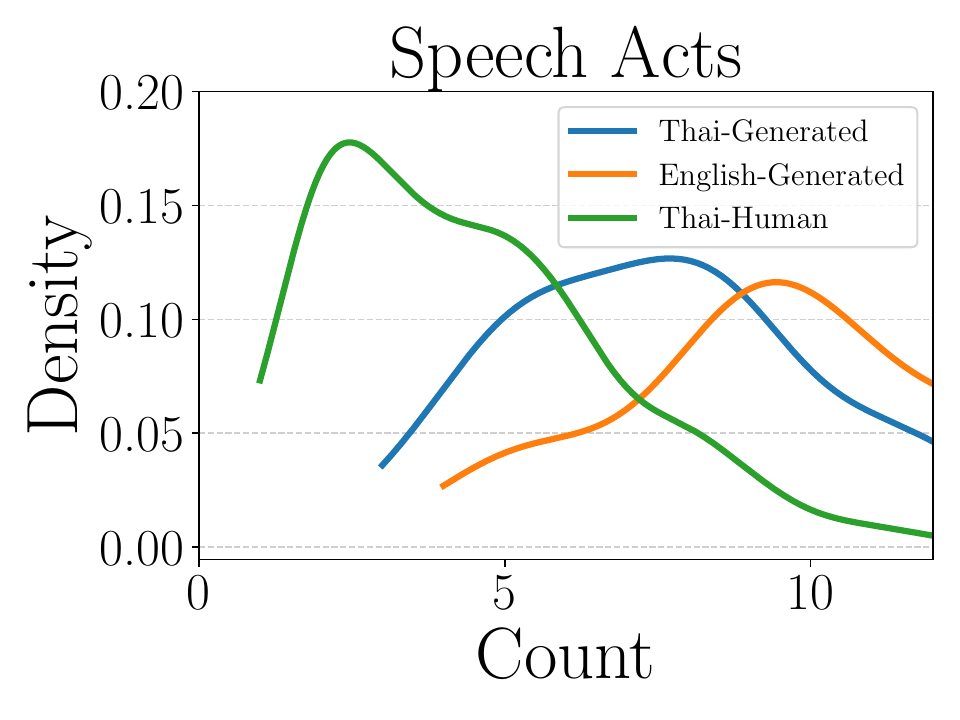}
        \caption{Speech Acts}
    \end{subfigure}
    \hfill 
    \begin{subfigure}[b]{0.24\textwidth} 
        \includegraphics[width=\textwidth]{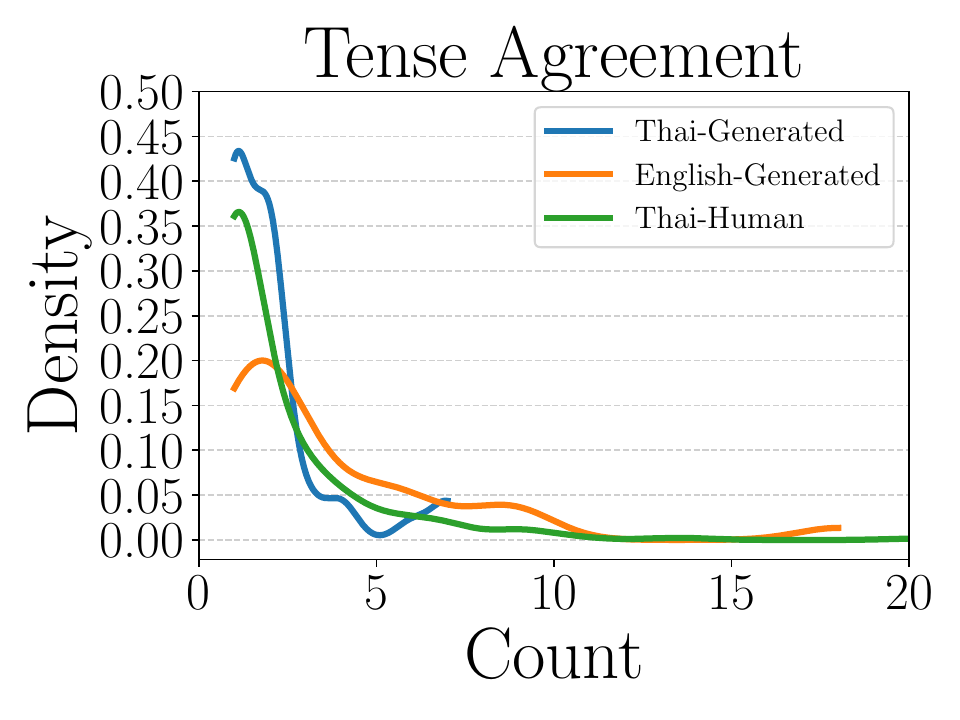}
        \caption{Tense Agreement}
    \end{subfigure}
    \hfill
    \begin{subfigure}[b]{0.24\textwidth}
        \includegraphics[width=\textwidth]{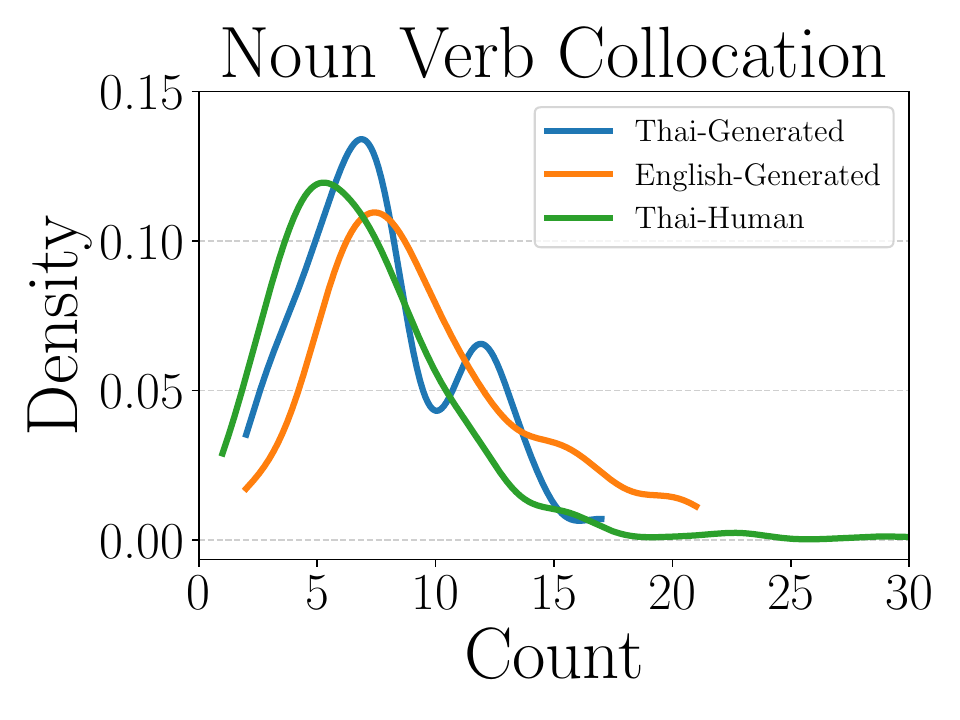}
        \caption{Noun-Verb Collocation}
    \end{subfigure}
    \hfill
    \begin{subfigure}[b]{0.24\textwidth}
    \includegraphics[width=\textwidth]{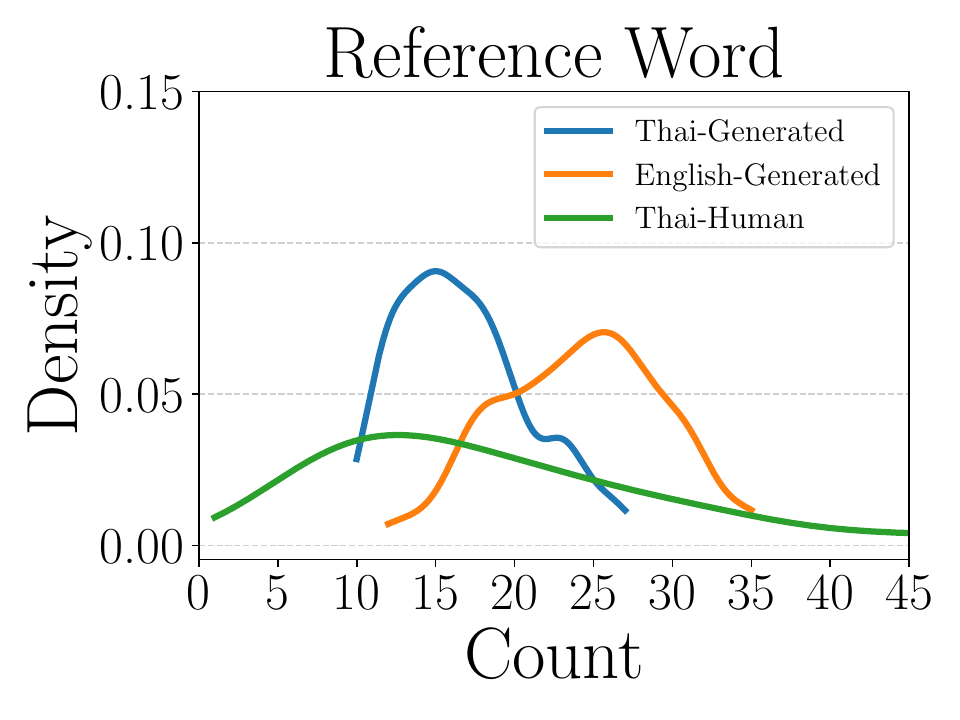}
        \caption{Reference Word}
    \end{subfigure}
    \\
    \begin{subfigure}[b]{0.24\textwidth}
        \includegraphics[width=\textwidth]{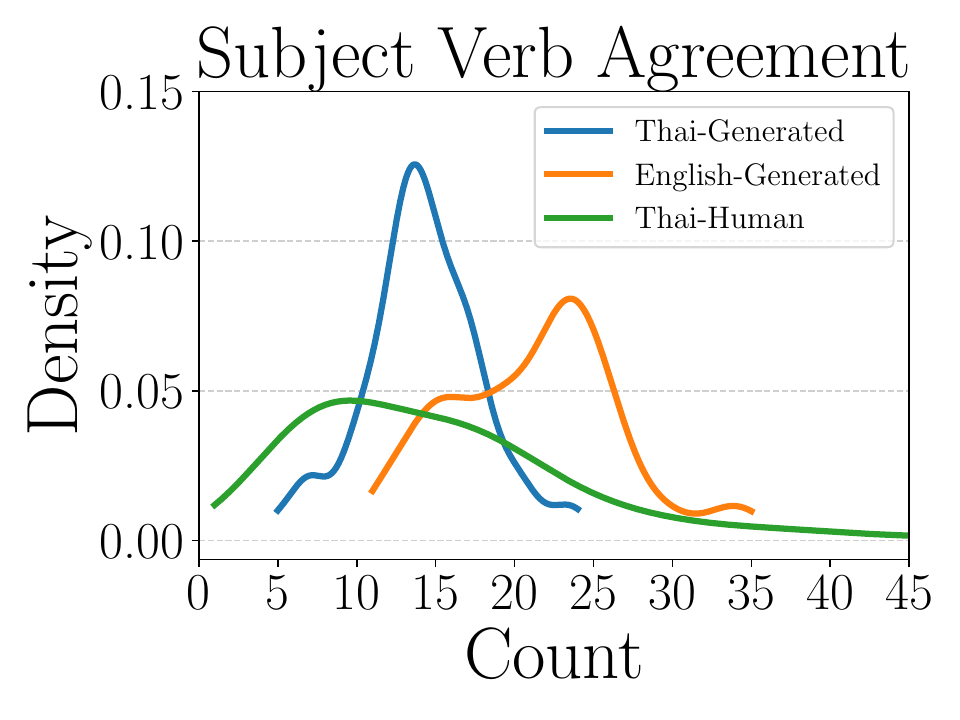}
        \caption{Subject Verb Agreement}
    \end{subfigure}
    \hfill
    \begin{subfigure}[b]{0.24\textwidth}
        \includegraphics[width=\textwidth]{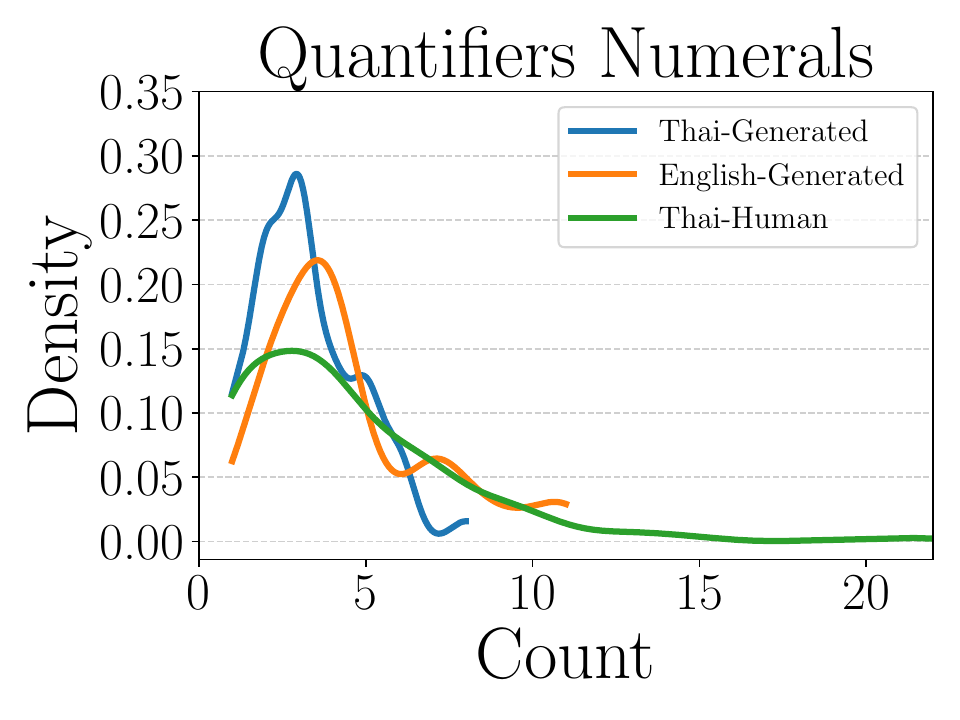}
        \caption{Quantifiers Numerals}
    \end{subfigure}
    \hfill
    \begin{subfigure}[b]{0.24\textwidth}
        \includegraphics[width=\textwidth]{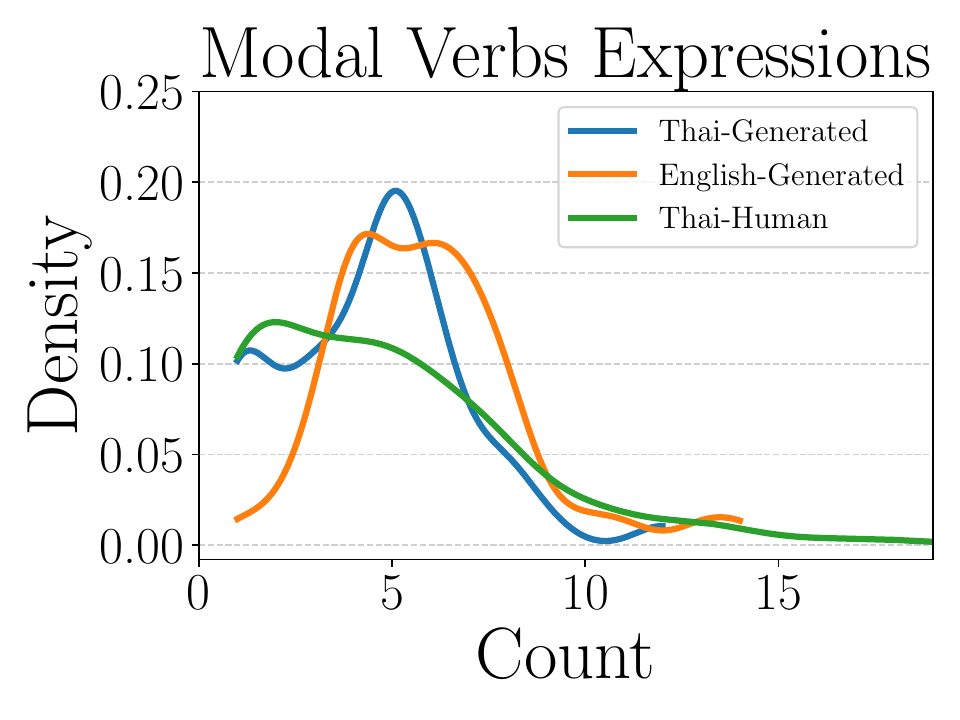}
        \caption{Modal Verbs Expressions}
    \end{subfigure}
    \hfill
    \begin{subfigure}[b]{0.24\textwidth}
    \includegraphics[width=\textwidth]{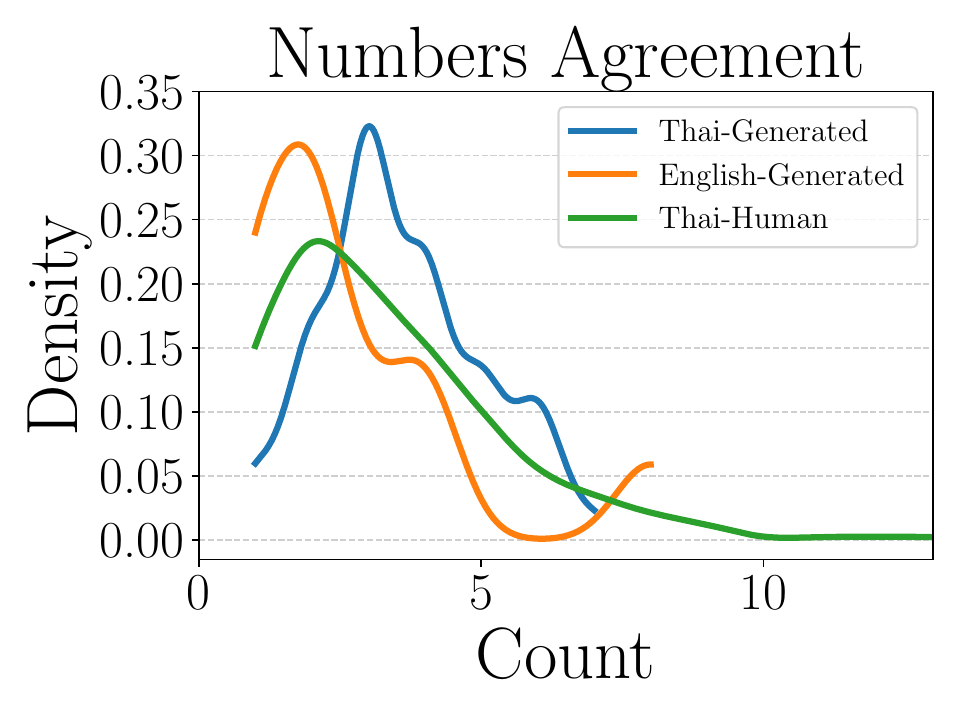}
        \caption{Numbers Agreement}
    \end{subfigure}
    \caption{Full density results for L2 generation dialogue via Thai L1s} 
    \label{fig:L2alldens_tha}
\end{figure*}

\begin{figure*}[!h]
    \centering
    \begin{subfigure}[b]{0.24\textwidth} 
        \includegraphics[width=\textwidth]{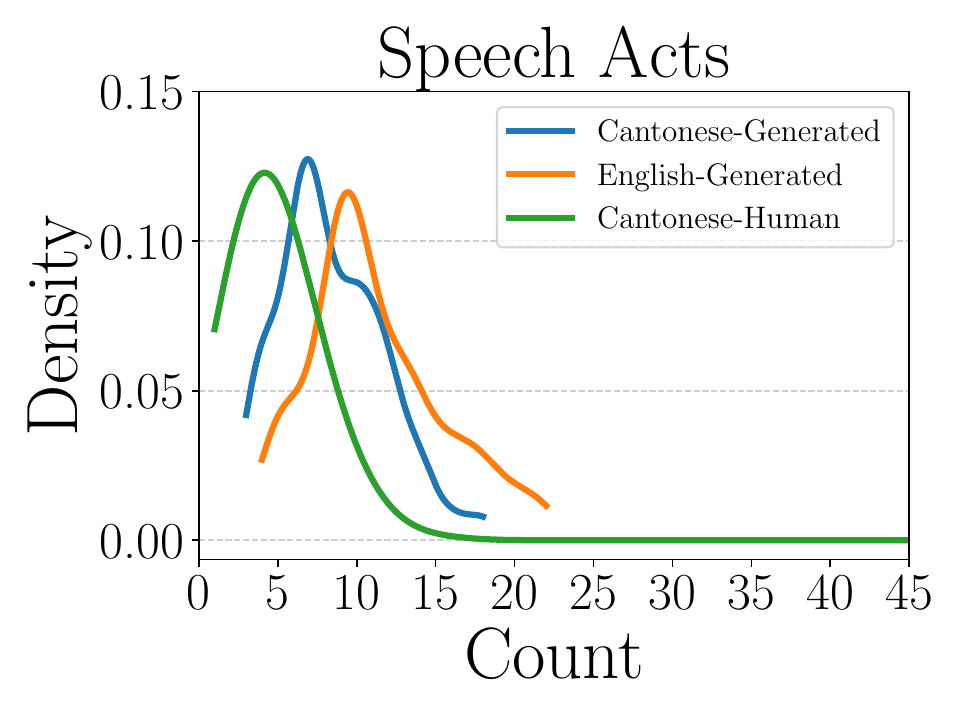}
        \caption{Speech Acts}
    \end{subfigure}
    \hfill 
    \begin{subfigure}[b]{0.24\textwidth} 
        \includegraphics[width=\textwidth]{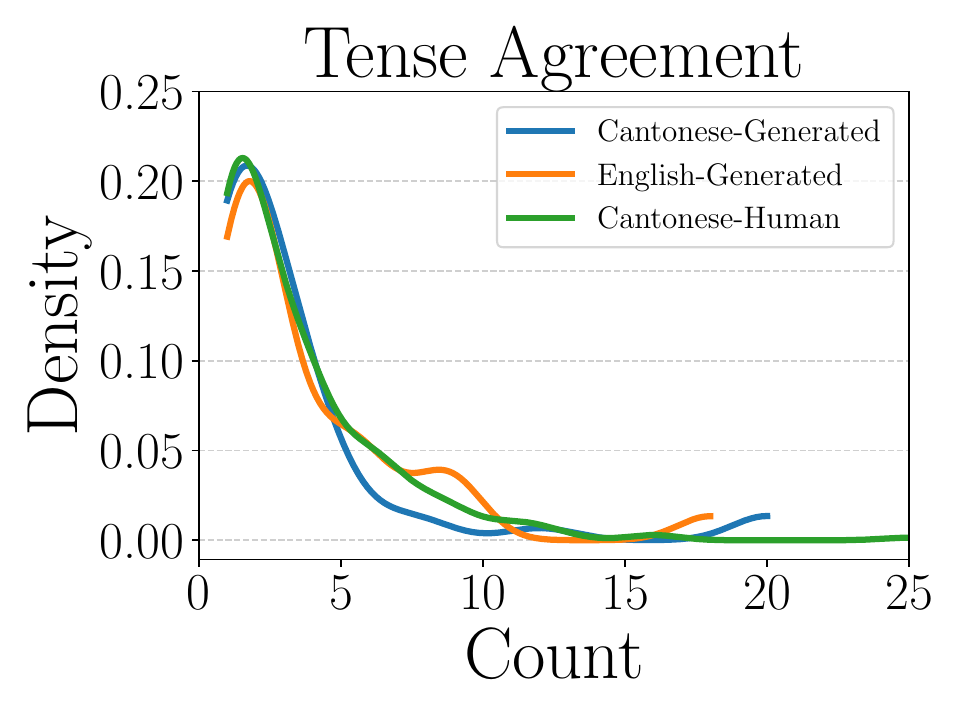}
        \caption{Tense Agreement}
    \end{subfigure}
    \hfill
    \begin{subfigure}[b]{0.24\textwidth}
        \includegraphics[width=\textwidth]{appendix_figs/Noun_Verb_Collocation_Cantonese.pdf}
        \caption{Noun-Verb Collocation}
    \end{subfigure}
    \hfill
    \begin{subfigure}[b]{0.24\textwidth}
    \includegraphics[width=\textwidth]{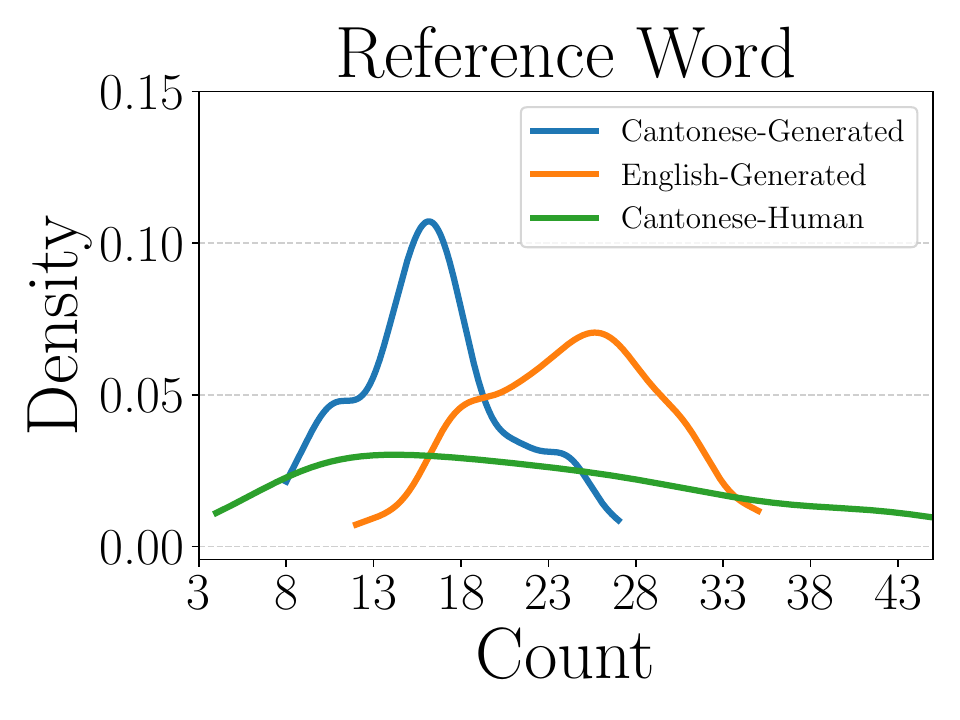}
        \caption{Reference Word}
    \end{subfigure}
    \\
    \begin{subfigure}[b]{0.24\textwidth}
        \includegraphics[width=\textwidth]{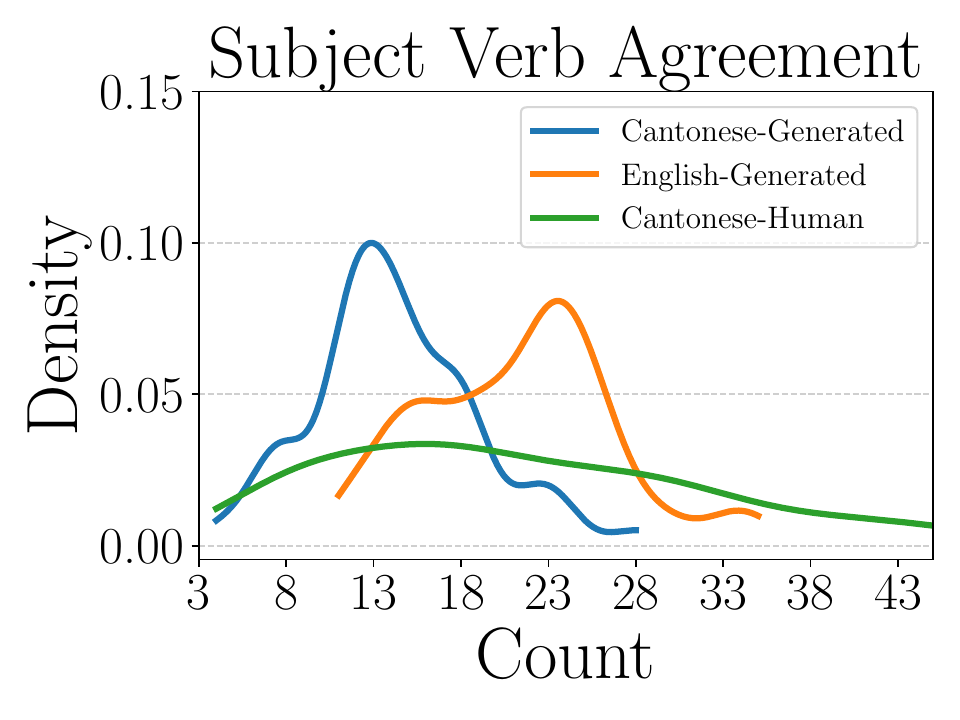}
        \caption{Subject Verb Agreement}
    \end{subfigure}
    \hfill
    \begin{subfigure}[b]{0.24\textwidth}
        \includegraphics[width=\textwidth]{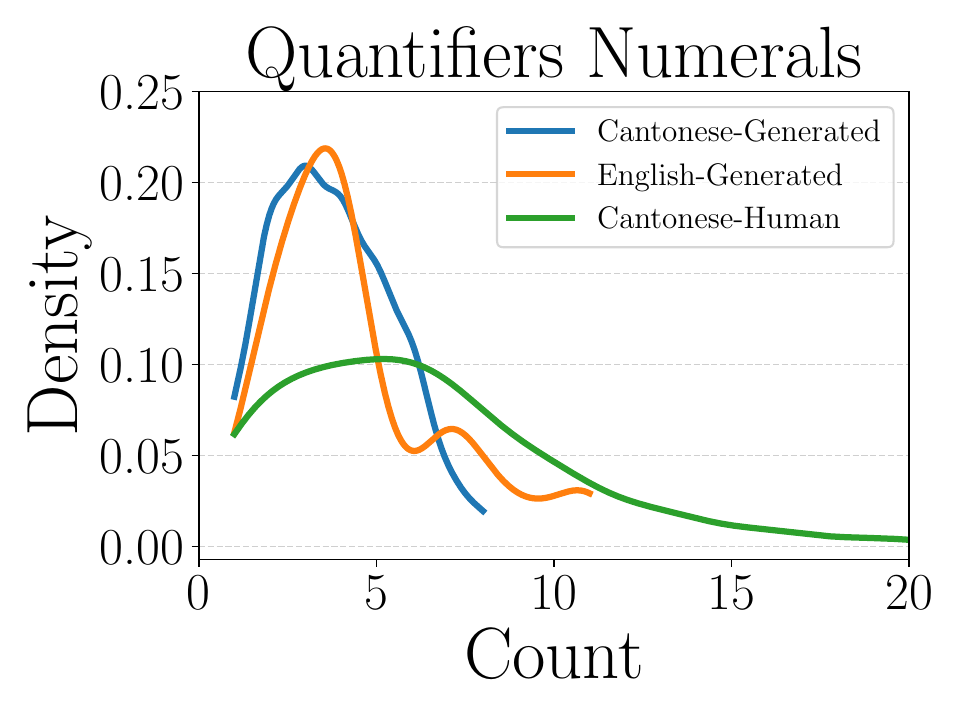}
        \caption{Quantifiers Numerals}
    \end{subfigure}
    \hfill
    \begin{subfigure}[b]{0.24\textwidth}
        \includegraphics[width=\textwidth]{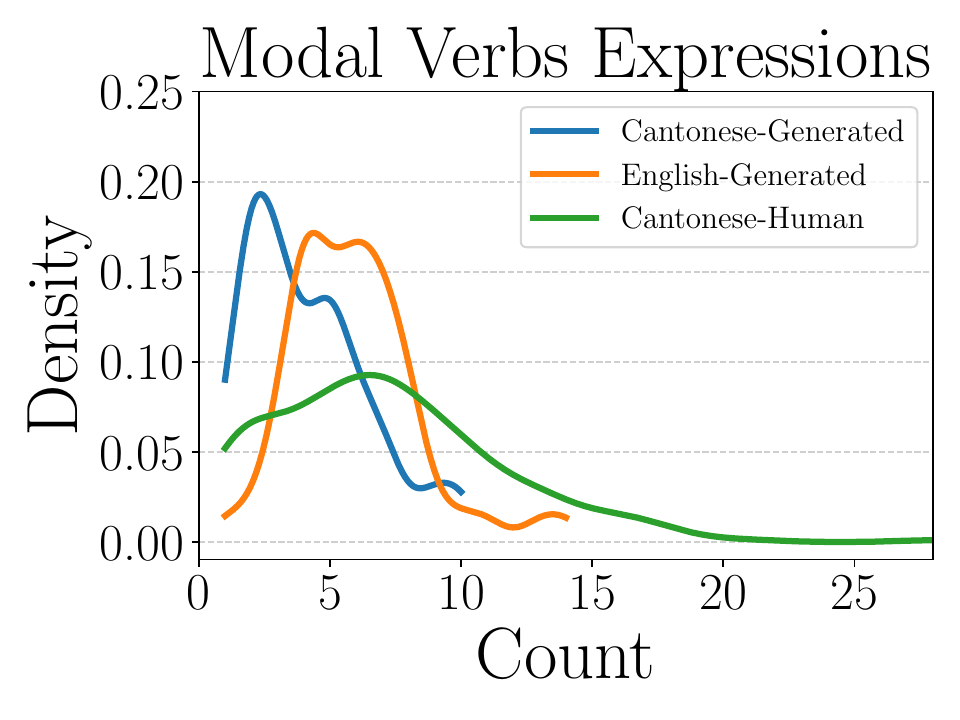}
        \caption{Modal Verbs Expressions}
    \end{subfigure}
    \hfill
    \begin{subfigure}[b]{0.24\textwidth}
    \includegraphics[width=\textwidth]{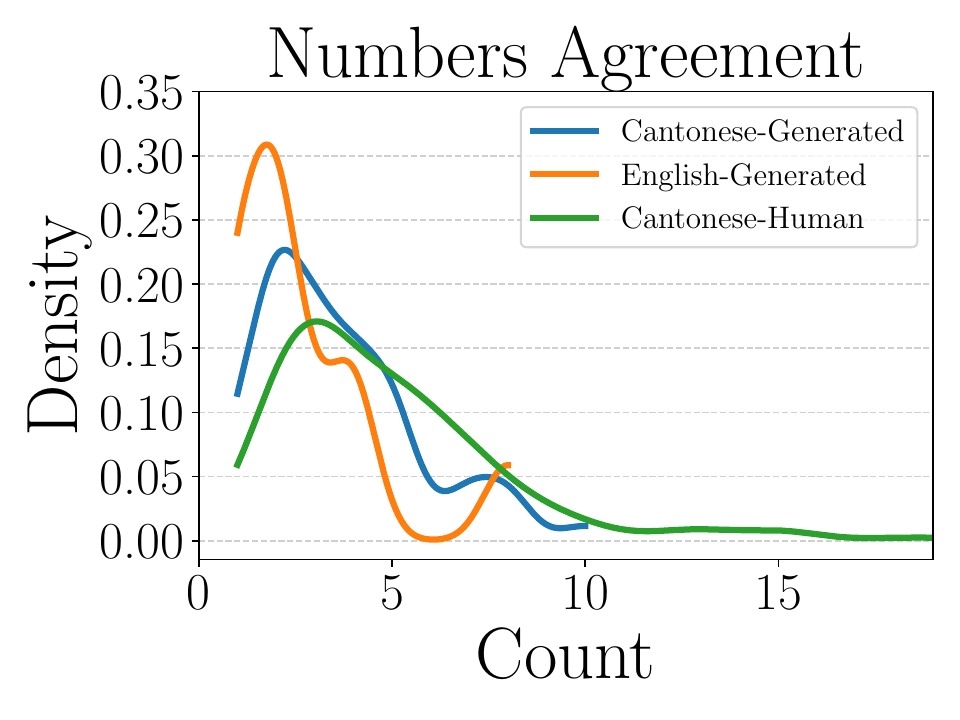}
        \caption{Numbers Agreement}
    \end{subfigure}
    \caption{Full density results for L2 generation dialogue via Cantonese L1s} 
    \label{fig:L2alldens_can}
\end{figure*}

\newpage 
\subsection{Distance Results under different models}\label{sec:allresults}

\begin{table}[!ht]
\centering
\resizebox{0.9\textwidth}{!}{%
\begin{tabular}{llllllllll}
\toprule
 & & \multicolumn{8}{c}{Distribution distance between humans' and LLMs' generated dialogues ($\downarrow$)} \\
 \cmidrule(lr){3-10}
\textbf{Lang.} & \textbf{Condition} & \textbf{\begin{tabular}[c]{@{}c@{}}Number \\ Agreement\end{tabular}} &
  \textbf{\begin{tabular}[c]{@{}c@{}}Tense \\ Agreement\end{tabular}} &
  \textbf{\begin{tabular}[c]{@{}c@{}}Subject-Verb \\ Agreement\end{tabular}} &
  \textbf{\begin{tabular}[c]{@{}c@{}}Modal Verbs \\ Expressions\end{tabular}} &
  \textbf{\begin{tabular}[c]{@{}c@{}}Quantifiers \\ Numerals\end{tabular}} &
  \textbf{\begin{tabular}[c]{@{}c@{}}Noun-Verb \\ Collocation\end{tabular}} &
  \textbf{\begin{tabular}[c]{@{}c@{}}Reference \\ Word\end{tabular}} &
  \textbf{\begin{tabular}[c]{@{}c@{}}Speech \\ Acts\end{tabular}} \\
  \cmidrule(lr){1-1} \cmidrule(lr){2-2} \cmidrule(lr){3-3} \cmidrule(lr){4-4} \cmidrule(lr){5-5} \cmidrule(lr){6-6} \cmidrule(lr){7-7} \cmidrule(lr){8-8} \cmidrule(lr){9-9} \cmidrule(lr){10-10}
\multirow{2}{*}{Cantonese}
& $d_\mathrm{bi}$   & \inc{0.047} & \dec{0.175} & \inc{0.034} & \inc{0.096} & \dec{0.106} & \inc{0.030} & \inc{0.063} & \inc{0.472} \\
& $d_\mathrm{mono}$ & 0.180       & 0.159       & 0.213       & 0.170       & 0.057       & 0.038       & 0.207       & 1.109  \\
  \cmidrule(lr){1-1} \cmidrule(lr){2-2} \cmidrule(lr){3-3} \cmidrule(lr){4-4} \cmidrule(lr){5-5} \cmidrule(lr){6-6} \cmidrule(lr){7-7} \cmidrule(lr){8-8} \cmidrule(lr){9-9} \cmidrule(lr){10-10}
\multirow{2}{*}{Thai}
& $d_\mathrm{bi}$   & \inc{0.053} & \inc{0.050} & \inc{0.084} & \inc{0.117} & \dec{0.055} & \inc{0.103} & \inc{0.038} & \inc{0.802} \\
& $d_\mathrm{mono}$ & 0.125       & 0.131       & 0.439       & 0.189       & 0.033       & 0.217       & 0.341       & 1.470 \\
  \cmidrule(lr){1-1} \cmidrule(lr){2-2} \cmidrule(lr){3-3} \cmidrule(lr){4-4} \cmidrule(lr){5-5} \cmidrule(lr){6-6} \cmidrule(lr){7-7} \cmidrule(lr){8-8} \cmidrule(lr){9-9} \cmidrule(lr){10-10}
\multirow{2}{*}{Japanese}
& $d_\mathrm{bi}$   & \inc{0.055} & \inc{0.027} & \inc{0.175} & \inc{0.142} & \inc{0.042} & \inc{0.024} & \inc{0.266} & \inc{1.324} \\
& $d_\mathrm{mono}$ & 0.087       & 0.106       & 0.510       & 0.403       & 0.162       & 0.252       & 0.382       & 2.301 \\
  \cmidrule(lr){1-1} \cmidrule(lr){2-2} \cmidrule(lr){3-3} \cmidrule(lr){4-4} \cmidrule(lr){5-5} \cmidrule(lr){6-6} \cmidrule(lr){7-7} \cmidrule(lr){8-8} \cmidrule(lr){9-9} \cmidrule(lr){10-10}
\multirow{2}{*}{Korean}
& $d_\mathrm{bi}$   & \inc{0.026} & \inc{0.027} & \inc{0.039} & \inc{0.014} & \inc{0.023} & \inc{0.039} & \inc{0.056} & \inc{1.611} \\
& $d_\mathrm{mono}$ & 0.141       & 0.132       & 0.270       & 0.188       & 0.075       & 0.259       & 0.234       & 2.781 \\
  \cmidrule(lr){1-1} \cmidrule(lr){2-2} \cmidrule(lr){3-3} \cmidrule(lr){4-4} \cmidrule(lr){5-5} \cmidrule(lr){6-6} \cmidrule(lr){7-7} \cmidrule(lr){8-8} \cmidrule(lr){9-9} \cmidrule(lr){10-10}
\multirow{2}{*}{Malay}
& $d_\mathrm{bi}$   & \inc{0.064} & \inc{0.069} & \inc{0.024} & \inc{0.058} & \dec{0.074} & \inc{0.020} & \inc{0.046} & \inc{0.712} \\
& $d_\mathrm{mono}$ & 0.118       & 0.084       & 0.250       & 0.123       & 0.031       & 0.079       & 0.204       & 1.523 \\
  \cmidrule(lr){1-1} \cmidrule(lr){2-2} \cmidrule(lr){3-3} \cmidrule(lr){4-4} \cmidrule(lr){5-5} \cmidrule(lr){6-6} \cmidrule(lr){7-7} \cmidrule(lr){8-8} \cmidrule(lr){9-9} \cmidrule(lr){10-10}
\multirow{2}{*}{Mandarin}
& $d_\mathrm{bi}$   & \dec{0.123} & \inc{0.024} & \inc{0.009} & \inc{0.025} & \dec{0.107} & \inc{0.132} & \inc{0.024} & \inc{0.666} \\
& $d_\mathrm{mono}$ & 0.099       & 0.086       & 0.319       & 0.135       & 0.043       & 0.171       & 0.267       & 1.523 \\
  \cmidrule(lr){1-1} \cmidrule(lr){2-2} \cmidrule(lr){3-3} \cmidrule(lr){4-4} \cmidrule(lr){5-5} \cmidrule(lr){6-6} \cmidrule(lr){7-7} \cmidrule(lr){8-8} \cmidrule(lr){9-9} \cmidrule(lr){10-10}
\multirow{2}{*}{Urdu}
& $d_\mathrm{bi}$   & \inc{0.012} & \inc{0.038} & \inc{0.049} & \inc{0.031} & \dec{0.080} & \inc{0.009} & \inc{0.132} & \inc{0.399} \\
& $d_\mathrm{mono}$ & 0.116       & 0.135       & 0.182       & 0.155       & 0.076       & 0.098       & 0.266       & 1.114 \\
\bottomrule
\end{tabular}
}
\caption{Distance Results with Deepseek V3 685B} \label{tab:deepseek-res}
\end{table}
\begin{table}[!h]
\centering
\resizebox{0.9\textwidth}{!}{%
\begin{tabular}{llllllllll}
\toprule
 & & \multicolumn{8}{c}{Distribution distance between humans' and LLMs' generated dialogues ($\downarrow$)} \\
 \cmidrule(lr){3-10}
\textbf{Lang.} & \textbf{Condition} & \textbf{\begin{tabular}[c]{@{}c@{}}Number \\ Agreement\end{tabular}} &
  \textbf{\begin{tabular}[c]{@{}c@{}}Tense \\ Agreement\end{tabular}} &
  \textbf{\begin{tabular}[c]{@{}c@{}}Subject-Verb \\ Agreement\end{tabular}} &
  \textbf{\begin{tabular}[c]{@{}c@{}}Modal Verbs \\ Expressions\end{tabular}} &
  \textbf{\begin{tabular}[c]{@{}c@{}}Quantifiers \\ Numerals\end{tabular}} &
  \textbf{\begin{tabular}[c]{@{}c@{}}Noun-Verb \\ Collocation\end{tabular}} &
  \textbf{\begin{tabular}[c]{@{}c@{}}Reference \\ Word\end{tabular}} &
  \textbf{\begin{tabular}[c]{@{}c@{}}Speech \\ Acts\end{tabular}} \\
  \cmidrule(lr){1-1} \cmidrule(lr){2-2} \cmidrule(lr){3-3} \cmidrule(lr){4-4} \cmidrule(lr){5-5} \cmidrule(lr){6-6} \cmidrule(lr){7-7} \cmidrule(lr){8-8} \cmidrule(lr){9-9} \cmidrule(lr){10-10}

\multirow{2}{*}{Cantonese}
& $d_{\mathrm{bi}}$   & \dec{0.085} & \dec{0.037} & \inc{0.040} & \dec{0.187} & \dec{0.141} & \inc{0.005} & \inc{0.230} & \inc{0.532} \\
& $d_{\mathrm{mono}}$ & 0.064       & 0.029       & 0.507       & 0.090       & 0.026       & 0.041       & 0.288       & 0.879       \\
  \cmidrule(lr){1-1} \cmidrule(lr){2-2} \cmidrule(lr){3-3} \cmidrule(lr){4-4} \cmidrule(lr){5-5} \cmidrule(lr){6-6} \cmidrule(lr){7-7} \cmidrule(lr){8-8} \cmidrule(lr){9-9} \cmidrule(lr){10-10}
\multirow{2}{*}{Thai}
& $d_{\mathrm{bi}}$   & \dec{0.294} & \inc{0.115} & \inc{0.253} & \dec{0.053} & \inc{0.046} & \inc{0.129} & \inc{0.247} & \inc{0.998} \\
& $d_{\mathrm{mono}}$ & 0.179       & 0.139       & 0.801       & 0.027       & 0.077       & 0.264       & 0.448       & 1.161       \\
  \cmidrule(lr){1-1} \cmidrule(lr){2-2} \cmidrule(lr){3-3} \cmidrule(lr){4-4} \cmidrule(lr){5-5} \cmidrule(lr){6-6} \cmidrule(lr){7-7} \cmidrule(lr){8-8} \cmidrule(lr){9-9} \cmidrule(lr){10-10}

\multirow{2}{*}{Japanese}
& $d_{\mathrm{bi}}$   & \dec{0.211} & \inc{0.099} & \inc{0.486} & \inc{0.061} & \inc{0.076} & \inc{0.124} & \dec{0.559} & \inc{1.425} \\
& $d_{\mathrm{mono}}$ & 0.117       & 0.290       & 0.928       & 0.166       & 0.256       & 0.306       & 0.498       & 1.980       \\
  \cmidrule(lr){1-1} \cmidrule(lr){2-2} \cmidrule(lr){3-3} \cmidrule(lr){4-4} \cmidrule(lr){5-5} \cmidrule(lr){6-6} \cmidrule(lr){7-7} \cmidrule(lr){8-8} \cmidrule(lr){9-9} \cmidrule(lr){10-10}

\multirow{2}{*}{Korean}
& $d_{\mathrm{bi}}$   & \dec{0.212} & \dec{0.023} & \inc{0.138} & \inc{0.032} & \inc{0.020} & \inc{0.164} & \inc{0.094} & \inc{1.784} \\
& $d_{\mathrm{mono}}$ & 0.158       & 0.004       & 0.608       & 0.054       & 0.112       & 0.301       & 0.324       & 2.379       \\
  \cmidrule(lr){1-1} \cmidrule(lr){2-2} \cmidrule(lr){3-3} \cmidrule(lr){4-4} \cmidrule(lr){5-5} \cmidrule(lr){6-6} \cmidrule(lr){7-7} \cmidrule(lr){8-8} \cmidrule(lr){9-9} \cmidrule(lr){10-10}

\multirow{2}{*}{Malay}
& $d_{\mathrm{bi}}$   & \dec{0.108} & \inc{0.033} & \inc{0.050} & \dec{0.095} & \inc{0.049} & \inc{0.086} & \inc{0.114} & \inc{0.834} \\
& $d_{\mathrm{mono}}$ & 0.033       & 0.138       & 0.572       & 0.016       & 0.087       & 0.099       & 0.285       & 1.253       \\
  \cmidrule(lr){1-1} \cmidrule(lr){2-2} \cmidrule(lr){3-3} \cmidrule(lr){4-4} \cmidrule(lr){5-5} \cmidrule(lr){6-6} \cmidrule(lr){7-7} \cmidrule(lr){8-8} \cmidrule(lr){9-9} \cmidrule(lr){10-10}

\multirow{2}{*}{Mandarin}
& $d_{\mathrm{bi}}$   & \inc{0.060} & \inc{0.040} & \inc{0.029} & \dec{0.066} & \inc{0.018} & \inc{0.154} & \inc{0.096} & \inc{0.813} \\
& $d_{\mathrm{mono}}$ & 0.074       & 0.140       & 0.668       & 0.016       & 0.062       & 0.201       & 0.361       & 1.247       \\
  \cmidrule(lr){1-1} \cmidrule(lr){2-2} \cmidrule(lr){3-3} \cmidrule(lr){4-4} \cmidrule(lr){5-5} \cmidrule(lr){6-6} \cmidrule(lr){7-7} \cmidrule(lr){8-8} \cmidrule(lr){9-9} \cmidrule(lr){10-10}

\multirow{2}{*}{Urdu}
& $d_{\mathrm{bi}}$   & \inc{0.037} & \inc{0.026} & \inc{0.179} & \dec{0.097} & \dec{0.131} & \inc{0.040} & \dec{0.385} & \inc{0.792} \\
& $d_{\mathrm{mono}}$ & 0.045       & 0.063       & 0.463       & 0.022       & 0.072       & 0.122       & 0.364       & 0.886       \\
\bottomrule
\end{tabular}
}
\caption{Distance Results with QWEN2.5 72B} \label{tab:qwen-res}
\end{table}

\begin{table}[!h]
\centering
\resizebox{0.9\textwidth}{!}{%
\begin{tabular}{llllllllll}
\toprule
 & & \multicolumn{8}{c}{Distribution distance between humans' and LLMs' generated dialogues ($\downarrow$)} \\
 \cmidrule(lr){3-10}
\textbf{Lang.} & \textbf{Condition} & \textbf{\begin{tabular}[c]{@{}c@{}}Number \\ Agreement\end{tabular}} &
  \textbf{\begin{tabular}[c]{@{}c@{}}Tense \\ Agreement\end{tabular}} &
  \textbf{\begin{tabular}[c]{@{}c@{}}Subject-Verb \\ Agreement\end{tabular}} &
  \textbf{\begin{tabular}[c]{@{}c@{}}Modal Verbs \\ Expressions\end{tabular}} &
  \textbf{\begin{tabular}[c]{@{}c@{}}Quantifiers \\ Numerals\end{tabular}} &
  \textbf{\begin{tabular}[c]{@{}c@{}}Noun-Verb \\ Collocation\end{tabular}} &
  \textbf{\begin{tabular}[c]{@{}c@{}}Reference \\ Word\end{tabular}} &
  \textbf{\begin{tabular}[c]{@{}c@{}}Speech \\ Acts\end{tabular}} \\
  \cmidrule(lr){1-1} \cmidrule(lr){2-2} \cmidrule(lr){3-3} \cmidrule(lr){4-4} \cmidrule(lr){5-5} \cmidrule(lr){6-6} \cmidrule(lr){7-7} \cmidrule(lr){8-8} \cmidrule(lr){9-9} \cmidrule(lr){10-10}
\multirow{2}{*}{Cantonese}
& $d_{\mathrm{bi}}$   & \inc{0.012} & \dec{0.349} & \inc{0.018} & \dec{0.195} & \dec{0.393} & \inc{0.009} & \dec{0.110} & \inc{0.451} \\
& $d_{\mathrm{mono}}$ & 0.179       & 0.007       & 0.496       & 0.078       & 0.050       & 0.091       & 0.003       & 1.160       \\
  \cmidrule(lr){1-1} \cmidrule(lr){2-2} \cmidrule(lr){3-3} \cmidrule(lr){4-4} \cmidrule(lr){5-5} \cmidrule(lr){6-6} \cmidrule(lr){7-7} \cmidrule(lr){8-8} \cmidrule(lr){9-9} \cmidrule(lr){10-10}
\multirow{2}{*}{Thai}
& $d_{\mathrm{bi}}$   & \inc{0.071} & \inc{0.048} & \inc{0.067} & \inc{0.022} & \dec{0.114} & \inc{0.169} & \dec{0.173} & \inc{1.127} \\
& $d_{\mathrm{mono}}$ & 0.321       & 0.073       & 0.788       & 0.061       & 0.016       & 0.398       & 0.007       & 1.572       \\
  \cmidrule(lr){1-1} \cmidrule(lr){2-2} \cmidrule(lr){3-3} \cmidrule(lr){4-4} \cmidrule(lr){5-5} \cmidrule(lr){6-6} \cmidrule(lr){7-7} \cmidrule(lr){8-8} \cmidrule(lr){9-9} \cmidrule(lr){10-10}
\multirow{2}{*}{Japanese}
& $d_{\mathrm{bi}}$   & \inc{0.020} & \dec{0.178} & \inc{0.113} & \inc{0.032} & \inc{0.065} & \inc{0.079} & \dec{0.256} & \inc{1.627} \\
& $d_{\mathrm{mono}}$ & 0.234       & 0.101       & 0.912       & 0.239       & 0.134       & 0.443       & 0.046       & 2.378       \\
  \cmidrule(lr){1-1} \cmidrule(lr){2-2} \cmidrule(lr){3-3} \cmidrule(lr){4-4} \cmidrule(lr){5-5} \cmidrule(lr){6-6} \cmidrule(lr){7-7} \cmidrule(lr){8-8} \cmidrule(lr){9-9} \cmidrule(lr){10-10}
\multirow{2}{*}{Korean}
& $d_{\mathrm{bi}}$   & \inc{0.027} & \dec{0.035} & \inc{0.012} & \inc{0.067} & \dec{0.289} & \inc{0.161} & \dec{0.046} & \inc{1.612} \\
& $d_{\mathrm{mono}}$ & 0.332       & 0.003       & 0.594       & 0.090       & 0.032       & 0.444       & 0.005       & 2.831       \\
  \cmidrule(lr){1-1} \cmidrule(lr){2-2} \cmidrule(lr){3-3} \cmidrule(lr){4-4} \cmidrule(lr){5-5} \cmidrule(lr){6-6} \cmidrule(lr){7-7} \cmidrule(lr){8-8} \cmidrule(lr){9-9} \cmidrule(lr){10-10}
\multirow{2}{*}{Malay}
& $d_{\mathrm{bi}}$   & \inc{0.034} & \inc{0.033} & \inc{0.053} & \dec{0.068} & \dec{0.200} & \inc{0.025} & \dec{0.258} & \inc{0.934} \\
& $d_{\mathrm{mono}}$ & 0.125       & 0.043       & 0.559       & 0.027       & 0.010       & 0.179       & 0.001       & 1.585       \\
  \cmidrule(lr){1-1} \cmidrule(lr){2-2} \cmidrule(lr){3-3} \cmidrule(lr){4-4} \cmidrule(lr){5-5} \cmidrule(lr){6-6} \cmidrule(lr){7-7} \cmidrule(lr){8-8} \cmidrule(lr){9-9} \cmidrule(lr){10-10}
\multirow{2}{*}{Mandarin}
& $d_{\mathrm{bi}}$   & \inc{0.047} & \dec{0.068} & \inc{0.010} & \dec{0.142} & \dec{0.123} & \inc{0.054} & \dec{0.102} & \inc{0.862} \\
& $d_{\mathrm{mono}}$ & 0.190       & 0.053       & 0.655       & 0.031       & 0.012       & 0.308       & 0.005       & 1.584       \\
  \cmidrule(lr){1-1} \cmidrule(lr){2-2} \cmidrule(lr){3-3} \cmidrule(lr){4-4} \cmidrule(lr){5-5} \cmidrule(lr){6-6} \cmidrule(lr){7-7} \cmidrule(lr){8-8} \cmidrule(lr){9-9} \cmidrule(lr){10-10}
\multirow{2}{*}{Urdu}
& $d_{\mathrm{bi}}$   & \inc{0.024} & \dec{0.093} & \inc{0.046} & \dec{0.065} & \dec{0.337} & \inc{0.049} & \dec{0.204} & \inc{0.571} \\
& $d_{\mathrm{mono}}$ & 0.148       & 0.019       & 0.452       & 0.040       & 0.027       & 0.212       & 0.007       & 1.163       \\

\bottomrule
\end{tabular}
}
\caption{Distance Results with LLAMA 70B} \label{tab:llama70b-res}
\end{table}

\newpage 
\begin{table}[!ht]
\centering
\resizebox{0.9\textwidth}{!}{%
\begin{tabular}{llllllllll}
\toprule
 & & \multicolumn{8}{c}{Distribution distance between humans' and LLMs' generated dialogues ($\downarrow$)} \\
 \cmidrule(lr){3-10}
\textbf{Lang.} & \textbf{Condition} & \textbf{\begin{tabular}[c]{@{}c@{}}Number \\ Agreement\end{tabular}} &
  \textbf{\begin{tabular}[c]{@{}c@{}}Tense \\ Agreement\end{tabular}} &
  \textbf{\begin{tabular}[c]{@{}c@{}}Subject-Verb \\ Agreement\end{tabular}} &
  \textbf{\begin{tabular}[c]{@{}c@{}}Modal Verbs \\ Expressions\end{tabular}} &
  \textbf{\begin{tabular}[c]{@{}c@{}}Quantifiers \\ Numerals\end{tabular}} &
  \textbf{\begin{tabular}[c]{@{}c@{}}Noun-Verb \\ Collocation\end{tabular}} &
  \textbf{\begin{tabular}[c]{@{}c@{}}Reference \\ Word\end{tabular}} &
  \textbf{\begin{tabular}[c]{@{}c@{}}Speech \\ Acts\end{tabular}} \\
  \cmidrule(lr){1-1} \cmidrule(lr){2-2} \cmidrule(lr){3-3} \cmidrule(lr){4-4} \cmidrule(lr){5-5} \cmidrule(lr){6-6} \cmidrule(lr){7-7} \cmidrule(lr){8-8} \cmidrule(lr){9-9} \cmidrule(lr){10-10}
\multirow{2}{*}{Cantonese}
& $d_{\mathrm{bi}}$   
  & \dec{0.090} & \dec{0.469} & \dec{0.156} & \dec{0.458} & \dec{0.294} & \dec{0.219} & \inc{0.005} & \inc{0.088} \\
& $d_{\mathrm{mono}}$ 
  & 0.046 & 0.044 & 0.007 & 0.097 & 0.086 & 0.067 & 0.021 & 0.791 \\
  \cmidrule(lr){1-1} \cmidrule(lr){2-2} \cmidrule(lr){3-3} \cmidrule(lr){4-4} \cmidrule(lr){5-5} \cmidrule(lr){6-6} \cmidrule(lr){7-7} \cmidrule(lr){8-8} \cmidrule(lr){9-9} \cmidrule(lr){10-10}
\multirow{2}{*}{Thai}
& $d_{\mathrm{bi}}$
  & \inc{0.088} & \inc{0.055} & \inc{0.023} & \inc{0.091} & \dec{0.064} & \inc{0.012} & \dec{0.053} & \inc{0.428} \\
& $d_{\mathrm{mono}}$
  & 0.198 & 0.060 & 0.038 & 0.105 & 0.019 & 0.348 & 0.032 & 1.041 \\
  \cmidrule(lr){1-1} \cmidrule(lr){2-2} \cmidrule(lr){3-3} \cmidrule(lr){4-4} \cmidrule(lr){5-5} \cmidrule(lr){6-6} \cmidrule(lr){7-7} \cmidrule(lr){8-8} \cmidrule(lr){9-9} \cmidrule(lr){10-10}
\multirow{2}{*}{Japanese}
& $d_{\mathrm{bi}}$
  & \dec{0.167} & \inc{0.109} & \inc{0.046} & \inc{0.019} & \inc{0.025} & \inc{0.010} & \inc{0.009} & \inc{0.766} \\
& $d_{\mathrm{mono}}$
  & 0.149 & 0.168 & 0.097 & 0.302 & 0.113 & 0.394 & 0.095 & 1.848 \\
  \cmidrule(lr){1-1} \cmidrule(lr){2-2} \cmidrule(lr){3-3} \cmidrule(lr){4-4} \cmidrule(lr){5-5} \cmidrule(lr){6-6} \cmidrule(lr){7-7} \cmidrule(lr){8-8} \cmidrule(lr){9-9} \cmidrule(lr){10-10}
\multirow{2}{*}{Korean}
& $d_{\mathrm{bi}}$
  & \inc{0.044} & \dec{0.235} & \inc{0.001} & \inc{0.058} & \dec{0.085} & \inc{0.125} & \inc{0.000} & \inc{1.315} \\
& $d_{\mathrm{mono}}$
  & 0.140 & 0.035 & 0.025 & 0.136 & 0.004 & 0.380 & 0.040 & 2.233 \\
  \cmidrule(lr){1-1} \cmidrule(lr){2-2} \cmidrule(lr){3-3} \cmidrule(lr){4-4} \cmidrule(lr){5-5} \cmidrule(lr){6-6} \cmidrule(lr){7-7} \cmidrule(lr){8-8} \cmidrule(lr){9-9} \cmidrule(lr){10-10}
\multirow{2}{*}{Malay}
& $d_{\mathrm{bi}}$
  & \dec{0.082} & \dec{0.237} & \dec{0.053} & \dec{0.391} & \dec{0.149} & \inc{0.025} & \inc{0.008} & \inc{0.409} \\
& $d_{\mathrm{mono}}$
  & 0.045 & 0.028 & 0.018 & 0.046 & 0.015 & 0.145 & 0.015 & 1.151 \\
  \cmidrule(lr){1-1} \cmidrule(lr){2-2} \cmidrule(lr){3-3} \cmidrule(lr){4-4} \cmidrule(lr){5-5} \cmidrule(lr){6-6} \cmidrule(lr){7-7} \cmidrule(lr){8-8} \cmidrule(lr){9-9} \cmidrule(lr){10-10}
\multirow{2}{*}{Mandarin}
& $d_{\mathrm{bi}}$
  & \inc{0.011} & \dec{0.097} & \dec{0.094} & \dec{0.176} & \dec{0.085} & \inc{0.191} & \dec{0.028} & \inc{0.375} \\
& $d_{\mathrm{mono}}$
  & 0.074 & 0.034 & 0.031 & 0.062 & 0.010 & 0.264 & 0.025 & 1.144 \\
  \cmidrule(lr){1-1} \cmidrule(lr){2-2} \cmidrule(lr){3-3} \cmidrule(lr){4-4} \cmidrule(lr){5-5} \cmidrule(lr){6-6} \cmidrule(lr){7-7} \cmidrule(lr){8-8} \cmidrule(lr){9-9} \cmidrule(lr){10-10}
\multirow{2}{*}{English}
& $d_{\mathrm{bi}}$
  & \dec{0.008} & \dec{0.134} & \dec{0.009} & \dec{0.062} & \dec{0.059} & \dec{0.079} & \dec{0.015} & \dec{0.823} \\
& $d_{\mathrm{mono}}$
  & 0.008 & 0.134 & 0.009 & 0.062 & 0.059 & 0.079 & 0.015 & 0.823 \\
  \cmidrule(lr){1-1} \cmidrule(lr){2-2} \cmidrule(lr){3-3} \cmidrule(lr){4-4} \cmidrule(lr){5-5} \cmidrule(lr){6-6} \cmidrule(lr){7-7} \cmidrule(lr){8-8} \cmidrule(lr){9-9} \cmidrule(lr){10-10}
\multirow{2}{*}{Urdu}
& $d_{\mathrm{bi}}$
  & \dec{0.099} & \inc{0.011} & \dec{0.042} & \dec{0.188} & \dec{0.054} & \inc{0.020} & \inc{0.010} & \inc{0.265} \\
& $d_{\mathrm{mono}}$
  & 0.057 & 0.016 & 0.016 & 0.075 & 0.004 & 0.174 & 0.035 & 0.798 \\
\bottomrule
\end{tabular}
}
\caption{Distance Results with LLAMA 8B} \label{tab:llama8b-res}
\end{table}

\end{document}